\documentclass{pkuai4m}

\usepackage{microtype}
\usepackage{graphicx}
\usepackage{pdfpages}
\usepackage{subcaption}
\usepackage{booktabs} 

\usepackage{hyperref}

\usepackage{amsmath}
\usepackage{amssymb}
\usepackage{mathtools}
\usepackage{amsthm}
\usepackage{algorithm}
\usepackage{algorithmic}
\usepackage[most]{tcolorbox}
\usepackage{caption}
\usepackage{multirow,multicol,booktabs,array}

\usepackage{fvextra}
\usepackage{enumitem}

\renewcommand{\titlefont}{\fontsize{16}{18}\selectfont}
\renewcommand\affiliationformat[2][]{{\affiliationfont {\sffamily $^{#1}$#2}}}

\makeatletter
\@ifundefined{Highlighting}{}{
  \DefineVerbatimEnvironment{Highlighting}{Verbatim}{
    breaklines=true,
    fontsize=\small
  }
}
\makeatother

\fvset{
  breaklines=true
}

\usepackage[fencedCode,hybrid,frozencache]{markdown}
\makeatletter

\def\markdownRendererUlBeginPrototype{\begin{itemize}[leftmargin=*,nosep]}
\def\markdownRendererUlBeginTightPrototype{\begin{itemize}[leftmargin=*,nosep]}

\def\markdownRendererUlEndPrototype{\end{itemize}}
\def\markdownRendererUlEndTightPrototype{\end{itemize}}
\def\markdownRendererOlBeginPrototype{\begin{enumerate}[leftmargin=*,nosep]}
\def\markdownRendererOlBeginTightPrototype{\begin{enumerate}[leftmargin=*,nosep]}

\def\markdownRendererOlItemWithNumberPrototype#1{\item\relax}

\def\markdownRendererOlEndPrototype{\end{enumerate}}
\def\markdownRendererOlEndTightPrototype{\end{enumerate}}

\makeatother

\theoremstyle{plain}
\newtheorem{theorem}{Theorem}[section]

\theoremstyle{definition}

\newtheorem{assumption}[theorem]{Assumption}
\theoremstyle{remark}

\usepackage[textsize=tiny]{todonotes}

\title{ReasFlow: Assisting Reasoning-Centric Scientific Discovery in Applied Mathematics via a Knowledge-Based Multi-Agent System}

\makeatletter
\renewcommand\authorformat[2][]{%
  \authorfont {\sffamily
    \mbox{\ifstrempty{#1}{#2}{#2$^{#1}$}}%
  }%
}
\makeatother

\author[1,*]{Yutong He}
\author[1,*]{Daibo Li}
\author[1,*]{Guohong Li}
\author[1,*]{Jiahe Geng}
\author[1,*]{Zhengyang Huang}
\author[1,*]{Can Ren}
\author[2,\dagger]{Zekun Zhang}
\author[3,\dagger]{Yifan Liu}
\author[1]{Shuchen Zhu}
\author[1]{Hengrui Zhang}
\author[1]{Boao Kong}
\author[1]{Ming Sun}
\author[4]{Shu Li}
\author[1]{Chenyi Li}
\author[4]{Jiang Hu}
\author[1,\P]{Kun Yuan}
\author[1,\P]{Zaiwen Wen}
\author[5,\P]{Pingwen Zhang}

\affiliation[]{
$^{1}$Peking University \\
$^{2}$University of Electronic Science and Technology of China\\
$^{3}$Beijing Normal University\\
$^{4}$Tsinghua University\\
$^{5}$Wuhan University
}
\contribution[*]{Equal Contribution}
\contribution[\dagger]{Work done while interning at Peking University}
\contribution[\P]{Corresponding author}

\abstract{
Recent advances in Large Language Models have fueled autonomous AI agents capable of tackling complex scientific tasks, yet existing automated research systems remain predominantly focused on empirically driven domains with quantitative benchmarks, leaving theory-driven discovery, particularly in mathematically grounded disciplines requiring rigorous proofs and synthesis of domain knowledge, largely underexplored. Key challenges include the difficulty of verifying theoretical reasoning at scale, insufficient reasoning ability for autonomous frontier exploration, and a scarcity of procedural heuristics in the literature. We introduce \textbf{ReasFlow}, an end-to-end autonomous agent system for reasoning-centric scientific discovery that operationalizes a collaborative paradigm where the human expert acts as Principal Investigator while the agent executes rigorous derivations as a capable graduate student. ReasFlow incorporates (i) a robust internal verification loop that audits logical coherence and corrects fundamental errors prior to human inspection, and (ii) an automated knowledge retrieval and self-improvement mechanism that proactively surfaces both declarative facts and overlooked procedural heuristics, substantially reducing expert intervention. The system unifies literature synthesis, algorithm design, theorem proving, experimentation, and manuscript preparation in a single system. Deployed to autonomously generate five complete research papers with rigorous theoretical and empirical content from minimal prompts, ReasFlow consistently achieves the highest evaluation scores among state-of-the-art open-access baselines under a curated LLM-based review rubric. ReasFlow is publicly accessible via the ReasLab platform, providing a collaborative workspace for AI-assisted theoretical research. Github repo: \url{https://github.com/reaslab/ReasFlow.git}.
}

\newcommand{\eg}{\textit{e.g.}}
\newcommand{\paperone}{\cite{he2026fedslop}}
\newcommand{\papertwo}{\cite{zhu2026subspace}}
\newcommand{\paperthree}{\cite{zhang2026suda}}
\newcommand{\paperfour}{\cite{sun2026accelerated}}
\newcommand{\paperfive}{\cite{li2026retraction}}
\definecolor{humanbubble}{RGB}{235, 245, 255}
\definecolor{aibubble}{RGB}{245, 245, 245}
\definecolor{loggray}{RGB}{230, 230, 230}
\definecolor{indexbg}{RGB}{248, 248, 248}
\definecolor{reaslab}{RGB}{112, 48, 160}
\definecolor{reaslight}{RGB}{246, 243, 255}

\newtcolorbox{interactionlog}[1][]{
  enhanced,
  arc=0pt, outer arc=0pt,
  colback=white, colframe=reaslab,
  boxrule=0.8pt,
  fonttitle=\bfseries\sffamily, 
  coltitle=reaslab, colbacktitle=reaslight,
  title={\if\relax\detokenize{#1}\relax\else #1\fi},
  halign title=center, attach title to upper,
  after title={\vspace{4pt}\hrule\vspace{10pt}},
  lower separated=true,
  segmentation style={solid, reaslab, line width=0.8pt},
  colbacklower=reaslab,
}

\newcommand{\human}[1]{%
  \noindent\begin{flushright}
    \begin{minipage}[c]{0.65\textwidth}
      \begin{tcolorbox}[
        enhanced,
        colback=humanbubble, colframe=black!15,
        arc=6pt, sharp corners=southeast, boxrule=0.5pt,
        left=6pt, right=6pt, top=4pt, bottom=4pt, boxsep=0pt
      ]\small #1\end{tcolorbox}
    \end{minipage}%
    \hspace{8pt}
    \begin{minipage}[c]{30pt}
      \footnotesize\sffamily\textbf{User}
    \end{minipage}
  \end{flushright}
  \vspace{5pt}
}

\newcommand{\ai}[2]{%
  \noindent\begin{flushleft}
    \begin{minipage}[c]{85pt}
      \footnotesize\sffamily\textbf{#1}
    \end{minipage}%
    \hspace{2pt}
    \begin{minipage}[c]{0.65\textwidth}
      \begin{tcolorbox}[
        enhanced,
        colback=reaslight,
        colframe=reaslab!20,
        arc=6pt, sharp corners=southwest, boxrule=0.5pt,
        left=6pt, right=6pt, top=4pt, bottom=4pt, boxsep=0pt
      ]\small #2\end{tcolorbox}
    \end{minipage}
  \end{flushleft}\vspace{5pt}
}
\newtcolorbox{LLMBox}[2][]{
    colback=reaslight,
    colframe=reaslab,
    fonttitle=\bfseries,
    title=#2,               
    arc=2mm,
    boxrule=0.5pt,
    left=8pt,               
    right=8pt,
    top=8pt,
    bottom=8pt,
    breakable,
    #1                      
}
\newtcolorbox{NBLLMBox}[2][]{
    colback=reaslight,
    colframe=reaslab,
    fonttitle=\bfseries,
    title=#2,               
    arc=2mm,
    boxrule=0.5pt,
    left=8pt,               
    right=8pt,
    top=8pt,
    bottom=8pt,
    #1                      
}

\begin{document}

\maketitle

\section{Introduction}

\textbf{Motivations and Challenges.} Recent advances in Large Language Models (LLMs) have demonstrated remarkable capabilities in natural language generation, reasoning, and planning. Coupled with the continued expansion of context windows, these developments have spurred growing interest in autonomous AI agents capable of tackling highly complex tasks, spanning competitive mathematics to end-to-end scientific research. Beyond general problem-solving, AI-assisted research has already yielded breakthrough discoveries \cite{romera2024mathematical, bubeck2025early}, underscoring its transformative potential across specialized domains. Notably, several AI agent systems have achieved medal-level performance in international Olympiad competitions \cite{trinh2024solving, huang2025winning, qiu2025physics}, demonstrating strong capacity for rigorous logical reasoning and complex task execution. The open-source agent {Sci-Master}~\citep{chai2025scimaster} further showed that inference-time tool augmentation and multi-agent orchestration can attain state-of-the-art performance on Humanity's Last Exam~\cite{phan2025humanity}, establishing a robust reasoning foundation for scientific AI. Concurrently, fully autonomous research frameworks, including the AI Scientist \cite{lu2024ai}, CycleResearcher \cite{weng2024cycleresearcher}, and DeepScientist \cite{weng2025deepscientist}, have begun to automate entire scientific discovery pipelines.

\begin{figure}[t]
    \centering
    \includegraphics[width=0.8\linewidth]{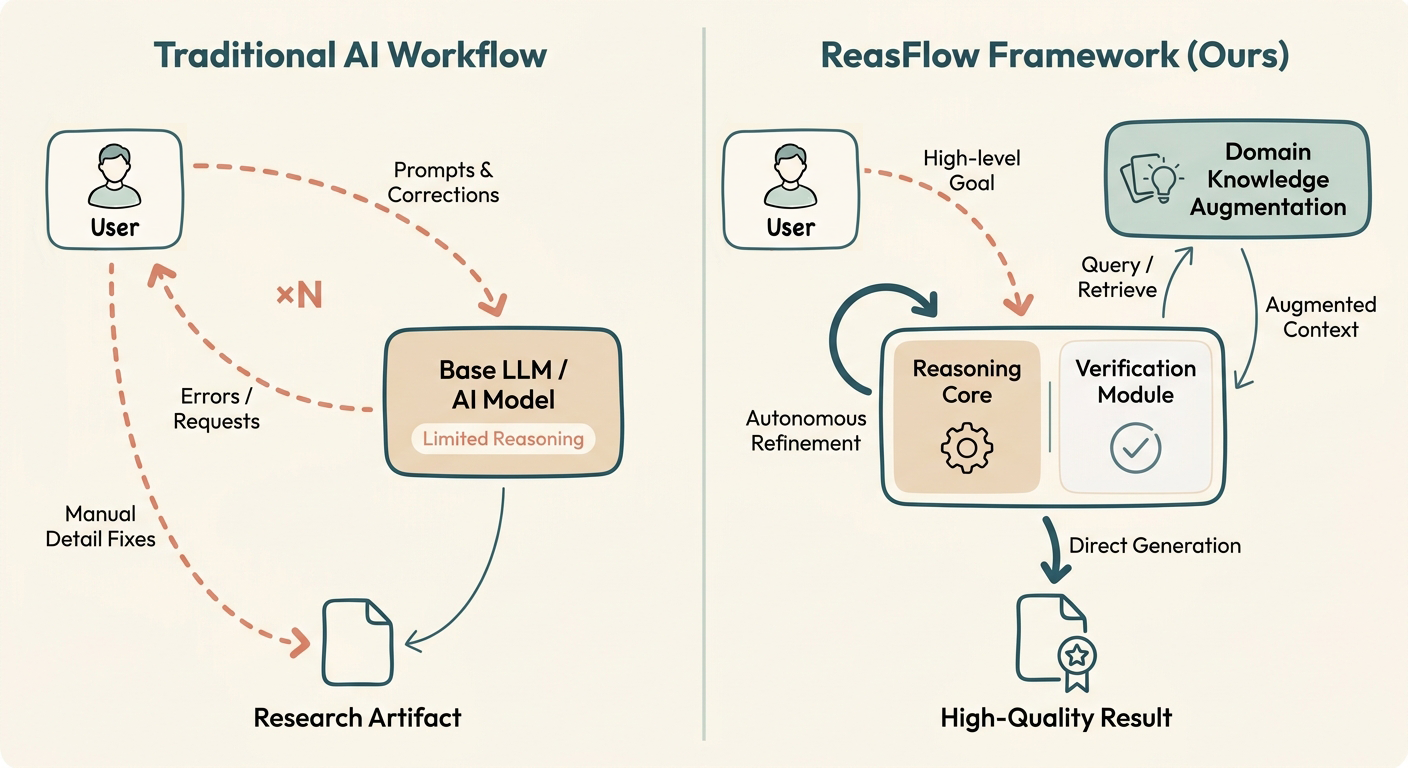}
    \caption{Illustration of traditional AI workflow with human guidance (left) and ReasFlow's workflow (right). ReasFlow uses automatic verfication loop and domain knowledge augmentation to reduce continuous human intervention.}
    \label{fig:intro_cmp}
\end{figure}

Despite these advances, the predominant focus of automated scientific discovery has remained on {empirically driven} research, domains such as machine learning \cite{yamada2025ai,weng2025deepscientist}, biology \cite{gottweis2025towards}, and physics \cite{miao2025physmaster}, where progress is gauged by quantitative benchmarks and experimental validation. In contrast, autonomous exploration of {theory-driven} research remains largely underexplored. In this work, we focus specifically on \textbf{applied mathematics}, a discipline that occupies a unique position in the research landscape. Unlike pure mathematics, which pursues abstraction for its own sake, applied mathematics demands tight integration between rigorous theorem proving and application-driven modeling in areas such as optimization, signal processing, statistical learning theory, and information theory. In contrast to empirical machine learning research, where progress is measured primarily by benchmark performance, applied mathematics centers on theoretical contributions, convergence guarantees, approximation bounds, and complexity characterizations, while relying on numerical experiments to validate tightness and demonstrate practical relevance. This ``theory-first, experiment-supporting'' paradigm distinguishes applied mathematics from both ends of the spectrum and renders existing automated research systems, which are designed either for competition-style problem solving or for benchmark-driven empirical workflows, inadequate for this domain.

Unlike mathematical competition problems, which are framed around well-specified questions with clearly defined targets, applied mathematical research is typically more open-ended: it requires synthesizing diverse domain knowledge across subfields (e.g., convex analysis, probability theory, and functional analysis may all be needed within a single proof), and it seeks broader conceptual insights accompanied by rigorous guarantees. For instance, analyzing the convergence rate of an optimization algorithm may involve constructing a suitable Lyapunov function, managing cascading inequality relaxations, and reasoning carefully about how different structural assumptions, such as strong convexity versus mere smoothness, affect the final bound. Existing automated research systems are not designed with such theoretical development in mind; the constituent agents in prior frameworks \cite{lu2024ai,weng2024cycleresearcher,weng2025deepscientist} lack specialized support for theorem proving, and the high-quality output they generate remains centered on empirical performance rather than foundational theory.

Developing AI-assisted systems for reasoning-centric research in applied mathematics presents several formidable challenges. {First, verifying theoretical reasoning is intrinsically difficult.} Unlike empirical research, where performance can be quantitatively benchmarked, proofs expressed in natural language lack reliable mechanisms for automated verification at scale. Beyond correctness alone, evaluating a theoretical contribution in applied mathematics also requires a multidimensional assessment: the tightness of derived bounds, the parsimony and reasonableness of assumptions (e.g., whether a Lipschitz continuity assumption is necessary or can be relaxed), and the broader significance of the result relative to existing theory. {Second, the reasoning capabilities of current LLMs remain insufficient for frontier applied mathematical research without substantial expert guidance.} Existing models are generally trained on broad, general-purpose corpora and lack systematic exposure to the specialized knowledge required for deep mastery of specific mathematical subfields. As a result, they cannot independently acquire the nuanced background knowledge, such as classical proof techniques in convex optimization, standard bounding arguments in high-dimensional statistics, or established conventions in approximation theory, that underpins theory-driven applied mathematical research. Consequently, frontier reasoning-centric exploration cannot proceed autonomously and remains critically dependent on human expertise and supervision. Sustained multi-turn interaction is therefore essential, as experts must iteratively provide domain-specific context, correct misaligned assumptions, and steer the reasoning process toward viable directions. This dependence on continuous human intervention limits the degree of end-to-end automation achievable in empirically driven workflows.

\textbf{Methodology.}  This paper proposes several designs to address these challenges. First, we introduce a robust internal verification loop into the agentic LLM system, enabling the agent to audit its own derivations before human inspection. This mechanism checks logical coherence, computational accuracy, and the validity of inequality relaxations, operations that are pervasive in applied mathematical proofs, such as bounding techniques, asymptotic analysis, and convergence rate derivations, while also detecting subtle issues such as implicit assumptions, loose bounds, and algebraic inconsistencies. By filtering out basic errors, it allows experts to focus on higher-level conceptual validation. Second, to reduce the need for continuous expert intervention, we replace live expert guidance with a domain-specific knowledge retrieval system. Rather than relying on experts to identify gaps and supply missing context, the agent actively retrieves the heuristics, proof techniques, and methodological conventions needed at each step of reasoning from a knowledge base. For example, when proving convergence of a first-order method, the agent can retrieve established potential function construction strategies or known descent lemma variants, rather than requiring an expert to suggest them in real time. This shifts the workflow from expert-driven correction to agent-driven knowledge acquisition, substantially reducing the researcher's cognitive burden. Finally, we emphasize that AI-assisted discovery remains inherently human-in-the-loop: the expert acts as a Principal Investigator, providing strategic direction and final judgment, while the AI serves as a capable Research Assistant, carrying out rigorous derivations and synthesizing domain-specific insights under supervision.

Constructing a knowledge base for reasoning-centric agents in applied mathematics is itself non-trivial. Manual curation is prohibitively expensive, and even the scientific literature contains subtle but consequential logical flaws, making it difficult to build a reliable gold-standard dataset. Moreover, much of the informal knowledge that drives theoretical discovery in applied mathematics, such as intuitions about when a particular relaxation will be tight, failed proof attempts that reveal structural obstacles, and heuristic shortcuts for constructing auxiliary functions, is rarely documented in published work, leaving open the question of how it should be formalized and quality-controlled for machine use. Broadly, the required knowledge falls into two categories: declarative domain knowledge, including explicit facts, definitions, and established theorems (e.g., classical convergence results, standard inequality chains, well-known counterexamples) that the agent can recognize as missing and actively retrieve; and procedural heuristics, which consist of practical reasoning know-how rather than explicit facts, including experience-based strategies (e.g., ``try a quadratic Lyapunov function first for strongly convex objectives''), proof insights (e.g., ``decouple the bias and variance terms before bounding each separately''), and cautionary patterns (e.g., ``this relaxation loses a logarithmic factor; check whether it matters for the final rate'') that the agent may fail to use because it does not recognize their relevance in the current reasoning context. Addressing this blind spot therefore requires mechanisms that can proactively surface relevant procedural knowledge without explicit prompts.

To circumvent these hurdles, we propose a systematic solution that automates the optimization of both knowledge extraction and retrieval procedures. The approach leverages the reasoning capabilities of LLMs to autonomously inspect and refine deficiencies in the extraction and retrieval pipelines, iterating until a measurable evaluation criterion is satisfied, for instance, until the agent's reasoning performance under internal knowledge retrieval matches that achieved under direct expert guidance. Importantly, this self-improvement loop also learns to associate subtle contextual cues with the need for procedural heuristics, enabling the system to retrieve and apply overlooked techniques even when the agent does not initially recognize their relevance. By grounding the process in an end-to-end performance metric rather than explicit manual verification of extracted knowledge, the framework avoids the intractable challenge of defining and evaluating informal knowledge representations, while ensuring that both declarative and procedural knowledge are effectively mobilized to support rigorous theoretical reasoning.

\textbf{Contributions.} Grounded in this perspective, we introduce \textbf{ReasFlow}, an end-to-end autonomous AI agent system for reasoning-centric scientific discovery in applied mathematics. ReasFlow implements a collaborative workflow in which the human researcher provides high-level ideas and supervisory oversight, while the agent independently carries out the labor-intensive stages of the research pipeline. The system includes dedicated modules for literature survey, algorithm design, theorem proving, experimental evaluation, manuscript writing, and self-review. By integrating these components, ReasFlow addresses a key limitation of prior automated research systems, which have focused primarily on empirically verifiable domains, and extends autonomous scientific discovery to the theory-driven paradigm characteristic of applied mathematics.

To validate ReasFlow, we used it to develop five theoretically significant algorithms and turn them into complete research papers in applied mathematics. Specifically, ReasFlow autonomously: (1) designed an accelerated distributed algorithm for stochastic strongly convex optimization that, for the first time, matches the lower complexity bound for this problem class \citep{sun2026accelerated}; (2) developed a low-rank communication-compressed algorithm for federated learning that achieves a faster theoretical convergence rate than prior human-designed methods \citep{zhu2026subspace}; (3) introduced a momentum-compatible extension that overcomes a key limitation of existing low-rank compression methods in federated learning \citep{he2026fedslop}; (4) proposed a distributed Muon optimizer for large-scale language model training that improves upon the best-known theoretical convergence rate of previous human-designed algorithms \citep{zhang2026suda}; and (5) developed the first retraction-free decentralized optimization algorithm over the Stiefel manifold \citep{li2026retraction}. All five works provide rigorous convergence guarantees and comprehensive empirical evaluations, collectively spanning distributed optimization, federated learning, manifold optimization, and LLM post-training. 

Remarkably, these advances emerged from minimal initial prompts and only a small set of core references, with ReasFlow autonomously conducting most of the algorithm design, theoretical analysis, experimentation, and manuscript preparation. We further benchmarked ReasFlow against several state-of-the-art open-source and publicly accessible AI research agents using identical initial prompts. A specialized LLM-based reviewer evaluated all outputs according to a carefully curated rubric. ReasFlow achieved the highest overall scores on theory-oriented research tasks and consistently received perfect scores from Gemini-3-Pro. These results demonstrate that ReasFlow can autonomously produce substantial theoretical advances and, in several cases, surpass the best results previously achieved by human experts in reasoning-intensive domains that have largely remained beyond the reach of automated research systems. Our contributions are summarized as follows:

\begin{itemize}
	\item \textbf{Systematic Enhancement of Reasoning Capabilities for Applied Mathematics.} We propose a principled framework for improving the theoretical rigor of AI research agents through integrated verification loops and automated domain-knowledge infusion, targeting the core bottlenecks of theory-driven discovery in applied mathematical research.

	\item \textbf{End-to-End Autonomous Research System.} We develop ReasFlow, an open-source AI assistant that unifies literature synthesis, algorithm design, theorem proving, experimentation, and manuscript preparation within a knowledge-based multi-agent system tailored to the applied mathematics research paradigm.

	\item \textbf{Comprehensive Benchmarking and Evaluation.} We introduce a multidimensional evaluation protocol to empirically compare ReasFlow with existing open-access AI research agents. The results show that ReasFlow consistently achieves superior generation quality and greater theoretical depth on applied mathematics research tasks.
\end{itemize}

ReasFlow is currently integrated into the \textbf{ReasLab} platform\footnote{\url{https://model.reaslab.io}}, where it is freely accessible to the public. Within ReasLab, each research project is associated with a dedicated {workspace} that stores all artifacts generated by ReasFlow, including literature summaries, algorithm drafts, proof sketches, experimental logs, and manuscript revisions. Users can inspect and edit these artifacts manually at any time, upload additional reference materials, and seamlessly resume collaboration with the agent. Interaction with ReasFlow resembles working with a research copilot: the user provides high-level research questions and periodic guidance through natural language dialogue, while the agent autonomously advances through literature survey, method development, theorem proving, experimentation, and manuscript writing, depositing its outputs directly into the shared workspace for review and iteration. Source codes, artifacts and review details for our experiments are available on Github: \url{https://github.com/reaslab/ReasFlow.git}.  A codex-oriented version is now available at \url{https://github.com/pkumelon/reasflow-codex.git}.

\begin{figure}[t]
    \centering
    \includegraphics[width=0.5\linewidth]{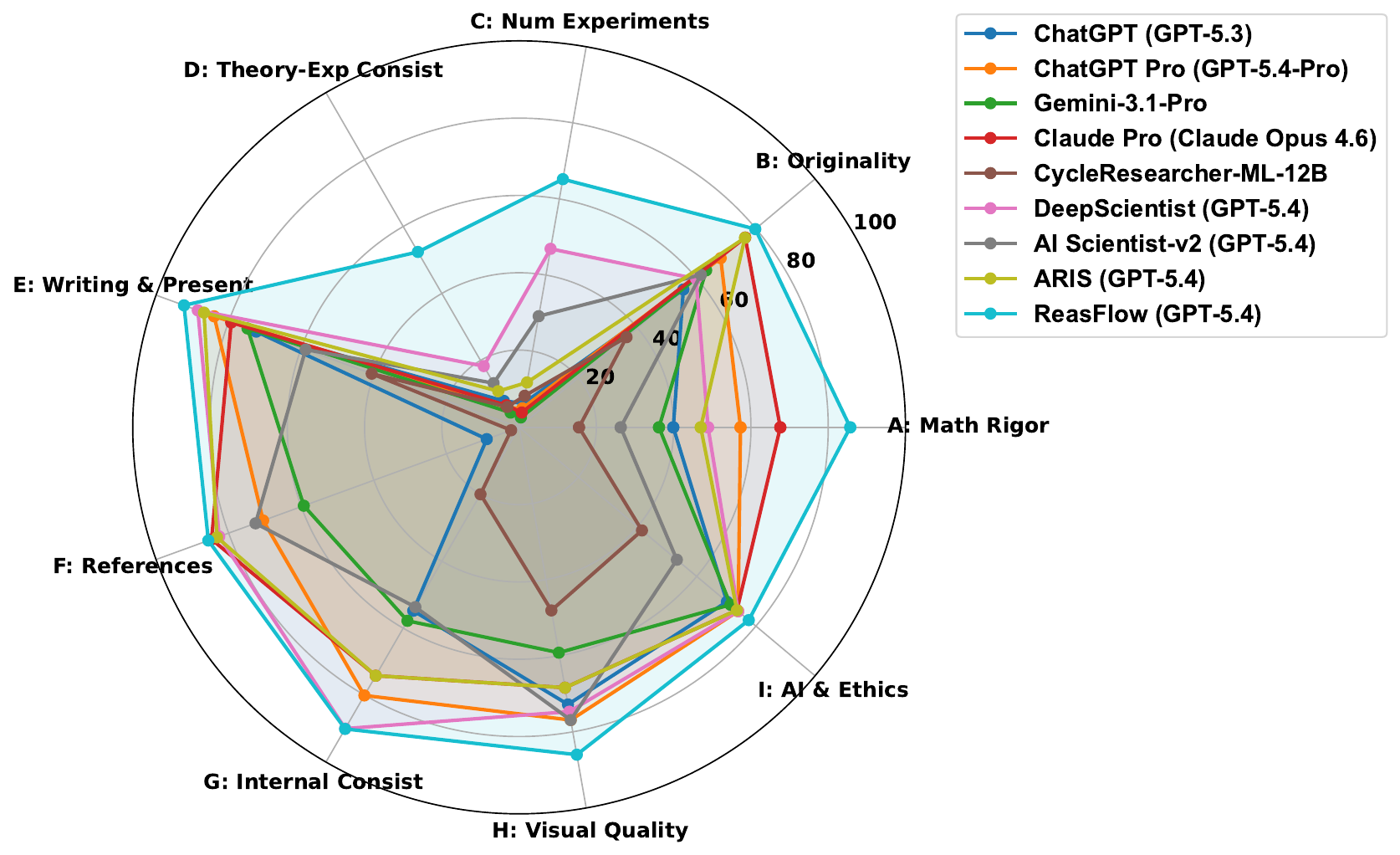}
    \caption{Radar chart comparing ReasFlow with baseline workflows across nine evaluation dimensions (A–I).}
    \label{fig:radar}
\end{figure}

\section{Related Work}

\noindent\textbf{AI-Assisted Scientific Research.}
The pursuit of autonomous scientific discovery has motivated the development of end-to-end agentic frameworks that seek to automate the full research lifecycle, from hypothesis generation to manuscript preparation. \citet{lu2024ai} introduced AI Scientist, a pioneering closed-loop system that combines automated experimentation with peer review; this framework was later extended in AI Scientist-v2 through agentic tree search \cite{yamada2025ai}. In parallel, several studies have focused on the iterative improvement of research quality. CycleResearcher \cite{weng2024cycleresearcher} implements an automated generation-and-review loop driven by iterative preference learning, DeepScientist \cite{weng2025deepscientist} targets the end-to-end discovery of state-of-the-art algorithms, and ML-Master \cite{liu2025ml} integrates MCTS-based exploration with steerable reasoning to automate machine learning engineering. Meanwhile, Agent Laboratory \cite{schmidgall2025agent} and AI co-scientist \cite{gottweis2025towards} further explore human-in-the-loop research workflows, emphasizing goal alignment and iterative feedback. Beyond these comprehensive frameworks, specialized agents have also been developed to address specific bottlenecks in the research process. For high-fidelity information retrieval and synthesis, PaperQA2 \cite{skarlinski2024language} and OpenScholar \cite{asai2024openscholar} achieve strong performance in citation-grounded question answering and literature synthesis, while AutoSurvey \cite{wang2024autosurvey} provides a systematic solution for large-scale automated survey generation. Further upstream in the research pipeline, Idea2Story \cite{xu2026idea2story} helps transform abstract ideas into more concrete and actionable research directions.

Despite this rapid progress, existing AI-assisted scientific research systems remain primarily oriented toward empirically driven research and often lack support for rigorous mathematical derivation and theorem proving. {ReasFlow} is designed to address this gap by enabling the autonomous generation of rigorous theoretical results and complex proofs, thereby extending AI-assisted discovery to theoretically demanding domains.

\noindent\textbf{AI-Assisted Mathematical Reasoning.}
Advances in large language models (LLMs) have extended automated reasoning from basic arithmetic to increasingly complex mathematical tasks. One prominent line of work focuses on competition-level challenges by augmenting models with heuristic search and test-time scaling, as exemplified by systems such as AlphaGeometry~\citep{trinh2024solving} and AlphaProof~\citep{hubert2025olympiad}, which achieved silver-medal performance at the International Mathematical Olympiad (IMO). Similar efforts including Kimina Prover~\cite{wang2025kimina}, DeepSeek-Prover-V2~\cite{ren2025deepseek}, BFS Prover~\cite{xin2025bfs}, BFS Prover V2~\cite{xin2025scaling}, Goedel Prover~\cite{lin2025goedel}, Goedel Prover-v2~\cite{lin2025goedelv2}, Seed Prover~\cite{chen2025seed}, Seed Prover 1.5~\cite{chen2025seedv1.5}, and LongCat-Flash-Prover~\cite{wang2026longcat} typically combine LLM reasoning with formal verification in Lean, and are actively advancing automated solving on benchmarks such as miniF2F~\citep{zheng2021minif2f} and PutnamBench~\cite{tsoukalas2024putnambench}. \citet{li2025sciagent} further extends to physics and chemistry Olympiads following a multi-agent paradigm. Similarly, M2F~\citep{wang2026m2f} demonstrates that LLMs can automatically formalize textbooks and research papers into buildable Lean projects.

In parallel, LLM-based systems have begun to tackle research-level problems, ranging from human-guided collaboration to fully autonomous verification. \citet{bubeck2025early} report several case studies in which \texttt{GPT-5}, under expert supervision, contributed key insights to open problems in combinatorics and online optimization. \citet{li2025advancing} proposed a human-in-the-loop proving workflow, illustrated through Riemannian optimization for Grover's quantum search. Pushing further toward autonomy, \citet{feng2026towards} introduced Aletheia, a natural-language agent that iteratively generates, verifies, and revises solutions, achieving strong results on the FirstProof~\cite{abouzaid2026first} benchmark and producing one fully autonomous research paper. While Aletheia operates entirely in natural language, \citet{ju2026automated} pursue a complementary direction by tightly coupling informal reasoning with formal verification in Lean 4, autonomously resolving a decade-old open problem in commutative algebra. Their system, Matlas~\cite{ju2026matlas}, searches over a knowledge base of  definitions and theorems, whereas ReasFlow is designed to leverage a richer body of domain knowledge that encompasses implicit insights and heuristic cues not easily captured from proofs. Beyond solving given problems, another line of work targets autonomous mathematical discovery: FunSearch~\citep{romera2024mathematical} and AlphaEvolve~\citep{novikov2025alphaevolve} demonstrate that LLM-guided search can produce verifiable new results and provably correct algorithms in an end-to-end manner.

A common feature of these systems is their focus on well-formulated problems with clearly specified assumptions and conclusions. Many scientific research tasks, however, are both open-ended and ill-defined: the appropriate hypotheses, the form of the desired result, and even the criteria for a meaningful contribution may not be known in advance. Addressing such tasks requires not only mathematical reasoning but also scientific judgment, particularly the ability to distinguish genuinely nontrivial structural insights from routine or trivial observations. ReasFlow is designed for this broader setting, aiming to autonomously formulate research questions and produce rigorous theoretical results without relying on pre-formalized problem statements.

\noindent\textbf{LLM-Based Automatic Evaluation and Review.}
The growing scalability and reasoning capabilities of LLMs have led to their increasing use as automated judges and reviewers in scientific workflows. Early foundational work by \citet{zheng2023judging} showed that strong proprietary models can function effectively as judges, exhibiting high correlation with human preferences in open-ended generation tasks. Building on this foundation, subsequent research has developed specialized systems for scholarly review that leverage LLMs' ability to provide structured critique. \citet{chu2024pre} proposed PRE, which aggregates judgments from multiple models to mitigate individual bias, while \citet{idahl2025openreviewer} and \citet{gao2025reviewagents} showed that fine-tuned agents or multi-agent ensembles can deliver diverse and professionally rigorous assessments of scientific manuscripts. In addition, systems such as those introduced by \citet{zhu2025deepreview} demonstrate that incorporating explicit literature retrieval and evidence-based argumentation can yield more reliable and verifiable review comments.

These advances strongly motivate the verification loops used in ReasFlow, suggesting that LLMs can serve as effective internal critics for iterative refinement. At the same time, ensuring fair and robust evaluation remains a key challenge. While \citet{ye2024justice} and \citet{li2025llms} identified task-specific distortions and prompt sensitivity, they also showed that these risks can be mitigated through careful system design. \citet{ye2024we} and \citet{lou2024aaar} further highlighted the value of structured benchmarks such as AAAR-1.0 and explicit evaluation rubrics in reducing favoritism and manipulation. Building on these insights, ReasFlow adopts a multi-criteria evaluation protocol based on standardized scientific rubrics to make its automated review process more impartial and better aligned with expert expectations. Taken together, these studies suggest that, with appropriate verification mechanisms, LLM-based judges can offer a scalable and credible alternative for high-stakes scientific evaluation.

\section{ReasFlow  System}\label{sec:reasflow_system}
We introduce \textbf{ReasFlow}, a multi-agent system tailored for reasoning-centric scientific research in applied mathematics, particularly in domains that require algorithmic design, theoretical convergence analysis, and numerical experimentation. ReasFlow encompasses the essential stages of the research process: (i) conducting literature surveys, (ii) designing algorithms, (iii) developing convergence proofs, (iv) performing numerical experiments, and (v) writing academic papers. Each stage is supported by specialized agents that incorporate domain-specific knowledge and can operate either autonomously or with human-in-the-loop interaction. With ReasFlow, researchers can conveniently translate their ideas into academic papers by providing simple descriptions and optionally related references. The system also supports flexible control and precise refinement based on user feedback, allowing adjustments when certain procedures are not sufficiently satisfactory. To enhance reasoning performance in specific research areas, users can further employ ReasFlow to extract knowledge cards that complement the built-in knowledge, for instance, in distributed optimization.

\begin{figure}[t]
	\centering
	\includegraphics[width=.8\linewidth]{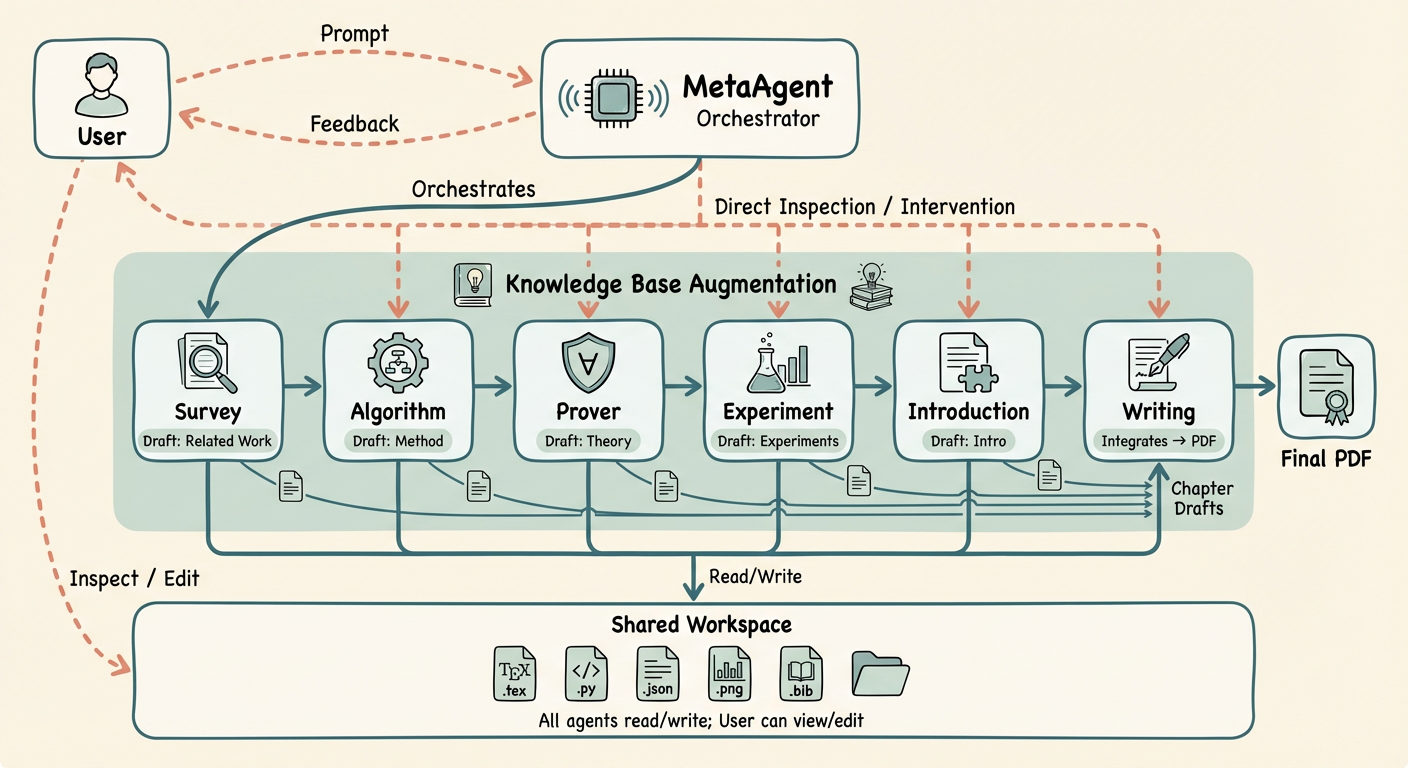}
	\caption{Illustration of the ReasFlow system.}
	\label{fig:reasflow}
\end{figure}

\subsection{System Overview}\label{subsec:overview}

ReasFlow is a multi-agent system that coordinates the full research pipeline for reasoning-centric scientific discovery in applied mathematics. As illustrated in Fig.~\ref{fig:reasflow}, the system is orchestrated by a central \texttt{MetaAgent}, which interacts with the user to clarify the research topic, generates a structured research plan, and sequentially invokes specialized sub-agents: \texttt{SurveyAgent}, \texttt{AlgorithmAgent}, \texttt{ProverAgent}, \texttt{ExperimentAgent}, \texttt{IntroductionAgent}, and \texttt{WritingAgent}.

Each agent is powered by a large language model and operates within a shared workspace, with access to basic utilities (\texttt{ReadFile}, \texttt{WriteFile}, \texttt{EditFile}, \texttt{RunTerminalCmd}) for file manipulation and command execution. {\color{black} Moreover, different agents are equipped with different specialized tools. To manage task complexity and context-length constraints, each agent has two auxiliary mechanisms:} \textbf{\texttt{skills}}, which return predefined prompts that impart specific in-context capabilities, and \textbf{\texttt{sub-agents}}, which allow an agent to delegate tasks to another specialized agent while maintaining isolated memory states. {\color{black} Taking the \texttt{ProverAgent} as an example, we provide it with a specialized tool to download arXiv papers for reference, a carefully designed skill to generate search queries, and sub-agents with clean contexts for mathematical proving and verfication.}

A key component of the system is a built-in repository of {knowledge cards} {\color{black} for applied mathematics}. This retrieval mechanism enables agents to draw upon domain-specific heuristics beyond their base training, and the system can further incorporate user-provided knowledge to extend its applicability, see Sec.~\ref{sec:knowledge_aug} for more detailed discussions. 

{\color{black} ReasFlow provides an automatic running scheme centered on MetaAgent as the orchestrator.} The \texttt{MetaAgent} workflow proceeds as follows. Upon user approval of the research plan, the \texttt{SurveyAgent} is invoked to conduct a literature survey and produce a draft of the \textit{Related Work} section. The \texttt{AlgorithmAgent} then generates candidate algorithms, writes pseudocode and a Python implementation, performs preliminary verification, and produces a report for the \textit{Method} section. Next, the \texttt{ProverAgent} establishes convergence theorems with full proofs, outputting a comprehensive LaTeX report. The \texttt{ExperimentAgent} subsequently designs and executes numerical experiments, including ablation studies, and generates a draft of the \textit{Experiment} section. Finally, the \texttt{IntroductionAgent} drafts the introduction, and the \texttt{WritingAgent} integrates all components into a polished manuscript. {\color{black} ReasFlow also supports an alternative scheme where users directly interact with \texttt{Sub-Agents} to achieve flexible execution.} Moreover, it enables a human-in-the-loop workflow in which domain experts can intervene at any stage, for example, collaborating with the \texttt{ProverAgent} to refine proof strategies or relax overly restrictive assumptions. 

Detailed tool specifications and default workflows for each agent are provided in Sec.~\ref{sec:agent_spec} and  Appendix~\ref{app:agent_specs}.

\begin{algorithm}[tb]
	\caption{Automatic Prompt Refinement for \texttt{CardGeneratorSkill}}
	\label{alg:CardGeneratorSkill-formal}
	\begin{algorithmic}[1]
		\STATE \textbf{Input:} Initial prompt \(s_0 \in \mathcal{S}\), {\color{black} reference paper set \(R \in \mathcal{R}\)}, proof task \(T \in \mathcal{T}\), iteration counter \(k \gets 0\)
		\STATE Compute baseline proof \(p_\star = \Phi_{\text{base}}(R, T)\)
		\WHILE{true}
		\STATE Extract knowledge cards \(C_k = \Psi(s_k, R)\)
		\STATE Generate current proof \(p_k = \Phi_{\text{cards}}(C_k, T)\)
		\IF{\(\Omega(p_k, p_\star) = \text{false}\)}
		\STATE Update prompt \(s_{k+1} = \Upsilon(s_k, R, C_k, p_k, p_\star)\)
		\STATE \(k \gets k + 1\)
		\ELSE
		\STATE \textbf{break}
		\ENDIF
		\ENDWHILE
	\end{algorithmic}
\end{algorithm}

\subsection{Knowledge Augmentation}\label{sec:knowledge_aug}
A central limitation of general-purpose large language models (LLMs) in reasoning-centric tasks is their lack of structured, domain-specific strategies. While models such as \texttt{GPT-5.1} or \texttt{Gemini-3-Pro} can attempt convergence proofs using basic techniques, they frequently struggle to identify the correct analytical pathways needed for tight theoretical results without expert guidance.

\noindent\textbf{Motivating Example: Stochastic Independence.}
We empirically observe that current LLMs often mishandle dependencies among stochastic variables. They may either apply lemmas requiring independence when the condition is unsatisfied, or overlook independence where it holds, leading to unnecessarily weak bounds. Consider independent stochastic vectors \(g_1, g_2, \dots, g_n\) with expectation \(g_\star \in \mathbb{R}^d\) and bounded variance \(\mathbb{E}[\|g_i - g_\star\|^2] \le \sigma^2\). A sharp bound can be derived via:
\[
	\mathbb{E}[\|\bar{g} - g_\star\|^2] = \sum_{i=1}^n \frac{1}{n^2} \mathbb{E}[\|g_i - g_\star\|^2] \le \frac{\sigma^2}{n},
\]
where \(\bar{g} := \frac{1}{n} \sum_{i=1}^n g_i\). In contrast, LLMs frequently default to applying Cauchy's inequality, yielding the looser bound:
\[
	\mathbb{E}[\|\bar{g} - g_\star\|^2] \le \frac{1}{n} \sum_{i=1}^n \mathbb{E}[\|g_i - g_\star\|^2] \le \sigma^2.
\]
While correct proofs can be elicited under direct human supervision, such guidance is impractical to provide for every proof step in an automated pipeline.

\noindent\textbf{Knowledge Cards.}
To address this, ReasFlow introduces {knowledge cards}, simple text files that encode targeted, contextual reasoning patterns and problem-solving heuristics. The content of a knowledge card varies by task. For the \texttt{ProverAgent}, cards emphasize general proof patterns and heuristics, focusing on \emph{how to prove} results for problems with specific structural characteristics. For the \texttt{SurveyAgent}, cards prioritize factual details and connections between {\color{black} literature}. We posit that a knowledge-augmented agent can achieve performance comparable to that of a non-augmented agent operating under direct, real-time expert guidance.

\begin{algorithm}[tb]
	\caption{Automatic Prompt Refinement for \texttt{CardRetrieverSkill}}
	\label{alg:CardRetrieverSkill-formal}
	\begin{algorithmic}[1]
		\STATE \textbf{Input:} Initial prompt \(s_0 \in \mathcal{S}\), proof task \(T \in \mathcal{T}\), ground-truth knowledge card {\color{black} set \(C_\star \in\mathcal{C}\)} for \(T\), iteration counter \(k \gets 0\)
		\STATE Compute baseline proof \(p_\star = \Phi_{\text{cards}}(C_\star, T)\)
		\WHILE{true}
		\STATE Generate query \(q_k = \Theta(T, s_k)\)
		\STATE Retrieve knowledge cards \(C_k = \Lambda(q_k)\)
		\STATE Generate current proof \(p_k = \Phi_{\text{cards}}(C_k, T)\)
		\IF{\(\Omega(p_k, p_\star) = \text{false}\)}
		\STATE Update prompt \(s_{k+1} = \tilde{\Upsilon}(s_k, C_k, C_\star, p_k, p_\star)\)
		\STATE \(k \gets k + 1\)
		\ELSE
		\STATE \textbf{break}
		\ENDIF
		\ENDWHILE
	\end{algorithmic}
\end{algorithm}

\noindent\textbf{Challenges in Knowledge Extraction and Retrieval.}
Effectively augmenting the \texttt{ProverAgent} with knowledge cards presents two primary challenges:
\begin{enumerate}
	\item \textbf{Quality control in knowledge extraction.} Proof techniques are often implicit. A poorly constructed card may describe its applicable scope too loosely, leading to inappropriate application, or too restrictively, limiting generalization. Thus, identifying which techniques to encode and how to describe them for correct generalization is non-trivial.
	\item \textbf{Effective retrieval of relevant cards.} Even with high-quality cards, retrieving the correct ones at the right time remains difficult. LLMs tend to be overconfident in their reasoning abilities. For a proof task requiring approximately five domain-specific cards, models often perform only one or two searches and may decline to read retrieved cards, judging them as trivial or relying solely on their abstracts. This behavior undermines the benefits of knowledge augmentation.
\end{enumerate}

\begin{figure}[htbp]
\centering
\begin{NBLLMBox}[width=\textwidth, fontupper=\small]{Example Knowledge Card}
		\textbf{Id:} TECH-DISTRIBUTED-VARIANCE-REDUCTION
        \vspace{5pt}
        
        \textbf{Type:} proof technique
        \vspace{5pt}
        
        \textbf{Domain:} optimization
        \vspace{5pt}
        
        \textbf{Title:} Distributed Variance Reduction (Linear Speedup)
        \vspace{5pt}
        
        \textbf{Description:} A fundamental technique for analyzing distributed/parallel stochastic algorithms. It leverages the independence of stochastic noise across N workers to reduce the variance of the aggregated gradient by a factor of 1/N, leading to linear speedup.
        \vspace{5pt}
        
        \textbf{Tags:} optimization, distributed-learning, federated-learning, variance-reduction, linear-speedup
        \vspace{5pt}
        
        \textbf{Problem Pattern:} 
        \begin{itemize}[nosep,leftmargin=*,label=-]
        \item "Analyzing convergence of distributed SGD or Federated Averaging."
        \item "Need to show that using N workers provides 1/N reduction in noise variance."
        \item "Standard Jensen's inequality bound (||mean||$^2$ <= mean(|| ||$^2$)) is too loose and loses the 1/N factor."
        \item "Aggregating independent zero-mean stochastic vectors."
        \end{itemize}
        \vspace{5pt}
        
\textbf{Structural Prerequisites:}
\begin{itemize}[nosep,leftmargin=*,label=-]
  \item "Noise vectors xi\_i must be zero-mean: E[xi\_i] = 0."
  \item "Noise vectors across workers must be independent: E[<xi\_i, xi\_j>] = 0 for i != j."
  \item "The aggregation operation is a simple average: 1/N sum(xi\_i)."
\end{itemize}
\vspace{5pt}

\textbf{Execution Logic:}
\begin{itemize}[nosep,leftmargin=*,label=-]
  \item step: "Decomposition into Mean and Noise"
  
    detail: "Decompose the stochastic update vector into its expected value (drift/signal) and zero-mean noise. Do NOT apply norm bounds to the whole sum immediately."
    
    algebraic prototype: "v\_i = bar\{v\} + xi\_i"

  \item step: "Orthogonality of Independent Noise"
  
    detail: "Expand the squared norm of the sum of noise vectors. Use the independence property to eliminate cross-terms E[<xi\_i, xi\_j>] = 0."
    
    dependency check: "Requires independence of noise across workers."
    
    algebraic prototype: "E[||sum xi\_i||$^2$] = sum E[||xi\_i||$^2$]"

  \item step: "Variance Reduction Scaling"
    detail: "Apply the 1/N$^2$ factor from the averaging operation to the sum of N variances. This results in a 1/N scaling factor for the final variance."
    
    algebraic\_prototype: "E[||1/N sum xi\_i||$^2$] = 1/N$^2$ * N * sigma$^2$ = sigma$^2$ / N"

  \item step: "Handling Bias/Drift Terms"
  
    detail: "For the bias/drift part (non-random or correlated), use standard inequalities (Jensen or Cauchy-Schwarz) which do not provide 1/N reduction. Combine with the reduced variance term."
    
    algebraic prototype: "E[||Update||$^2$] <= ||Drift||$^2$ + sigma$^2$/N"
\end{itemize}
\vspace{5pt}

\textbf{Assumptions Required:}
\begin{itemize}[nosep,leftmargin=*,label=-]
  \item "Independent Noise"
  \item "Unbiased Estimators (for the noise part)"
  \item "Bounded Variance"
\end{itemize}
\vspace{5pt}

\textbf{Expected Outcome Form:} "Variance term scales as O(sigma$^2$ / N) instead of O(sigma$^2$)."
	\end{NBLLMBox}
	\captionof{figure}{Example card extracted by refined skill prompts.}
	\label{fig:card-example}
\end{figure}

\noindent\textbf{Automatic Prompt Refinement.}
To overcome these challenges, we design an automatic refinement procedure for the skill prompts of \texttt{CardGeneratorSkill} and \texttt{CardRetrieverSkill}, analogous to optimizing model parameters. Formally, let \(\mathcal{S}\) be the set of all possible skill prompts, \(\mathcal{R}\) the set of reference paper sets, \(\mathcal{T}\) the space of target proof tasks, \(\mathcal{P}\) the space of generated proofs, and \(\mathcal{C}\) the space of knowledge card sets. Define:
\begin{itemize}
	\item \(\Phi_{\text{base}}: \mathcal{R} \times \mathcal{T} \to \mathcal{P}\): generates a baseline proof {\color{black} \(p_\star\in\mathcal{P}\)} using original reference paper set {\color{black}\(R\in\mathcal{R}\)}.
	\item \(\Phi_{\text{cards}}: \mathcal{C} \times \mathcal{T} \to \mathcal{P}\): generates a proof {\color{black}\(p\in\mathcal{P}\)} given knowledge cards {\color{black} \(C\in\mathcal{C}\)}.
	\item \(\Psi: \mathcal{S} \times \mathcal{R} \to \mathcal{C}\): extracts knowledge cards {\color{black} \(C\in\mathcal{C}\)} from {\color{black} \(R\subseteq\mathcal{R}\)} using prompt {\color{black} \(s\in\mathcal{S}\)}.
	\item \(\Omega: \mathcal{P} \times \mathcal{P} \to \{\text{true}, \text{false}\}\): compares whether {\color{black} \(p_k\in\mathcal{P}\)} is at least as good as {\color{black} \(p_\star\in\mathcal{P}\)}.
	\item \(\Upsilon: \mathcal{S} \times \mathcal{R} \times \mathcal{C} \times \mathcal{P} \times \mathcal{P} \to \mathcal{S}\): updates the prompt {\color{black} \(s_{k+1}\in\mathcal{S}\)} by analyzing deficiencies in {\color{black} \(p_k\in\mathcal{P}\)} relative to {\color{black} \(p_\star\in\mathcal{P}\)}.
\end{itemize}
Algorithm~\ref{alg:CardGeneratorSkill-formal} formalizes the iterative refinement of the \texttt{CardGeneratorSkill} prompt.

Similarly, we refine the \texttt{CardRetrieverSkill} prompt. Let \(\mathcal{Q}\) be the set of all knowledge-search queries. Define:
\begin{itemize}
	\item \(\Theta: \mathcal{S} \times \mathcal{T} \to \mathcal{Q}\): formulates a query {\color{black} \(q\in\mathcal{Q}\)} for task \(T\) using prompt {\color{black} \(s\in\mathcal{S}\)}.
	\item \(\Lambda: \mathcal{Q} \to \mathcal{C}\): returns knowledge cards retrieved in response to {\color{black} \(q\in\mathcal{Q}\)}.
	\item \(\tilde{\Upsilon}: \mathcal{S} \times \mathcal{C} \times \mathcal{C} \times \mathcal{P} \times \mathcal{P} \to \mathcal{S}\): updates the prompt by comparing retrieved card set {\color{black} \(C_k\in\mathcal{C}\)} and proof {\color{black} \(p_k\in\mathcal{P}\)} against ground-truth card set {\color{black} \(C_\star\in\mathcal{C}\)} and proof {\color{black} \(p_\star\in\mathcal{P}\)}.
\end{itemize}
Algorithm~\ref{alg:CardRetrieverSkill-formal} details this iterative process.

\noindent\textbf{Example Output.}
We present the extracted knowledge card corresponding to the motivating example in Fig.~\ref{fig:card-example}. As shown, the card accurately captures the logical structure underlying the proof technique and provides precise yet general descriptions of the scenarios in which it is applicable. The core information focuses on \emph{when} and \emph{where} the technique should be applied, thereby enabling the agent to retrieve this knowledge even when it is not explicitly aware of the specific technique itself.

\noindent\textbf{Remark.}
We empirically observe that after refinement of the skill prompts, the \texttt{ProverAgent} equipped with built-in knowledge cards achieves performance comparable to the agent with direct access to original reference papers. This outcome substantiates the efficacy of the proposed domain knowledge augmentation method in addressing two fundamental challenges inherent to constructing reasoning-oriented knowledge bases from literature. First, by synthesizing knowledge cards through LLM-driven comprehension of the source material rather than surface-level extraction, the approach effectively reveals the often implicit high-level insights that underpin formal proofs but remain tacit in the original exposition. Second, the workflow's reliance on the agent's autonomous verification loop during proof reconstruction (see Sec.~\ref{sec:prover_agent}) provides a practical safeguard against errors or specious claims that may be present in the published literature. More broadly, this finding highlights the potential of our method to enhance the reasoning capabilities of large language models in specialized domains by bridging the gap between informal scholarly discourse and rigorous, verifiable inference. 

\section{Agent Specifications}\label{sec:agent_spec}
In this section, we elaborate the design of all the sub-agents, including the workflow and knowledge augmentation approaches for the \texttt{SurveyAgent}, \texttt{AlgorithmAgent}, \texttt{ProverAgent}, \texttt{ExperimentAgent}, \texttt{IntroductionAgent} and \texttt{WritingAgent}.

\subsection{Survey Agent}\label{sec:survey_agent}

\begin{figure}[t]
	\centering
	\includegraphics[width=0.8\linewidth]{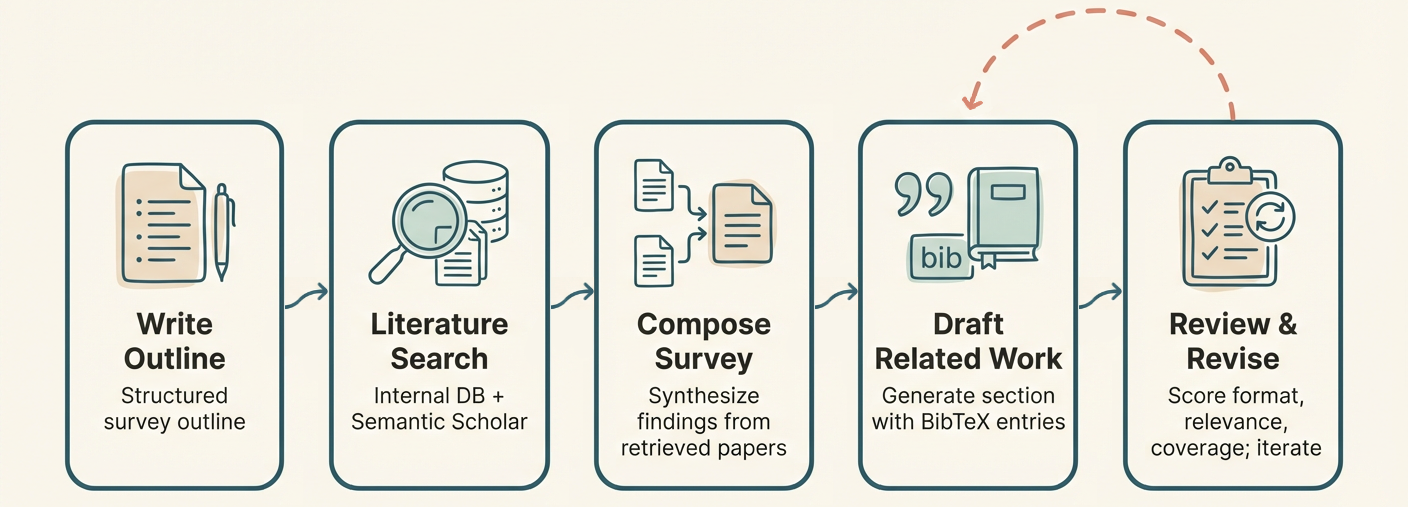}
	\caption{Workflow illustration of the \texttt{SurveyAgent}.}
	\label{fig:survey_cartoon}
\end{figure}

The \texttt{SurveyAgent} conducts comprehensive literature reviews on a given topic. By default, the \texttt{MetaAgent} invokes it first, as its task, identifying relevant references and drafting the \textit{Related Work} section, depends least on the outputs of other agents. The agent retrieves papers from both a built-in FAISS vector database, which indexes knowledge cards for papers in the target domain, and online sources via the Semantic Scholar Graph API. In addition, the agent can query individual paper metadata, citation networks, and reference lists through dedicated literature tools. A complete list of tools is provided in Appendix~\ref{app:agent_specs}.

\noindent\textbf{Workflow.} As illustrated in Fig.~\ref{fig:survey_cartoon}, the \texttt{SurveyAgent} proceeds through four stages upon receiving a user prompt:
\begin{enumerate}
	\item \textbf{Outline writing:} The agent identifies the survey topic and invokes \texttt{OutlineWriterTool} to produce a structured outline in the workspace.
	\item \textbf{Survey composition:} It calls \texttt{LiteratureSearchTool} to locate additional papers beyond the internal database, then uses \texttt{SurveyWriterTool} to generate a comprehensive survey report drawing from both sources.
	\item \textbf{Related work drafting:} The agent first activates \texttt{BibtexFormattingSkill} to enable standard bibtex writing, and then calls \texttt{RelatedWorkWriterTool} to produce a draft of \textit{Related Work} with correctly formatted citations.
	\item \textbf{Quality review:} The agent invokes \texttt{ReviewTool} to automatically evaluate the drafted \textit{Related Work} along four dimensions: format integrity (e.g., Bibtex field validation and cite-key consistency), citation relevance, key reference coverage, and instruction following. If the review score falls below the acceptance threshold, the agent iteratively revises and re-reviews the draft until quality requirements are met.
\end{enumerate}
Throughout this process, the user may inspect any generated files in the workspace to verify progress and intervene if necessary.

\noindent\textbf{Knowledge Augmentation.} To support effective literature reviews, we construct knowledge cards for each paper in the database and index them using a FAISS-based vector store with sentence-transformer embeddings. Each card encapsulates not only basic metadata (title, URL, and citation information) but also analytical insights such as strengths and weaknesses. The agent queries this database via \texttt{PaperLibrarySearchTool} to retrieve semantically relevant cards for the target topic. This structured representation enhances both the efficiency and quality of the survey process within the target domain, and improves the precision with which the agent describes individual papers when drafting the \textit{Related Work} section.

\subsection{Algorithm Agent}\label{sec:algorithm_agent}

\begin{figure}[t]
	\centering
	\includegraphics[width=0.8\linewidth]{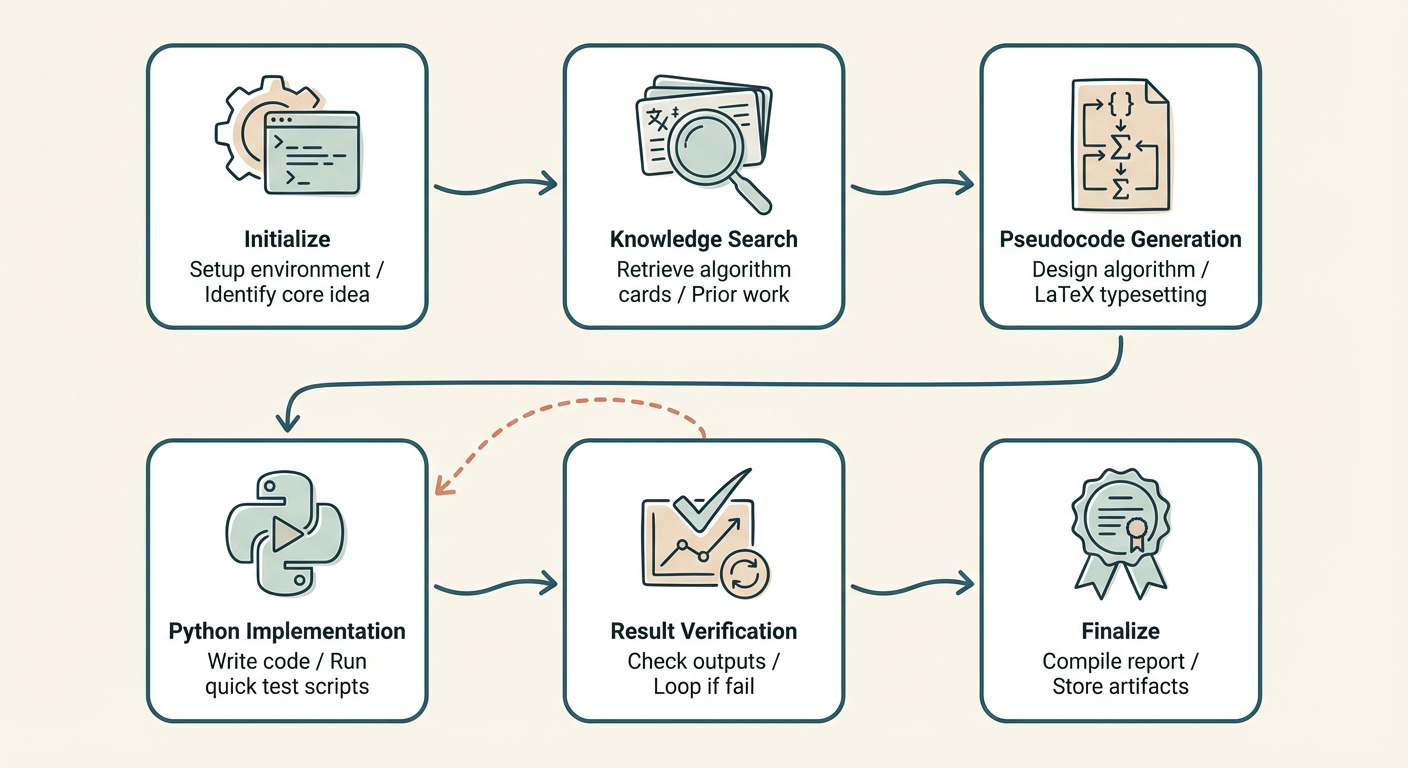}
	\caption{Workflow illustration of the \texttt{AlgorithmAgent}.}
	\label{fig:algorithm_cartoon}
\end{figure}

The \texttt{AlgorithmAgent} designs, implements, validates, and reports a new algorithm end-to-end based on a user prompt, which may be provided directly or generated by the \texttt{MetaAgent}. A complete list of tools is provided in Appendix~\ref{app:agent_specs}.

\noindent\textbf{Workflow.} As illustrated in Fig.~\ref{fig:algorithm_cartoon}, the agent proceeds through several stages:
\begin{enumerate}
	\item \textbf{Initialization:} The agent sets up the coding environment, identifies the core algorithmic idea, and prepares the workspace.
	\item \textbf{Knowledge search:} It invokes \texttt{KnowledgeCardSearch} to locate prior work and extract critical structural components, which inform the subsequent pseudocode design.
	\item \textbf{Pseudocode generation:} The agent performs algorithmic design and leverages the \texttt{LatexWriterAgent} to typeset formal pseudocode.
	\item \textbf{Python implementation:} It writes the algorithm and simple test scripts, then conducts quick experiments by executing the scripts and visualizing training dynamics via \texttt{SmartPlotTool}.
	\item \textbf{Result verification:} The agent checks experimental outputs against expected behavior using the \texttt{ImageReaderTool}. \textcolor{black}{The agent checks experimental outputs against expected behavior through the results in json file. } If validation fails, the process loops back for refinement; otherwise, it proceeds to finalization.
	\item \textbf{Finalization:} The agent compiles a final report, summarizing the algorithmic idea, pseudocode, implementation details, and experimental results, using the \texttt{LatexWriterAgent}. All artifacts are stored in the workspace.
\end{enumerate}
Throughout the process, the user may inspect progress artifacts and intervene if necessary.

\noindent\textbf{Knowledge Augmentation.} Knowledge cards for the \texttt{AlgorithmAgent} are constructed from research papers to capture key aspects of proposed algorithms, including pseudocode, advantages, and convergence properties. The agent consults these cards via \texttt{KnowledgeCardSearch} to access critical algorithmic design knowledge, accelerating principled design choices, guiding pseudocode synthesis, and providing reference criteria for automated verification.

\subsection{Prover Agent}\label{sec:prover_agent}

\begin{figure}[t]
	\centering
	\includegraphics[width=0.8\linewidth]{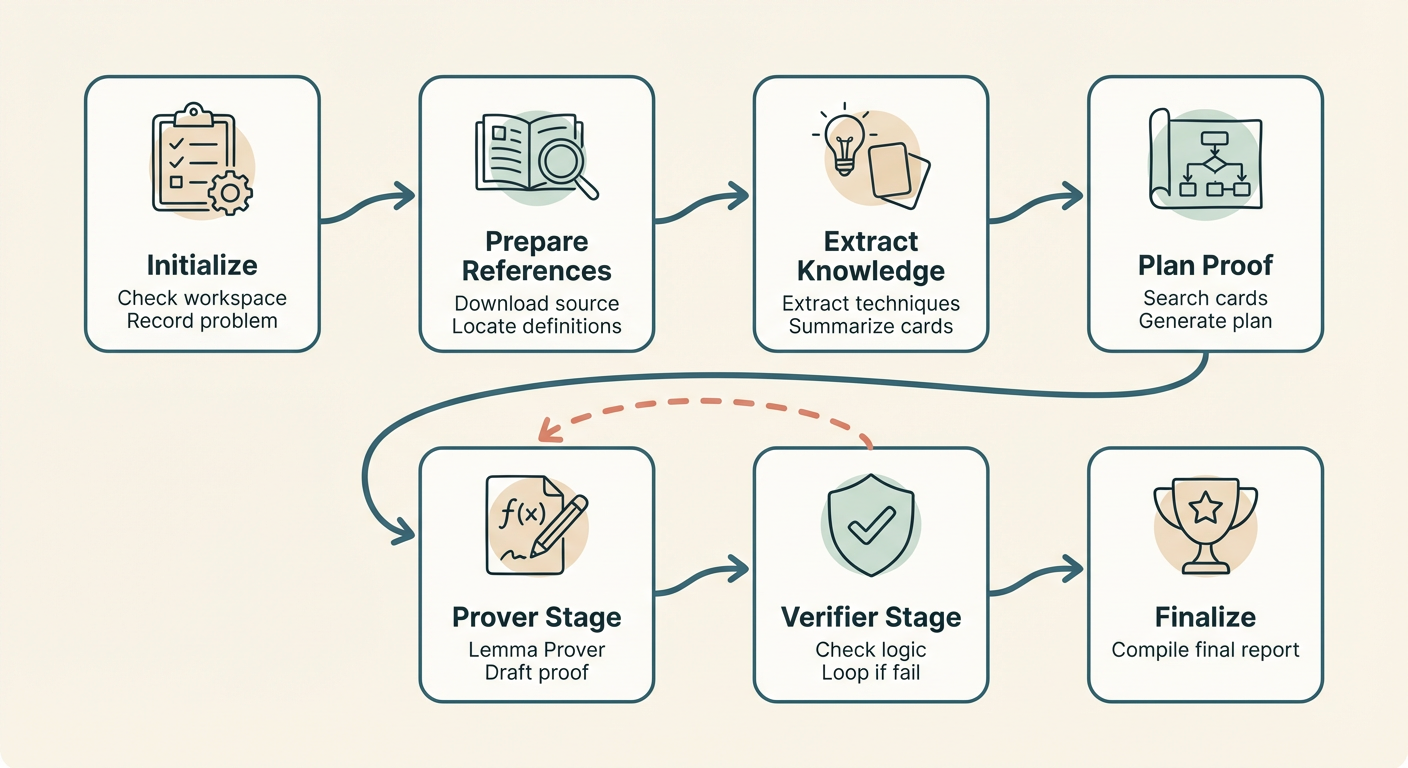}
	\caption{Workflow illustration of the \texttt{ProverAgent}.}
	\label{fig:prover_cartoon}
\end{figure}
\begin{figure}[t]
	\centering
	\includegraphics[width=0.8\linewidth]{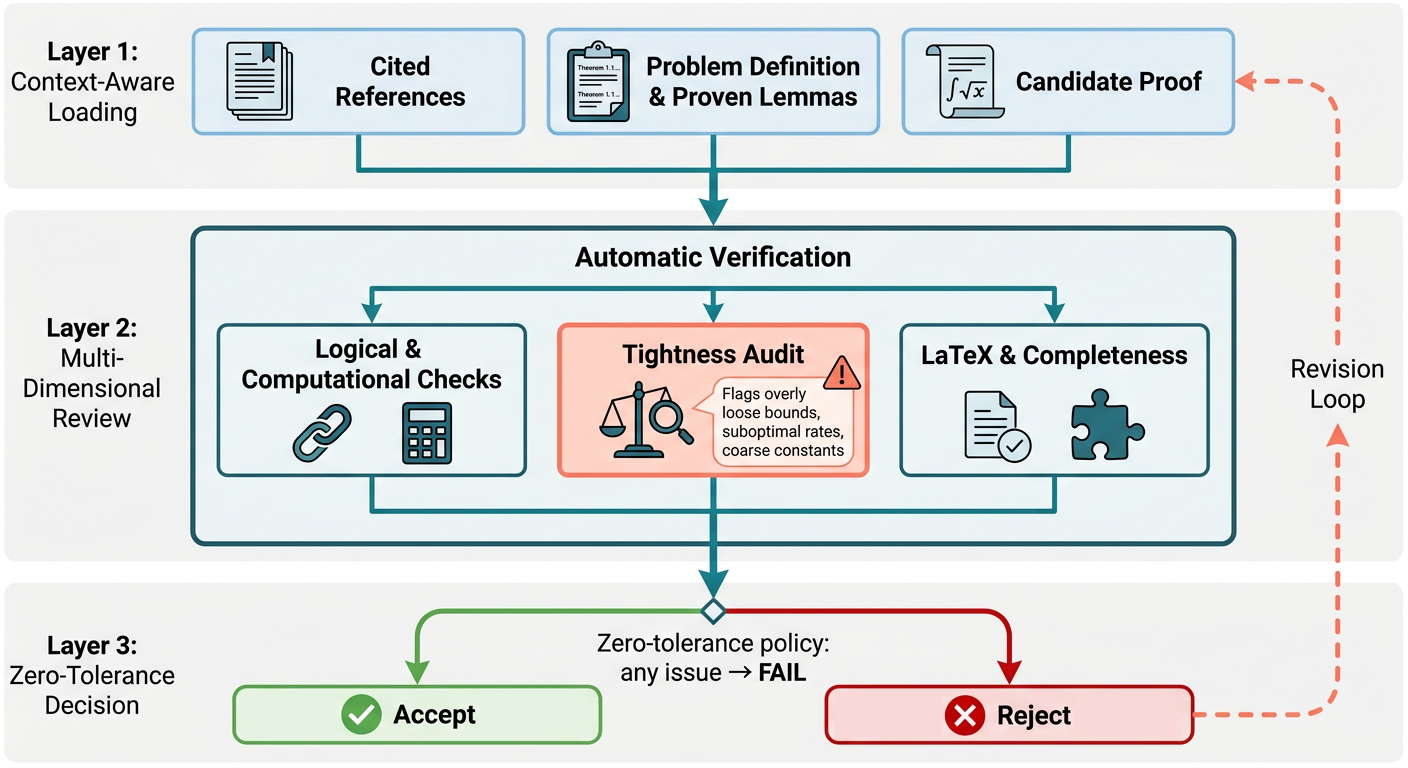}
	\caption{Illustration of the automatic verification procedure.}
	\label{fig:verifier_cartoon}
\end{figure}

The \texttt{ProverAgent} establishes theoretical convergence results for a given algorithm through mathematically rigorous proofs. The entire process is carried out via natural language reasoning; consequently, even a successfully generated theoretical report may still contain errors, and human evaluation remains necessary to ensure correctness. A complete list of tools is provided in Appendix~\ref{app:agent_specs}.

\noindent\textbf{Workflow.} As illustrated in Fig.~\ref{fig:prover_cartoon}, the agent proceeds through the following stages:
\begin{enumerate}
	\item \textbf{Initialization:} The agent checks workspace status, records the problem details, and prepares necessary files.
	\item \textbf{Reference preparation:} It uses \texttt{RefDownloaderSkill} to obtain required materials and \texttt{MathContentSearch} to locate relevant sections, definitions, and prior results.
	\item \textbf{Knowledge extraction:} The agent creates customized knowledge cards using \texttt{CardGeneratorSkill}, summarizing critical lemmas, assumptions, and proof strategies. This stage is primarily for domain knowledge extraction and may be optionally skipped.
	\item \textbf{Proof planning:} The agent searches for and reads relevant knowledge cards using \texttt{KnowledgeSearchTool} and \texttt{CardRetrieverSkill}, then generates a draft proof plan including both the general framework and detailed sub-steps.
	\item \textbf{Prover stage:} Guided by \texttt{LemmaProverSkill}, the agent leverages the \texttt{LemmaProverAgent} to generate draft proofs for individual lemmas or theorems. By default, only one statement is proved at a time to prevent error propagation.
	\item \textbf{Verifier stage:} The agent calls the \texttt{LemmaVerifierAgent} to check for logical soundness, gap-free reasoning, and adherence to the proof plan. If issues are detected, the process loops back to the prover stage for refinement.
	\item \textbf{Finalization:} Once all proofs are completed, the agent generates a comprehensive final report including the problem statement, assumptions used, and the statements and proofs of all lemmas, theorems, and corollaries.
\end{enumerate}
Throughout the process, the user may inspect progress artifacts, knowledge cards, proof plan, draft proofs, verification logs, and the final report, and intervene if necessary.

{\color{black}
\noindent\textbf{Automatic Verification. }Beyond basic syntax and consistency checks, the \texttt{LemmaVerifierAgent} is designed to address a critical pain point in automated mathematical reasoning: the subtle gap between a proof that looks correct and one that is truly rigorous. LLM-based proof generators often produce arguments that appear plausible but contain hidden flaws, \eg, unjustified inequality relaxations, tacitly assumed preconditions, or steps that skip essential justifications. As illustrated in Fig.~\ref{fig:verifier_cartoon}, The verifier tackles these challenges through three distinctive design choices.

First, it performs context-aware verification. Unlike a standalone checker, \texttt{LemmaVerifierAgent} loads not only the candidate proof but also the original problem description, previously proven lemmas, and any cited references. This allows it to detect mismatches such as invoking an unproven lemma, using symbols inconsistent with prior definitions, or assuming a property that has not been established.

Second, it explicitly audits tightness. In many proofs, a valid but overly loose bound still yields a correct statement qualitatively, yet renders the result non-publishable because the convergence rate or complexity becomes suboptimal. The verifier flags such “correct but weak” steps, \eg, replacing a fine-grained bound with a coarse universal constant where the problem demands sharp dependence on the number of clients or condition number.

Third, it adopts a zero-tolerance policy: any identified issue, regardless of perceived severity, results in a \texttt{FAIL}. This conservative criterion is motivated by the observation that in mathematical research, a single overlooked flaw, such as a missing intermediate inequality or an incorrectly computed coefficient, can invalidate an entire contribution. Consequently, the \texttt{LemmaVerifierAgent} rejects proofs that contain unstructured thinking traces (\eg, chain-of-thought remnants embedded in the proof file), ambiguous notation, or even a single step lacking explicit justification. The rejected proof is returned to the \texttt{LemmaProverAgent} with a detailed issue report, initiating a revision loop that continues until the proof meets the required standard of rigor. Empirical results from our case studies show that this strict verification loop catches common pitfalls, including cyclic dependencies, hidden assumptions, and mishandled boundary cases, that would otherwise remain undetected by surface-level checks.}

\noindent\textbf{Knowledge Augmentation.} Unlike other agents that rely primarily on factual information, the \texttt{ProverAgent} requires high-level insights that underpin the logical structure of proofs and the specific techniques used to complete them. These insights are often implicit and must be inferred rather than directly extracted. For instance, current LLMs frequently mishandle dependencies among stochastic variables, either misapplying independence lemmas or overlooking independence where it holds, leading to suboptimal bounds. Our approach extracts useful proof techniques from existing papers and encodes them as knowledge cards to enhance reasoning capabilities. The challenges of quality control in extraction and effective retrieval, along with the automatic prompt refinement procedures addressing them, were detailed in Sec.~\ref{sec:knowledge_aug}.

\subsection{Experiment Agent}\label{sec:experiment_agent}

\begin{figure}[t]
	\centering
	\includegraphics[width=0.8\linewidth]{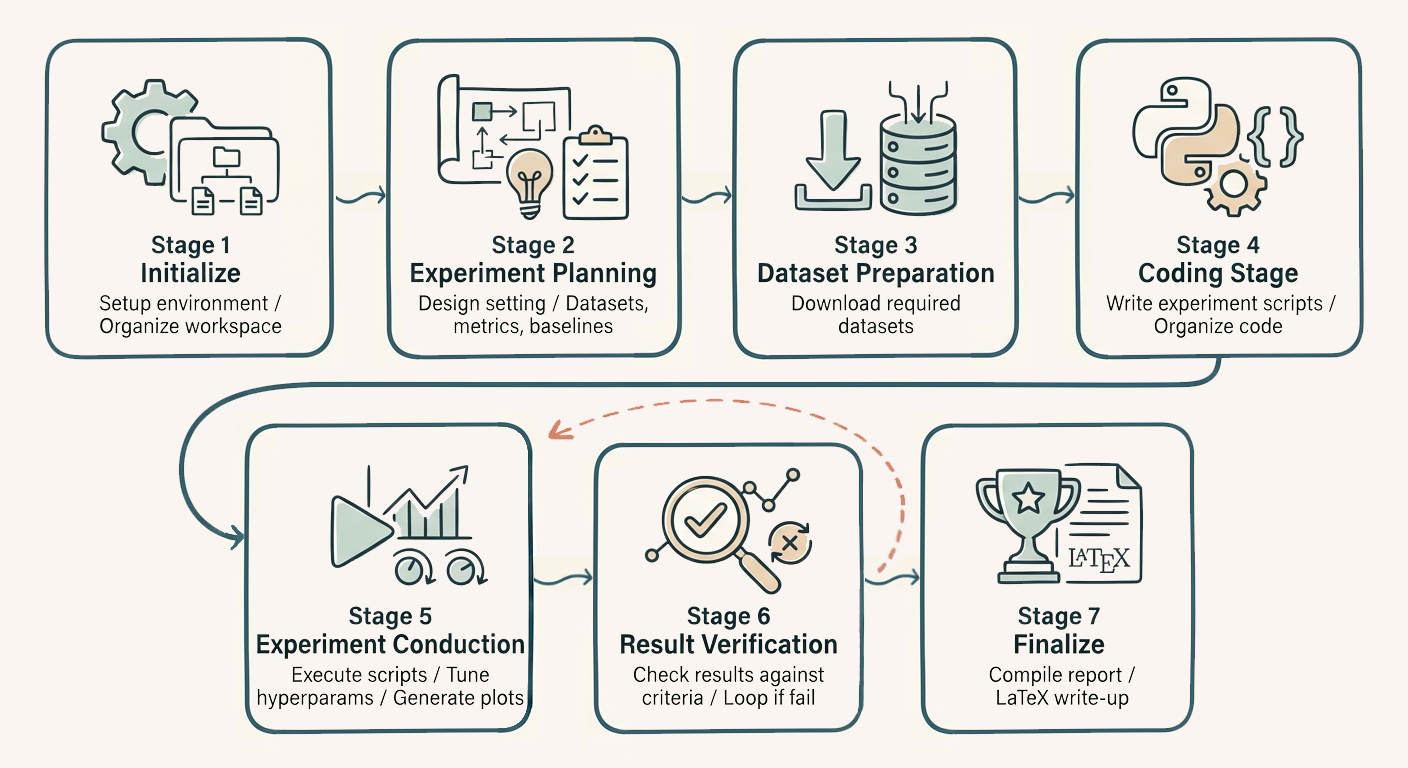}
	\caption{Workflow illustration of the \texttt{ExperimentAgent}.}
	\label{fig:experiment_cartoon}
\end{figure}

The \texttt{ExperimentAgent} designs, implements, conducts, and verifies computational experiments for a target algorithm, producing reproducible code, publication-ready figures, and a comprehensive final report that serves as the draft for the \textit{Experiment} section. A complete list of tools is provided in Appendix~\ref{app:agent_specs}.

\noindent\textbf{Workflow.} As illustrated in Fig.~\ref{fig:experiment_cartoon}, the agent proceeds through the following stages:
\begin{enumerate}
	\item \textbf{Initialization:} The agent sets up the coding environment and organizes the workspace.
	\item \textbf{Experiment planning:} It searches prior knowledge and designs the experimental setting, datasets, metrics, baselines, and hyperparameters, using \texttt{KnowledgeCardSearch}.
	\item \textbf{Dataset preparation:} The agent downloads required datasets via \texttt{TerminalCommandTool}.
	\item \textbf{Coding stage:} It writes runnable experiment scripts and organizes them in the workspace.
	\item \textbf{Experiment conduction:} The agent executes scripts, tunes hyperparameters with \texttt{AutoTuningTool}, and produces visualizations using \texttt{SmartPlotTool}.
	\item \textbf{Result verification:} It checks whether results meet expected criteria using \texttt{PlotAnalysisTool}. \textcolor{black}{It checks whether results meet expected criteria. } If issues arise, the process loops back for refinement.
	\item \textbf{Finalization:} The agent compiles a comprehensive report, experiment plan, datasets, configurations, results and figures, and key insights, using the \texttt{LatexWriterAgent}.
\end{enumerate}
Throughout, the user may inspect downloaded data, experiment codes, figures, verification logs, and the final report to intervene if needed.

\noindent\textbf{Knowledge Augmentation.} The \texttt{ExperimentAgent} leverages knowledge extracted from existing papers in two key aspects. First, it learns conventional experimental design choices from knowledge cards, enabling it to propose plans that align with academic standards, including target tasks, appropriate datasets, and commonly used baseline algorithms. Second, the pseudocode embedded in these cards allows the agent to efficiently implement baseline algorithms without downloading and analyzing the full original papers.

\subsection{Introduction Agent}\label{sec:introduction_agent}

\begin{figure}[t]
	\centering
	\includegraphics[width=0.8\linewidth]{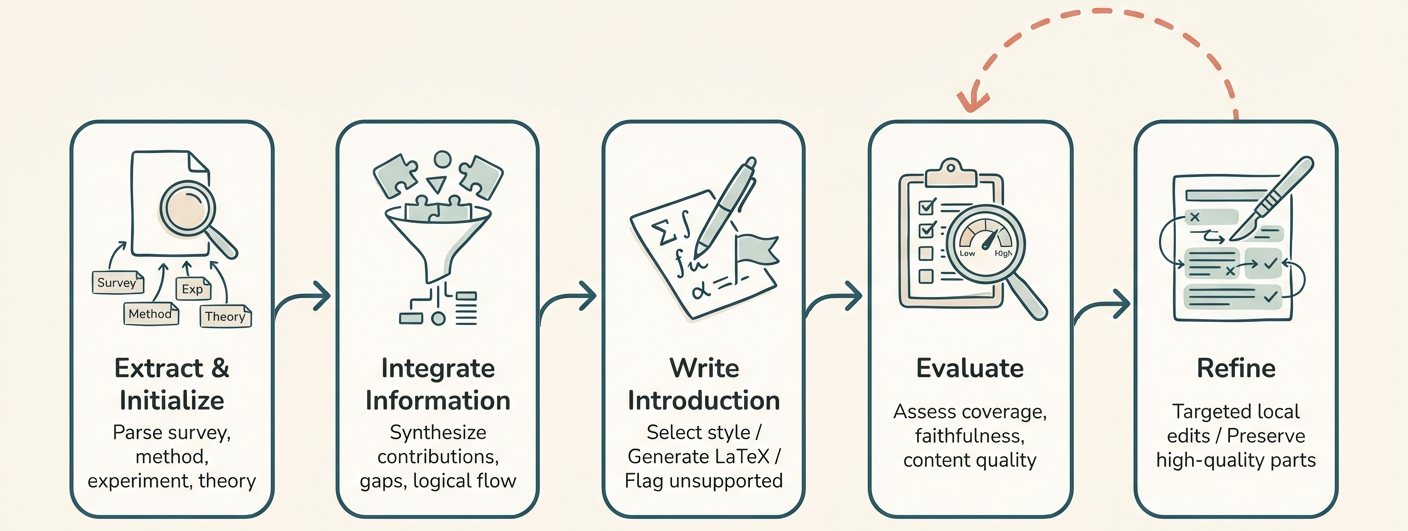}
	\caption{Workflow illustration of the \texttt{IntroductionAgent}.}
	\label{fig:introduction_cartoon}
\end{figure}

The \texttt{IntroductionAgent} synthesizes research progress and composes a comprehensive \textit{Introduction} section for the paper. By default, the \texttt{MetaAgent} invokes it prior to the \texttt{WritingAgent}, as the introduction requires outputs from both the \texttt{ProverAgent} and \texttt{ExperimentAgent} to accurately summarize contributions. To reliably extract structured information from prior agents' outputs, the \texttt{IntroductionAgent} is equipped with a suite of dedicated extraction tools that parse survey reports, method descriptions, experimental results, and theoretical contributions into structured summaries, preventing hallucination by grounding all content in source materials. Beyond generation, the agent incorporates an automated \emph{evaluation--refinement} loop: a dedicated evaluation tool assesses the generated introduction along three quality dimensions, and a refinement tool performs targeted local edits on low-scoring aspects while preserving high-quality content. A complete list of tools is provided in Appendix~\ref{app:agent_specs}.

\noindent\textbf{Workflow.} As illustrated in Fig.~\ref{fig:introduction_cartoon}, the agent proceeds through the following stages:
\begin{enumerate}
	\item \textbf{Initialization:} The agent checks workspace status, reads progress documents generated by previous agents, and invokes extraction tools, \texttt{ExtractSurveyInfo}, \texttt{ExtractMethodInfo}, \texttt{ExtractExperimentInfo}, and \texttt{ExtractTheoryInfo}, to parse each document into structured summaries of background, methods, results, and theoretical contributions.
	\item \textbf{Information integration:} It calls \texttt{OrganizeExtractedInfo} to synthesize the collected materials, identifying key contributions, research gaps, and the logical flow that motivates the work.
	\item \textbf{Introduction writing:} The agent determines an appropriate writing style based on the research topic, choosing among an \textit{ML style} (hook$\to$background$\to$gap$\to$contribution), a \textit{Math style} (hook$\to$objective$\to$ background$\to$related work$\to$gap$\to$contribution$\to$notation), or a \textit{Default style} (hook$\to$background$\to$survey $\to$gap$\to$contributions), and invokes \texttt{IntroductionWriterTool} to generate the LaTeX draft with inline hallucination annotations that flag any claims not directly supported by the extracted materials. The agent optionally supplements references using \texttt{LiteratureSearchTool} (Google Scholar as primary, Semantic Scholar as fallback) and \texttt{BibtexFormattingSkill}.
	\item \textbf{Evaluation:} After writing, the agent invokes \texttt{EvalIntroduction} to assess the generated introduction via a three-stage pipeline. First, it performs a structured decomposition of the reference \texttt{.tex} files (experiment sections, proof sections, related work sections) into categorized claims with importance labels (\texttt{critical}, \texttt{important}, \texttt{supplementary}). Second, it extracts all contribution and factual statements from the generated introduction as prediction claims. Third, it evaluates three quality dimensions via LLM-as-a-judge: \emph{coverage} (structure-aware weighted coverage of reference content), \emph{faithfulness} (fraction of claims supported by source materials), and \emph{content quality} (six sub-dimensions assessing domain clarity, algorithmic detail, background completeness, core idea prominence, goal specificity, and substantiveness). The evaluation produces a structured JSON report with per-dimension scores and an overall score, along with a three-tier threshold verdict (\texttt{auto\_refine} below 0.60, \texttt{suggest\_refine} below 0.75, \texttt{good} otherwise).
	\item \textbf{Refinement:} Based on the evaluation report, the agent invokes \texttt{RefineIntroduction} to perform targeted local edits on low-scoring dimensions. The tool first generates a structured refinement plan (identifying specific paragraphs and issues), then executes local rewrites guided by six hard constraints: no unsupported technical details, no alteration of contribution semantics, no fabricated numerical results, no forged citations, no structural reorganization, and consistent citation formatting. The original file is automatically backed up, and a line-level diff summary is produced for user review. This evaluation--refinement cycle can be iterated until quality thresholds are met.
\end{enumerate}

\noindent\textbf{Knowledge Augmentation.} Similar to the \texttt{SurveyAgent}, the \texttt{IntroductionAgent} is augmented with a built-in database of knowledge cards that encapsulate materials helpful for accurately describing each paper's contribution, including its strengths and weaknesses. This structured knowledge enables the agent to provide fair and precise descriptions when citing existing work.

\subsection{Writing Agent}\label{sec:writing_agent}

\begin{figure}[t]
	\centering
	\includegraphics[width=0.8\linewidth]{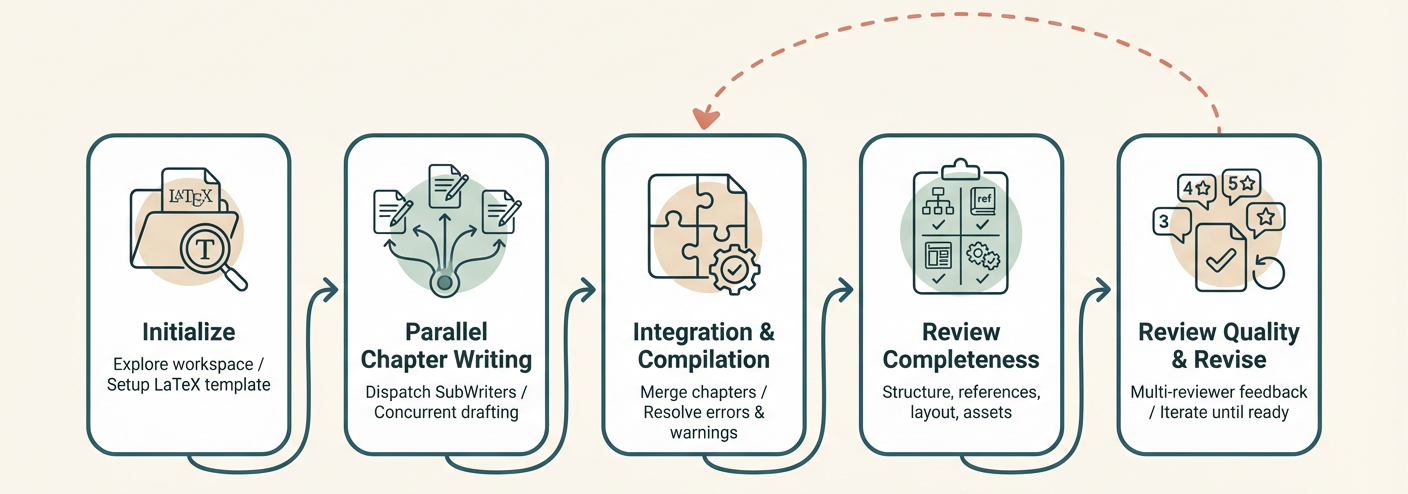}
	\caption{Workflow illustration of the \texttt{WritingAgent}.}
	\label{fig:writing_cartoon}
\end{figure}

The \texttt{WritingAgent} assembles and polishes the draft materials generated by preceding agents into a complete, publication-ready academic paper. It adopts a two-tier architecture: a \texttt{Coordinator} that plans the paper structure, dispatches chapter-level tasks, and integrates the final manuscript; and multiple \texttt{SubWriter} instances that work on individual chapters in parallel. A complete list of tools is provided in Appendix~\ref{app:agent_specs}.

\noindent\textbf{Workflow.} As illustrated in Fig.~\ref{fig:writing_cartoon}, the default workflow proceeds as follows:
\begin{enumerate}
	\item \textbf{Initialization:} The \texttt{Coordinator} explores the workspace, reads the asset inventory, and copies the venue-specific LaTeX template suite (document class, style files, and a pre-configured \texttt{main.tex} with conditional-inclusion guards) into the output directory.
	\item \textbf{Parallel chapter writing:} The \texttt{Coordinator} plans the chapter structure, assigns specific asset files to each chapter, and dispatches writing tasks to \texttt{SubWriter} instances via \texttt{DispatchTask}. Each \texttt{SubWriter} independently collects and merges its assigned materials, writes the chapter \texttt{.tex} file, compiles, and visually verifies the output. Multiple \texttt{SubWriter}s run concurrently; as each chapter completes, the Coordinator incrementally performs cross-chapter coherence editing (adding cross-references and transitions).
	\item \textbf{Integration and compilation:} The \texttt{Coordinator} integrates all chapters into \texttt{main.tex}, merges bibliography files, and compiles the full paper, iteratively fixing all errors and warnings until a clean build is achieved.
	\item \textbf{Review and revision:} The \texttt{Coordinator} invokes \texttt{ReviewCompleteness} to identify structural issues and missing content, addresses every reported item, then invokes \texttt{ReviewWritingQuality} to obtain multi-reviewer feedback and revises accordingly. A final page-by-page visual inspection with \texttt{ScreenshotAnalysisTool} ensures correct layout. Steps~3--4 are repeated until the paper is ready for publication.
\end{enumerate}

{\color{black} \noindent\textbf{Knowledge Augmentation.} To support coherent and format-compliant writing, we construct modular knowledge documents for the \texttt{WritingAgent}, including an asset-handling guide that specifies how to incorporate various file types (e.g., \texttt{.tex} proofs, \texttt{.bib} references, images, and experiment reports), venue-specific formatting rules for different publication outlets, and chapter-compilation scripts for assembling multi-section manuscripts. Users can further extend this knowledge base by adding custom skills. The agent accesses these documents through skills, enabling it to apply correct formatting, manage cross-references and assets, and structure multi-chapter papers consistently. This knowledge augmentation significantly improves the adherence to target venue requirements and reduces manual intervention in the document assembly process.}

\section{Experiment}
The primary goal of ReasFlow is to assist researchers in autonomously exploring promising research ideas and producing academic papers, with a broad focus on reasoning-intensive problems in applied mathematics. To assess its effectiveness across diverse applied mathematics settings, we use ReasFlow to generate five high-quality research papers spanning federated learning, decentralized signal processing, manifold optimization, and LLMs fine-tuning. These areas encompass a wide range of theoretical and computational challenges arising in statistical learning, networked systems, numerical optimization, and large-scale machine learning. We then evaluate ReasFlow's overall performance by comparing it with several publicly available AI research systems, including AI Scientist-v2~\cite{yamada2025ai}, Cycle Researcher~\cite{weng2024cycleresearcher}, DeepScientist~\cite{weng2025deepscientist}, and ARIS~\cite{yang2026aris}. Finally, we assess the capabilities of ReasFlow's individual sub-agents by benchmarking each of them against the corresponding baselines.

\subsection{Case Studies: End-to-End Discovery}\label{sec:case_studies}

\begin{figure}[t]
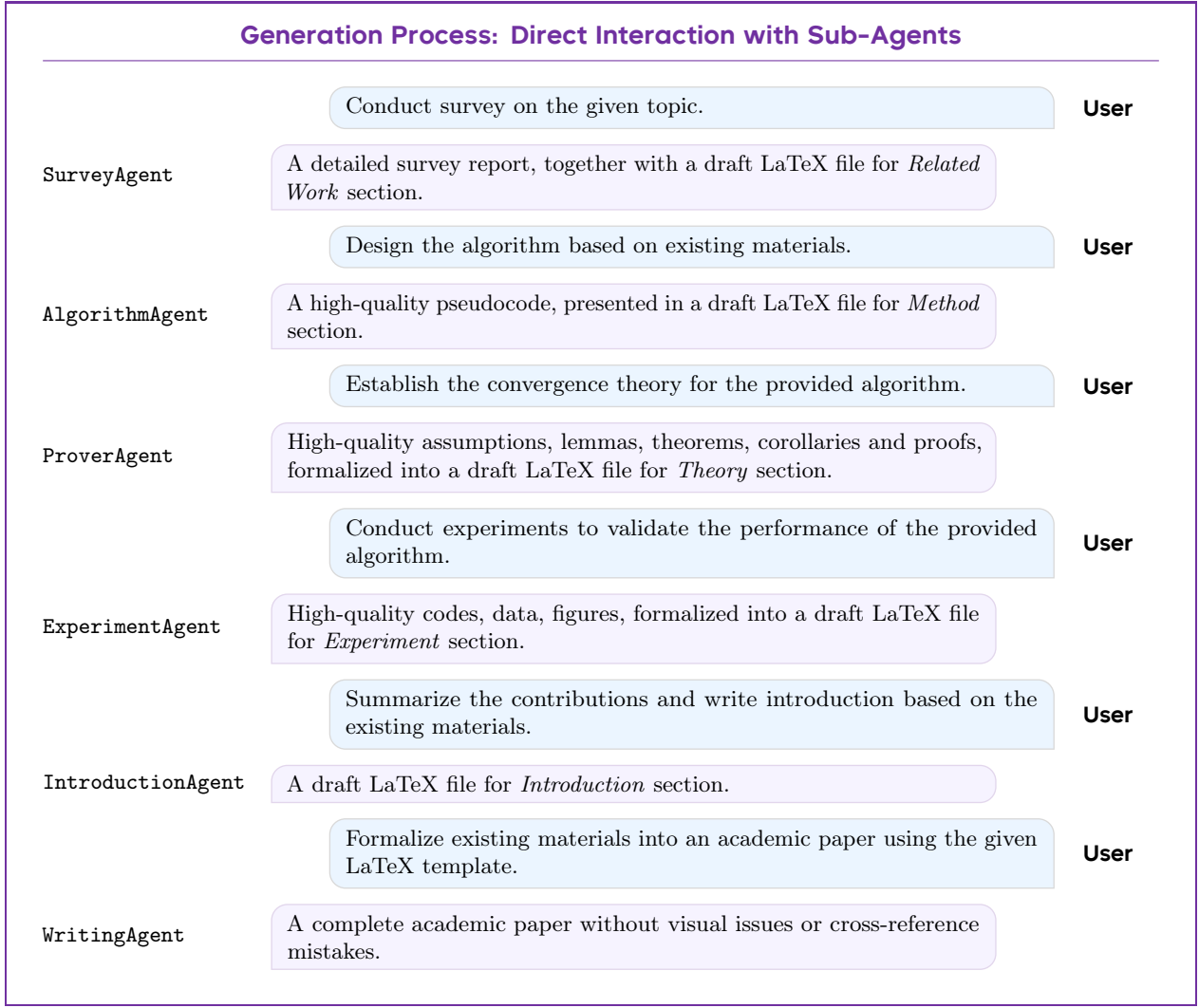

	\centering
	\begin{interactionlog}[Generation Process: Direct Interaction with Sub-Agents]

		\human{Conduct survey on the given topic.}

		\ai{\texttt{SurveyAgent}}{A detailed survey report, together with a draft LaTeX file for \textit{Related Work} section.}

		\human{Design the algorithm based on existing materials.}

		\ai{\texttt{AlgorithmAgent}}{A high-quality pseudocode, presented in a draft LaTeX file for \textit{Method} section.}

		\human{Establish the convergence theory for the provided algorithm.}

		\ai{\texttt{ProverAgent}}{High-quality assumptions, lemmas, theorems, corollaries and proofs, formalized into a draft LaTeX file for \textit{Theory} section.}

		\human{Conduct experiments to validate the performance of the provided algorithm.}

		\ai{\texttt{ExperimentAgent}}{High-quality codes, data, figures, formalized into a draft LaTeX file for \textit{Experiment} section.}

		\human{Summarize the contributions and write introduction based on the existing materials.}

		\ai{\texttt{IntroductionAgent}}{A draft LaTeX file for \textit{Introduction} section.}

		\human{Formalize existing materials into an academic paper using the given LaTeX template.}

		\ai{\texttt{WritingAgent}}{A complete academic paper without visual issues or cross-reference mistakes.}

	\end{interactionlog}
	\caption{The process of generating high-quality research papers by directly interacting with ReasFlow's sub-agents. Each sub-agent is not limited to a single trial. Refinements are allowed based on the user feedback.}
	\label{fig:generation-process-direct}
\end{figure}

\begin{figure}[t]
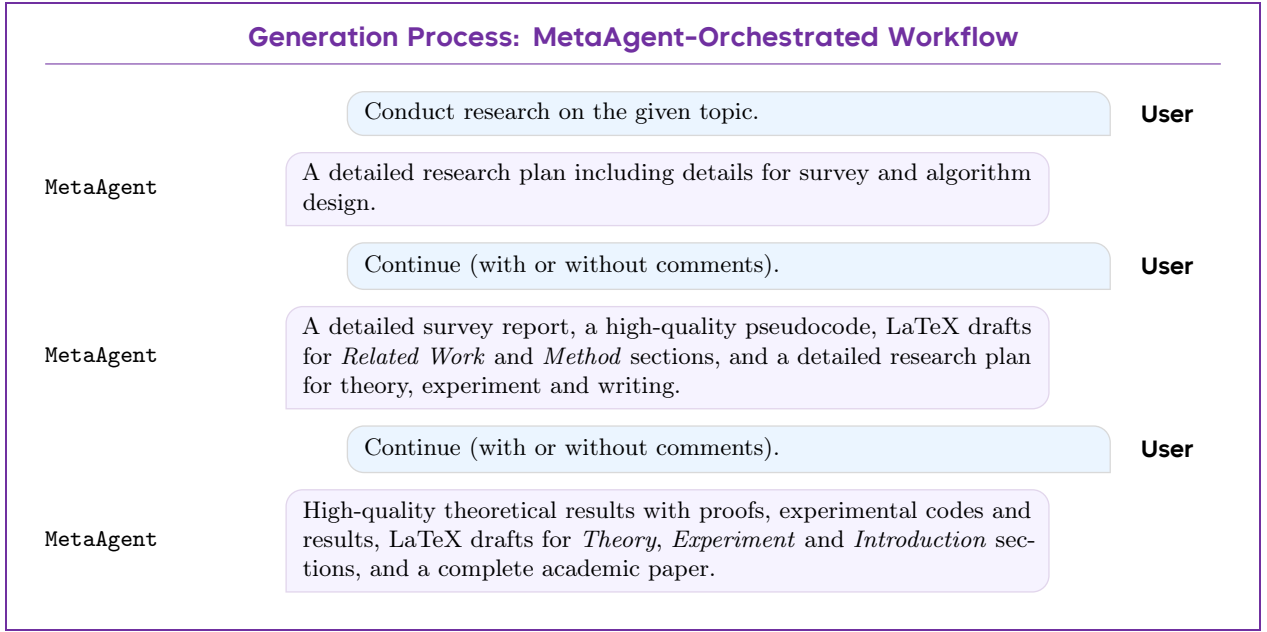

	\centering
	\begin{interactionlog}[Generation Process: MetaAgent-Orchestrated Workflow]

		\human{Conduct research on the given topic.}

		\ai{\texttt{MetaAgent}}{A detailed research plan including details for survey and algorithm design.}

		\human{Continue (with or without comments).}

		\ai{\texttt{MetaAgent}}{A detailed survey report, a high-quality pseudocode, LaTeX drafts for \textit{Related Work} and \textit{Method} sections, and a detailed research plan for theory, experiment and writing.}

		\human{Continue (with or without comments).}

		\ai{\texttt{MetaAgent}}{High-quality theoretical results with proofs, experimental codes and results, LaTeX drafts for \textit{Theory}, \textit{Experiment} and \textit{Introduction} sections, and a complete academic paper.}

	\end{interactionlog}
	\caption{The process of generating high-quality research papers by interacting with the \texttt{MetaAgent}. Refinements are allowed based on the user feedback.}
	\label{fig:generation-process-indirect}
\end{figure}

To assess ReasFlow's capacity for autonomous research across diverse areas of applied mathematics, we conducted five case studies spanning federated learning, decentralized signal processing, manifold optimization, and LLM fine-tunings under two distinct interaction paradigms. In the first three cases, human experts interacted directly with individual specialized sub-agents, providing each with a concise research idea and a small set of key references. Under this direct-prompting workflow, the sub-agents carried out their respective stages of the research pipeline, including literature review, algorithm design, theoretical analysis, experimental validation, and manuscript preparation, using \texttt{GPT-5.1} as the LLM backbone. A representative interaction log for this workflow is provided in Fig.~\ref{fig:generation-process-direct}. In the remaining two cases, human experts interacted exclusively with the \texttt{MetaAgent}, which autonomously coordinated the required specialized sub-agents, as illustrated in Fig.~\ref{fig:generation-process-indirect}. Across all five cases, human intervention after generation was limited to minor presentational refinements that did not alter the technical substance.

\subsubsection{Mode I: Direct Prompting with Specialized Sub-Agents (Case 1-3)}

\noindent\textbf{Case 1: Momentum-Compatible Federated Subspace Optimization.}
The first paper, \paperone, investigates how federated optimization can be performed efficiently within dynamically evolving low-dimensional subspaces. By projecting the original high-dimensional problem onto a sequence of lower-dimensional subproblems, the proposed method substantially reduces both client-side optimizer memory and client-server communication. Its key algorithmic contribution is a subspace update mechanism that remains compatible with momentum, overcoming a fundamental limitation of previous federated subspace methods. Building upon \cite{yang2021achieving} and \cite{he2025subspace}, the \texttt{AlgorithmAgent} designed the algorithm, while the \texttt{ProverAgent} established a convergence rate of $\mathcal{O}(1/\sqrt{NT})$, explicitly demonstrating linear speedup with respect to the number of clients $N$. 

The \texttt{ExperimentAgent} evaluated the method on MNIST using a three-layer MLP. At a compression ratio of $1/7$, the proposed algorithm achieved a final test accuracy of $0.9511\pm0.0028$, comparable to uncompressed FedAvg-M ($0.9551\pm0.0015$) and substantially outperforming memory-efficient baselines, including FedLoRA-M ($0.9247\pm0.0063$), FedMef ($0.8965\pm0.0062$), NeuLite ($0.8923\pm0.0055$), and FederatedSelect ($0.8759\pm0.0060$). Ablation studies further confirmed its robustness under severe data heterogeneity. To the best of our knowledge, \paperone~is the first federated subspace optimization method that simultaneously supports momentum, reduces memory and communication costs, and provides convergence guarantees with linear speedup.

\vspace{1mm}
\noindent\textbf{Case 2: Heterogeneity-Corrected Primal-Dual Lite Federated Learning.} While the first case focuses on momentum-compatible optimization in low-dimensional subspaces, the second paper, \papertwo, addresses a distinct challenge: correcting client drift under data heterogeneity without compromising the efficiency of subspace optimization. The paper develops a primal-dual federated subspace framework that combines low-dimensional updates with a heterogeneity-correction mechanism. A central technical challenge is preserving accumulated correction information as the active subspace evolves. To address this challenge, the \texttt{AlgorithmAgent} designed a backfill-style update that transfers residual primal-dual information across successive subspaces. The \texttt{ProverAgent} established a non-asymptotic convergence rate of $\tilde{\mathcal{O}}(1/T+\sigma/\sqrt{NKT})$, where $N$ denotes the number of clients, $K$ the number of local updates, $T$ the number of communication rounds, and $\sigma^2$ the gradient variance. This guarantee exhibits linear speedup with respect to both client participation and local computation. Most notably, it strictly improves upon the best convergence rate previously established for human-designed federated subspace methods. For example, the method in~\cite{zhang2025efficient} achieves a rate of $\mathcal{O}(1/\sqrt{T})$ in the deterministic setting with $\sigma=0$, whereas the algorithm generated by ReasFlow \papertwo~achieves the sharper rate of $\mathcal{O}(1/T)$. This result provides a compelling example of ReasFlow autonomously discovering an algorithm with a theoretical guarantee that surpasses the prior state of the art.

The \texttt{ExperimentAgent} validated the algorithm on two fronts: a controlled matrix-regression toy problem and a ResNet-110 benchmark on CIFAR-100. Over 100 training epochs, the proposed method achieved a final test accuracy of 45.43\%, surpassing FedAvg (34.35\%) and FedSub (23.17\%), while narrowing the gap to full-dimensional SCAFFOLD (52.92\%). Ablation studies confirmed stability across varying subspace dimensions and  data heterogeneity. Together, these results demonstrate that the proposed subspace method combines superior theoretical guarantees with strong empirical performance under heterogeneous federated learning settings.

\vspace{1mm}
\noindent\textbf{Case 3: Topology-Independent Decentralized Muon Optimizers for Training LLMs.}
The third paper, \paperthree, addresses a fundamental challenge in extending Muon, a recently developed optimizer for LLM training that orthogonalizes matrix-valued updates using the nonlinear matrix-sign operator $\mathrm{msgn}$, to sparse peer-to-peer networks. Because $\mathrm{msgn}$ does not commute with gossip averaging, Muon cannot be directly integrated into standard decentralized optimization methods. The proposed approach makes two main contributions. First, the \texttt{AlgorithmAgent} developed a general primal-dual framework parameterized by $(A,B,C)$ that systematically integrates Muon with a broad family of classical decentralized algorithms. The framework recovers EXTRA, ED/D$^2$, and gradient tracking as special cases, while carefully sequencing decentralized communication and matrix polarization to resolve the noncommutativity between $\mathrm{msgn}$ and network averaging. Second, the \texttt{ProverAgent} established a topology-separated convergence rate of $\mathcal{O}\bigl(T^{-1/4}\bigr)$ in nuclear-norm geometry. Crucially, the leading-order term is independent of the network topology. This result improves upon the strongest convergence guarantee previously established for human-designed decentralized Muon methods~\cite{he2025demuon}, whose leading term scales as $\mathcal{O}\bigl(T^{-1/4}/(1-\beta)\bigr)$, where $1-\beta$ measures network connectivity. As the network becomes increasingly sparse and $\beta\to 1$, the factor $1/(1-\beta)$ can become prohibitively large. In contrast, the leading convergence rate of \paperthree~remains unaffected by network sparsity.

The \texttt{ExperimentAgent} validated the effectiveness of the proposed method for LLM fine-tuning and further revealed the regime-dependent effects of network topology. The proposed variants matched DeMuon on a ring-network benchmark (ATC-GT: $45.56\%$ vs.\ $45.50\%$) and in near-IID GPT-2 training on WikiText-2 (ED: $18.1396$ vs.\ $18.1401$). On a 20-node non-IID CIFAR-100 benchmark trained for 100 epochs, both ED and ATC-GT substantially outperformed DeMuon (ED: $52.89\%$, ATC-GT: $48.58\%$, and DeMuon: $43.94\%$). These results demonstrate that the proposed framework preserves Muon's effectiveness for large-scale model training while delivering substantial improvements in challenging decentralized settings characterized by sparse connectivity and heterogeneous data.

\subsubsection{Mode II: MetaAgent-Orchestrated Workflow (Case 4–5)}

\noindent\textbf{Case 4: Optimal Acceleration for Decentralized Stochastic Signal Processing.}
The fourth paper, \paperfour, addresses the challenging problem of accelerating decentralized stochastic gradient descent, a fundamental computational primitive in decentralized signal processing. Acceleration in this setting is particularly difficult because stochastic gradient noise and imperfect communication introduce two coupled sources of error. Consensus errors accumulate across the network and interact with stochastic noise, while momentum-based acceleration can amplify both errors and destabilize the optimization process. Moreover, the algorithm must simultaneously achieve fast optimization in terms of the condition number $L/\mu$, efficient information propagation governed by the network spectral gap $1-\beta$, and optimal dependence on the noise level $\sigma^2$ and the number of nodes $n$.

Under the orchestration of the \texttt{MetaAgent}, ReasFlow built upon prior work on accelerated distributed methods and fast gossip~\cite{jakovetic2014fast,scaman2017optimal,berthier2020accelerated} to develop \emph{Multi-Gossip Accelerated DSGD} (MG-ADSGD). The proposed algorithm combines Nesterov-type primal-dual extrapolation with multiple rounds of accelerated gossip averaging in each outer iteration. Under smoothness, strong convexity, and bounded-variance assumptions, the \texttt{ProverAgent} established the non-asymptotic complexity
\[
\widetilde{\mathcal O}\!\left(
\frac{\sigma^2}{\mu n \epsilon}
+
\sqrt{\frac{L}{\mu}}\frac{1}{\sqrt{1-\beta}}
\right).
\]
This guarantee matches, up to logarithmic factors, the fundamental lower bound
\[
\tilde{\Omega}\!\left(
\frac{\sigma^2}{\mu n \epsilon}
+
\sqrt{\frac{L}{\mu}}\frac{1}{\sqrt{1-\beta}}
\right)
\]
for decentralized stochastic strongly convex optimization. The analysis constructs a coupled Lyapunov function that simultaneously tracks optimization progress and network disagreement, carefully controlling the interaction among stochastic noise, momentum amplification, and communication error. It further identifies the optimal communication depth
$R=\widetilde{\mathcal O}(1/\sqrt{1-\beta})$,
which balances accelerated optimization with network-wide information propagation. To the best of our knowledge, \paperfour~is the first decentralized stochastic optimization method to attain this lower complexity bound. It therefore strictly improves upon the strongest convergence guarantees previously achieved by human-designed algorithms and establishes the optimal complexity for decentralized stochastic strongly convex optimization, up to logarithmic factors.

\vspace{1mm}
\noindent\textbf{Case 5: Decentralized Retraction-Free Optimization on the Stiefel Manifold.}
The fifth paper, \paperfive, develops a decentralized retraction-free method for optimization over the Stiefel manifold by integrating an approximate gradient mapping with an EXTRA-type primal-dual recursion. Under the coordination of the \texttt{MetaAgent}, the resulting algorithm, RF-EXTRA, avoids computationally expensive retraction operations while retaining the simple peer-to-peer communication structure of EXTRA. A central technical challenge is to control the violation of the orthogonality constraint while performing primal-dual-type updates directly in the ambient Euclidean space for the nonconvex manifold constrained optimization problem. The theoretical analysis establishes an $\mathcal{O}(1/K)$ convergence rate with a constant stepsize. The proof develops a joint-error contraction inequality that simultaneously controls network disagreement, tracking error, and constraint violation, together with a careful descent analysis of the original objective function. Experiments on decentralized principal component analysis and low-rank matrix completion demonstrate that RF-EXTRA compares favorably with existing decentralized baselines and achieves strong communication efficiency across the tested Stiefel-manifold optimization tasks. To the best of our knowledge, \paperfive~is the first decentralized retraction-free primal-dual-type method for optimization over the Stiefel manifold and achieves the state-of-the-art $\mathcal{O}(1/K)$ convergence rate under standard assumptions and a constant stepsize without communicating gradients or their compressions.

These five case studies collectively demonstrate ReasFlow's ability to produce research outputs of publishable quality across diverse applied mathematics domains, varying levels of technical difficulty, and different modes of human interaction. The first three cases highlight the system's effectiveness under direct prompting, where it addressed challenges including tight variance analysis under stochastic independence, the integration of subspace projection with heterogeneity correction, and the noncommutativity between the $\mathrm{msgn}$ operator and gossip averaging. The final two cases demonstrate the scalability and robustness of the \texttt{MetaAgent}, which autonomously coordinated specialized sub-agents to solve complex problems involving optimal multi-gossip acceleration and retraction-free optimization over constrained manifolds. Collectively, these results show that ReasFlow's knowledge-augmented architecture can generate rigorous theoretical advances and comprehensive empirical validation across federated learning, decentralized signal processing, manifold optimization, and LLM training.

{\color{black}
\begin{table}[t]
	\centering
	\caption{Scores of the generated papers reviewed by the \texttt{PaperReviewAgent}. We use \texttt{GPT-5.1}, \texttt{GPT-5.4}, and \texttt{Gemini-3-Pro} as the reviewers' LLM backend, respectively. \textbf{Bold}: best; \underline{underlined}: second best.}
	\label{tab:paper_scores}
	\begin{tabular}{lcccc}
		\toprule
		AI Agent \textbackslash Reviewer Backend                    & \texttt{GPT-5.1}          & \texttt{GPT-5.4}         & \texttt{Gemini-3-Pro}          & Average          \\
		\midrule
		ChatGPT (\texttt{GPT-5.3})            & 51.7             & 32.2             & 29.8             & 37.9             \\
		ChatGPT Pro (\texttt{GPT-5.4-Pro})    & 62.4             & 44.5             & 41.9             & 49.6             \\
		Gemini (\texttt{Gemini-3.1-Pro})               & 51.5             & 31.4             & 35.1             & 39.3             \\
		Claude Pro (\texttt{Claude Opus 4.6}) & 64.4             & 32.9             & \underline{61.4} & 52.9             \\
		\texttt{CycleResearcher-ML-12B}       & 34.6             & 19.1             & 10.1             & 21.2             \\
		DeepScientist (\texttt{GPT-5.4})      & \underline{71.9} & \underline{56.6} & 43.5             & \underline{57.3} \\
		AI Scientist-v2 (\texttt{GPT-5.4})    & 60.6             & 34.7             & 29.6             & 41.6             \\
		ARIS (\texttt{GPT-5.4})               & 59.3             & 52.9             & 38.6             & 50.3             \\
        \midrule
		ReasFlow (\texttt{GPT-5.4})           & \textbf{80.5}    & \textbf{65.6}    & \textbf{87.6}    & \textbf{77.9}    \\
		\bottomrule
	\end{tabular}
\end{table}
}
\begin{table}[t]
\centering
\caption{Per-backbone evaluation of the \texttt{SurveyAgent} vs.\ the base LLM. Each score is the total (Content Accuracy + Citation Relevance, out of 20) averaged across three evaluation models.}
\label{tab:survey_summary}
\begin{tabular}{llccc}
\toprule
Task & Backbone & Base & ReasFlow & Adv. \\
\midrule
\multirow{5}{*}{\shortstack[l]{\paperone}}
 & \texttt{DeepSeek-v3.2}   & 12.65 & 17.93 & +5.27 \\
 & \texttt{GPT-5.1}    & 19.20 & 17.78 & $-1.41$ \\
 & \texttt{GPT-5.4}    & 9.08  & 16.75 & +7.67 \\
 & \texttt{Claude Sonnet 4.6} & 11.07 & 17.24 & +6.17 \\
\cmidrule(lr){2-5}
 & Average    & 13.00 & 17.42 & +4.42 \\
\midrule    
\multirow{5}{*}{\shortstack[l]{\papertwo}}
 & \texttt{DeepSeek-v3.2}   & 13.28 & 14.81 & +1.53 \\
 & \texttt{GPT-5.1}    & 14.52 & 16.63 & +2.11 \\
 & \texttt{GPT-5.4}    & 13.23 & 16.15 & +2.92 \\
 & \texttt{Claude Sonnet 4.6} & 14.03 & 14.90 & +0.88 \\
\cmidrule(lr){2-5}
 & Average    & 13.77 & 15.62 & +1.86 \\
\midrule
\multirow{5}{*}{\shortstack[l]{\paperthree}}
 & \texttt{DeepSeek-v3.2}   & 8.14  & 18.36 & +10.23 \\
 & \texttt{GPT-5.1}    & 7.00  & 16.03 & +9.03 \\
 & \texttt{GPT-5.4}    & 13.30 & 14.51 & +1.20 \\
 & \texttt{Claude Sonnet 4.6} & 17.32 & 11.62 & $-5.70$ \\
\cmidrule(lr){2-5}
 & Average    & 11.44 & 15.13 & +3.69 \\
\midrule
\multicolumn{2}{l}{Overall Average} & 12.73 & 16.06 & \textbf{+3.32} \\
\bottomrule
\end{tabular}
\end{table}

\subsection{Performance Evaluation on Automatic Full-Paper Generation}\label{subsec:exp-paper}

To evaluate ReasFlow's proficiency in autonomous end-to-end manuscript generation, we conducted a controlled experiment on the automated production of \paperone. All baseline systems were provided with an identical initial prompt, comprising the core research idea and a set of seminal references (see Appendix~\ref{app:exp-detail}), and evaluated after a single autonomous trial without human intervention. In our framework, the \texttt{MetaAgent} (powered by \texttt{GPT-5.4}) orchestrated the entire multi-agent pipeline.

We benchmark ReasFlow against several publicly available autonomous research systems: (i) proprietary web-based reasoning systems \textit{ChatGPT} (\texttt{GPT-5.3}), \textit{ChatGPT Pro} (\texttt{GPT-5.4-Pro}), \textit{Gemini} (\texttt{Gemini-3.1-Pro}), and \textit{Claude Pro} (\texttt{Claude Opus 4.6}); (ii) \textit{DeepScientist}~\cite{weng2025deepscientist} (\texttt{GPT-5.4}); (iii) \textit{CycleResearcher}~\cite{weng2024cycleresearcher} (\texttt{CycleResearcher -ML-12B}); (iv) \textit{AI-Scientist-v2}~\cite{yamada2025ai} (\texttt{GPT-5.4}); and (v) \textit{ARIS}~\cite{yang2026aris} (\texttt{GPT-5.4}).

To ensure impartial assessment, we developed a \texttt{PaperReviewAgent} that evaluates generated manuscripts across nine scholarly dimensions: (i) mathematical rigor, (ii) originality and significance, (iii) quality of numerical experiments, (iv) consistency between theoretical claims and empirical results, (v) writing and presentation, (vi) citation accuracy, (vii) internal logical consistency, (viii) visual and formatting quality, and (ix) AI and ethics compliance. Detailed scoring criteria and evaluation prompts are provided in Appendix~\ref{app:exp-detail}.

As shown in Table~\ref{tab:paper_scores}, ReasFlow achieves an average score of \textbf{77.9}, outperforming the strongest baseline (DeepScientist, 57.3) by over 20 points. Notably, ReasFlow maintains robust performance across all three reviewer backends, indicating that its advantage is not an artifact of evaluator bias. The performance gap is particularly pronounced in dimensions requiring deep reasoning, mathematical rigor and theory-experiment consistency, which we analyze further in Sec.~\ref{subsubsec:exp-prover}.

\subsection{Performance Evaluation on Section-Level Generation}

To isolate the contributions of individual components, we evaluate ReasFlow's specialized agents on their respective subtasks. We focus particularly on the \texttt{ProverAgent}, which addresses the most challenging reasoning bottleneck.

\subsubsection{Performance Evaluation for the {Survey Agent}}\label{subsubsec:exp-survey}

We evaluate the \texttt{SurveyAgent} by comparing the \textit{Related Work} section it generates against that produced by the base LLM across three survey tasks (\paperone, \papertwo, and \paperthree). Each task is generated with four LLM backbones (\texttt{DeepSeek-v3.2}, \texttt{GPT-5.1}, \texttt{GPT-5.4}, and \texttt{Claude Sonnet~4.6}) and independently scored by three evaluation models (\texttt{GPT-5.1}, \texttt{Sonnet~4.6}, and \texttt{DeepSeek-v3.2}) for robustness. The evaluation assesses two dimensions on a 1-10 scale: (i) \textit{Content Accuracy}, which verifies whether cited papers are correctly described by cross-referencing a 339-paper FAISS vector database; and (ii) \textit{Citation Relevance}, which measures whether the selected references align with the survey plan. Full evaluation details, including the rubric and per-evaluator breakdowns, are provided in Appendix~\ref{app:exp-survey}.

\begin{table}[t]
\centering
\caption{Algorithm script first-run success rate. S.T. represents the number of script-test task records evaluated for each system, and S.R. represents the first-run success rate. \textbf{Bold}: best; \underline{underlined}: second best.}
\label{tab:runnability}
\begin{tabular}{lcc}
\toprule
System & S.T. & S.R. \\
\midrule
ReasFlow (\texttt{GPT-5.4}) & 35 & \textbf{100\%} \\
ReasFlow (\texttt{GPT-5.1}) & 35 & 94\% \\
Codex (\texttt{GPT-5.4})                                 & 35 & \underline{97\%} \\
Claude Code (\texttt{Claude Sonnet 4.6})                                & 35 & 71\% \\
AI Scientist-v2 (\texttt{GPT-5.4})                  & 35 & 60\% \\
ARIS (\texttt{Claude Sonnet 4.6})                                       & 35 & 46\% \\
\bottomrule
\end{tabular}
\end{table}

Table~\ref{tab:survey_summary} reports the per-backbone results. The \texttt{SurveyAgent} achieves an overall advantage of \textbf{+3.32 points}, with the largest gain of \textbf{+10.23} observed for \texttt{DeepSeek-v3.2} on \paperthree. The improvement is predominantly driven by Content Accuracy (+2.97 on average), as the agent's citations are retrieved from a verified paper library rather than relying on parametric knowledge. Across all 36 paired comparisons (4 backbones $\times$ 3 tasks $\times$ 3 evaluation models), the \texttt{SurveyAgent} wins in \textbf{29 cases (80.6\% win rate)}.

{\color{black}
\subsubsection{Performance Evaluation for the Algorithm Agent}\label{subsubsec:exp-alg}

Following MLR-Bench~\cite{chen2026mlr}, we introduce \textbf{OR-Bench}, a benchmark designed to evaluate AI agents' algorithmic design and experiment implementation capabilities, with a focus on open-ended optimization research. We compare the \texttt{AlgorithmAgent} (using \texttt{GPT-5.1} and \texttt{GPT-5.4}) against ARIS (\texttt{Claude Sonnet 4.6}), Codex (\texttt{GPT-5.4}), Claude Code (\texttt{Claude Sonnet 4.6}), and AI Scientist-v2 (\texttt{GPT-5.4}) across 35 tasks (20 from MLR-Bench and 15 from OR-Bench). Given the same task description and research-idea input, each agent is required to produce an algorithm design and executable Python implementation within 3,600 seconds. As reported in Table~\ref{tab:runnability}, ReasFlow with \texttt{GPT-5.4} achieves a leading first-run success rate of 100\%, compared with 46\%--97\% for the baselines, demonstrating the robustness of ReasFlow's \texttt{AlgorithmAgent}.
}

\subsubsection{Performance Evaluation for the {Prover Agent}}\label{subsubsec:exp-prover}

We assess the \texttt{ProverAgent} by evaluating its ability to generate a complete convergence analysis for a given algorithm. The algorithm proposed in \papertwo\ is selected as the target, ensuring that no prior theoretical results are publicly available; therefore, all systems must derive the theory independently. Both ReasFlow's \texttt{ProverAgent} and baseline agents are provided with the same initial prompt and allowed only a single trial, with no iterative refinement or additional hints.

We compare against the following baselines: (i) general-purpose conversational agents accessible via official web interfaces (ChatGPT and Gemini); (ii) \textit{ReasLingo}, a publicly available agent on the ReasLab platform built on the Gemini CLI; (iii) \textit{ProofGrader} \cite{ma2025reliable}, an open-source AI system designed for automated mathematical proving; and (iv) \textit{QED} \cite{an2026qed}, a multi-agent pipeline for generating rigorous natural-language proofs for given mathematical problems. All agents in ReasFlow operate with a \texttt{medium} reasoning level to reduce computational overhead.

To evaluate the generated reports, we developed a \texttt{ProofReviewAgent} that scores each report across five dimensions: (i) academic formalism and formatting (10 points), (ii) target alignment and relevance (10 points), (iii) consistency of assumptions and notation (10 points), (iv) logical deduction and correctness (50 points), and (v) theoretical significance and strength (20 points). The complete evaluation prompt is provided in Appendix~\ref{app:exp-detail}.

{\color{black}
\begin{table}[t]
	\centering
	\caption{Scores of the theoretical reports reviewed by the \texttt{ProofReviewAgent}. We use \texttt{GPT-5.1}, \texttt{GPT-5.4}, and \texttt{Gemini-3-Pro} as the reviewers' LLM backend, respectively. \textbf{Bold}: best; \underline{underlined}: second best.}
	\label{tab:prover_scores}
	\begin{tabular}{lccccc}
		\toprule
		AI Agent                 & Reasoning Effort & \texttt{GPT-5.1}       & \texttt{GPT-5.4}       & \texttt{Gemini-3-Pro}      & Average          \\
		\midrule
		ChatGPT (\texttt{GPT-5.3})        & Default          & 29             & 15             & 7            & 17               \\
		Gemini (\texttt{Gemini-3.1-Pro})           & Default          & 81             & 24             & 48           & 51               \\
		ReasLingo (\texttt{Gemini-3-Pro}) & High             & 91             & 48             & 40           & 59.7             \\
		ReasLingo (\texttt{GPT-5.1})      & High             & 87             & 35             & 40           & 54.0            \\
		ReasLingo (\texttt{GPT-5.4})      & High             & 92             & 57             & 60           & 69.7             \\
		ProofGrader (\texttt{GPT-5.1})    & High             & 92             & 34             & 70           & 65.3             \\
		ProofGrader (\texttt{GPT-5.4})    & High             & 93             & 39             & \underline{75} & 69               \\
		QED (\texttt{GPT-5.4})            & High             & \textbf{98}    & 51             & 70           & 73               \\
        \midrule
		ReasFlow (\texttt{Gemini-3-Pro})  & Medium           & 91             & \underline{62} & \textbf{100} & \underline{84.3} \\
		ReasFlow (\texttt{GPT-5.1})       & Medium           & 93             & 45             & \textbf{100} & 79.3             \\
		ReasFlow (\texttt{GPT-5.4})       & Medium           & \underline{94} & \textbf{78}    & \textbf{100} & \textbf{90.7}    \\
		\bottomrule
	\end{tabular}
\end{table}
}

As reported in Table~\ref{tab:prover_scores}, ReasFlow with \texttt{GPT-5.4} achieves the highest average score of \textbf{90.7}, compared to 73 for the best-performing baseline (QED with \texttt{GPT-5.4}). Notably, when evaluated by \texttt{Gemini-3-Pro}, ReasFlow attains a perfect score of 100 across all three LLM backbones. Detailed dimension-wise scores are provided in Table~\ref{tab:prover_scores_detail} (Appendix~\ref{app:exp-detail}). These results confirm that the knowledge augmentation pipeline (Sec.~\ref{sec:knowledge_aug}) effectively instills domain-specific proof discipline into the LLM, enabling it to handle complex reasoning tasks that baselines fail to address.

{\color{black}
\subsubsection{Performance Evaluation for the Experiment Agent}\label{subsubsec:exp-exp}

To evaluate the \texttt{ExperimentAgent}, we adopt an experimental setup similar to that in Sec.~\ref{subsubsec:exp-alg}. Specifically, each agent is given its own outputs from the algorithm design phase (Sec.~\ref{subsubsec:exp-alg}) and must drive real experiments to completion within a time limit of 7,200 seconds. We evaluate performance using \texttt{GPT-5.4} and \texttt{Claude Sonnet 4.6} as judges, based on the task description, the execution artifacts, and the resulting reports. Beyond the six evaluation dimensions inherited from MLR‑Bench (Soundness, Consistency, Completeness, Novelty, Insightfulness, and Significance), we introduce five additional metrics: (i) \textit{Design Soundness}, which measures the quality of the experimental setup including baselines, evaluation metrics, ablation studies, and internal logic; (ii) \textit{Design Completeness}, which assesses whether all necessary components of the experiment are included in the plan; (iii) \textit{Algorithm–Experiment Alignment}, which evaluates how well the experimental design exploits the algorithm’s core contributions; (iv) \textit{Experimental Effectiveness}, which examines whether the results convincingly support the algorithm’s claims; and (v) \textit{Writeup Readability}, which captures whether the report provides the information necessary for downstream paper writing. Scores are averaged over three independent runs from each judge and then across tasks. We additionally evaluate \textit{Hallucination Suspicion}: a task is flagged if any judge run finds that the agent's workspace lacks result files or contains only trivial or synthetic outputs, indicating insufficient evidence of genuine execution. As shown in Table~\ref{tab:chain2_scores}, ReasFlow outperforms the baselines in many aspects, particularly on metrics related to experimental design and writeup readability, while ReasFlow with \texttt{GPT-5.4} also incurs the lowest hallucination suspicion, demonstrating strong performance under the evaluation protocol.

\begin{table*}[t]
\centering
\caption{Experiment Execution scores (1--10) on Soundness (Sdn.), Consistency (Cst.), Completeness (Cpl.), Novelty (Nvl.), Insightfulness (Isf.), Significance (Sgn.), Design Soundness (D.S.), Design Completeness (D.C.), Algorithm--Experiment Alignment (AEA.), Experimental Effectiveness (E.E.), Writeup Readability (W.R.), and Hallucination Suspicion (H.S.); lower H.S. is better. \textbf{Bold}: best; \underline{underline}: second best.}
\label{tab:chain2_scores}
\resizebox{\linewidth}{!}{%
\begin{tabular}{lccccccccccccc}
\toprule
System & $n$ & Sdn. & Cst. & Cpl. & Nvl. & Isf. & Sgn. & D.S. & D.C. & AEA. & E.E. & W.R. & H.S. \\
\midrule
ReasFlow (\texttt{GPT-5.4}) & 35 & \textbf{8.49} & \textbf{8.21} & \textbf{7.96} & \textbf{6.41} & \textbf{7.83} & \textbf{5.02} & \textbf{7.99} & \textbf{8.00} & \textbf{8.95} & \textbf{8.07} & \textbf{8.13} & \textbf{9\%} \\
ReasFlow (\texttt{GPT-5.1}) & 35 & \underline{7.26} & \underline{6.77} & 6.03 & 5.27 & \underline{6.21} & \underline{3.45} & \underline{6.85} & \underline{6.50} & \underline{7.49} & \underline{6.01} & \underline{6.58} & 37\% \\
\midrule
ARIS~ (\texttt{Claude Sonnet 4.6})                     & 35 & 5.32 & 5.21 & 4.77 & 5.51 & 4.93 & 3.35 & 6.23 & 5.70 & 6.84 & 5.04 & 4.59 & 46\% \\
Codex (\texttt{GPT-5.4})                                    & 35 & 6.50 & 6.55 & \underline{6.17} & 4.81 & 5.95 & 3.44 & 6.01 & 5.91 & 6.61 & 5.83 & 5.65 & \underline{26\%} \\
Claude Code (\texttt{Claude Sonnet 4.6})                                   & 35 & 5.91 & 5.68 & 5.53 & 4.98 & 5.42 & 3.40 & 5.86 & 5.72 & 6.50 & 5.40 & 5.11 & 43\% \\
AI Scientist-v2 (\texttt{GPT-5.4}) & 35 & 1.28 & 3.07 & 1.72 & \underline{5.64} & 4.94 & 2.35 & 6.39 & 6.14 & 7.09 & 3.16 & 3.84 & 100\% \\
\bottomrule
\end{tabular}%
}
\end{table*}
}

\subsubsection{Performance Evaluation for the {Introduction Agent}}\label{subsubsec:exp-intro}

We evaluate the \texttt{IntroductionAgent} by comparing the \textit{Introduction} section it generates against that produced by the base model (\texttt{GPT-4o}) across the three papers. For each configuration, five independent runs are performed and results are averaged. Three independent evaluation models (\texttt{GPT-5.4}, \texttt{Claude Opus 4.6}, and \texttt{Gemini-3.1-Pro-Preview}) assess each output along three axes: (i) \textit{Content Coverage}, measuring how comprehensively the introduction captures the paper's key results; (ii) \textit{Faithfulness}, quantifying hallucination avoidance as $1 - (\text{\#unsupported claims}/\text{\#total claims})$; and (iii) \textit{Content Quality}, a rubric-based score assessing six dimensions including domain emphasis, algorithm detail, and experimental depth. Two composite scores, Overall Balanced (OB) and Overall Technical Focus (OTF), are computed as weighted combinations; the latter up-weights content quality and core scientific coverage. Table~\ref{tab:intro_summary} summarizes per-evaluator results for these two composites; axis-wise scores and additional detail are in Appendix~\ref{app:exp-intro}.

\begin{table}[htbp]
\centering
\caption{Per-evaluator results of the \texttt{IntroductionAgent} (RF) vs.\ the base model on two composite metrics. Each cell is the mean over 5 independent runs.}
\label{tab:intro_summary}
\begin{tabular}{ll|ccc|ccc}
\toprule
\multirow{2}{*}{Paper} & \multirow{2}{*}{Evaluator} & \multicolumn{3}{c|}{Overall Balanced (OB)} & \multicolumn{3}{c}{Overall Tech.\ Focus (OTF)} \\
\cmidrule(lr){3-5} \cmidrule(lr){6-8}
 & & Base & RF & Adv. & Base & RF & Adv. \\
\midrule
\multirow{4}{*}{\paperone}
 & \texttt{GPT-5.4} & 0.554 & 0.683 & +23.2\% & 0.520 & 0.661 & +27.0\% \\
 & \texttt{Opus 4.6} & 0.510 & 0.739 & +45.0\% & 0.461 & 0.708 & +53.6\% \\
 & \texttt{Gemini-3.1-Pro-Preview} & 0.678 & 0.795 & +17.3\% & 0.658 & 0.788 & +19.8\% \\
\cmidrule(lr){2-8}
 & Average & 0.580 & 0.739 & +27.3\% & 0.547 & 0.719 & +31.6\% \\
\midrule
\multirow{4}{*}{\papertwo}
 & \texttt{GPT-5.4} & 0.552 & 0.643 & +16.4\% & 0.510 & 0.603 & +18.3\% \\
 & \texttt{Opus 4.6} & 0.527 & 0.642 & +21.6\% & 0.474 & 0.602 & +27.0\% \\
 & \texttt{Gemini-3.1-Pro-Preview} & 0.656 & 0.705 & +7.6\% & 0.627 & 0.674 & +7.4\% \\
\cmidrule(lr){2-8}
 & Average & 0.578 & 0.663 & +14.7\% & 0.537 & 0.626 & +16.6\% \\
\midrule
\multirow{4}{*}{\paperthree}
 & \texttt{GPT-5.4} & 0.465 & 0.648 & +39.5\% & 0.418 & 0.622 & +48.8\% \\
 & \texttt{Opus 4.6} & 0.482 & 0.657 & +36.2\% & 0.433 & 0.629 & +45.5\% \\
 & \texttt{Gemini-3.1-Pro-Preview} & 0.587 & 0.733 & +25.0\% & 0.548 & 0.719 & +31.3\% \\
\cmidrule(lr){2-8}
 & Average & 0.511 & 0.679 & +32.9\% & 0.466 & 0.657 & +40.9\% \\
\midrule
\multicolumn{2}{l|}{Overall Average} & 0.557 & 0.694 & +24.6\% & 0.517 & 0.667 & \textbf{+29.2\%} \\
\bottomrule
\end{tabular}
\end{table}

As shown in Table~\ref{tab:intro_summary}, the \texttt{IntroductionAgent} consistently outperforms the base model across all evaluators and papers, with an overall average gain of \textbf{+24.6\%} on OB and \textbf{+29.2\%} on OTF. At the axis level (detailed in Appendix~\ref{app:exp-intro}), the most substantial improvement is in Content Quality (\textbf{+52.3\%} on average), with a peak of \textbf{+121.4\%} on \paperthree\ under the Opus 4.6 evaluator, indicating that the agent produces introductions with significantly richer technical depth. Improvements are most pronounced on \paperthree, where the base model struggles most, suggesting that the \texttt{IntroductionAgent} scales better on harder tasks.

\subsubsection{Performance Evaluation for the {Writing Agent}}\label{subsubsec:exp-writer}

We evaluate the \texttt{WritingAgent} on the task of assembling the complete \paperone~asset bundle (proved theorems, algorithm pseudocode, experiment report with raw numerical data and pre-generated figures, introduction and related-work drafts, and reference \texttt{.bib} files) into a compilable CSIAM-AM manuscript. All compared systems receive exactly the same input bundle, the same task prompt (the shared \texttt{PROMPT.md} of Appendix~\ref{app:exp-writer}), and the same CSIAM-AM template; only interface-level adapters differ.

Eleven systems are compared: our \texttt{WritingAgent} run with a \texttt{GPT-5.1} backend, and ten baselines obtained by pairing two general-purpose coding agents (\textit{Claude Code} and \textit{Codex}) with two skill configurations (vanilla and the scientific-writing skill pack ARIS~\cite{yang2026aris}) across two model backends (\texttt{GPT-5.1} and \texttt{GPT-5.4}), plus the scientific-writing system \textit{ReasLingo} with \texttt{GPT-5.1} and \texttt{GPT-5.4}.

Each generated manuscript is scored along four dimensions: (I)~\textit{Faithfulness}, whether every statement in the paper can be traced back to the assets without fabrication or distortion; (II)~\textit{Coverage}, whether the core theorems, algorithms, experiments, figures, and references are actually incorporated; (III)~\textit{Writing Quality}, whether the prose is clear, coherent, well-argued, and uses evidence properly; and (IV)~\textit{Submittability}, whether the paper compiles cleanly, presents an acceptable visual layout, and obeys the template. Each dimension is decomposed into sub-dimensions (5/5/4/3) that are scored independently on a $0$--$10$ scale by three reviewer models (\texttt{GPT-5.1}, \texttt{GPT-5.4}, \texttt{Gemini-3-Pro}), with Phase~IV additionally using a vision describer plus programmatic compile and template checks. Sub-dimension scores are averaged with equal weights within a dimension, across the three reviewers, and finally across the four dimensions, giving the overall equal-weight score reported below. Complete per-phase scoring rules, prompts, and sub-dimension tables are provided in Appendix~\ref{app:exp-writer}.

\begin{table}[htbp]
	\centering
	\caption{Evaluation of the manuscripts produced by the \texttt{WritingAgent} and ten baselines on \paperone. Columns~I--IV are the per-dimension scores (each the mean across its sub-dimensions and across three reviewer models); \textit{Avg.} is the equal-weight mean of the four dimensions. \textbf{Bold}: best; \underline{underlined}: second best.}
	\label{tab:writer_scores}
	\begin{tabular}{lccccc}
		\toprule
		System (Executor, Skills, Model)               & I. Faith.     & II. Cov.      & III. Write.   & IV. Subm.      & Avg.          \\
		\midrule		
		Claude Code + ARIS (\texttt{GPT-5.1})                   & 7.67          & 6.33          & 5.42          & 2.00          & 5.36          \\
        Codex, vanilla (\texttt{GPT-5.4})                       & 6.47          & 5.33          & 4.58          & 6.71          & 5.77          \\
        Codex + ARIS (\texttt{GPT-5.4})                         & 8.47          & 4.33          & 5.58          & 7.24          & 6.41          \\
        ReasLingo (\texttt{GPT-5.4})                            & 8.47          & 4.80          & 5.08          & 7.87          & 6.56          \\
        Claude Code, vanilla (\texttt{GPT-5.1})                 & 7.87          & 7.20          & 7.17          & 4.62          & 6.72          \\
        Codex, vanilla (\texttt{GPT-5.1})                       & 7.67          & 5.67          & 7.33          & 7.82          & 7.12          \\
        ReasLingo (\texttt{GPT-5.1})                            & 8.47          & 7.60          & 6.25          & 8.84          & 7.79          \\
        Claude Code + ARIS (\texttt{GPT-5.4})                   & \underline{9.07}          & 8.13          & 7.33          & 7.56          & 8.02          \\
        Codex + ARIS (\texttt{GPT-5.1})                         & 7.60          & 8.07          & \underline{7.75}          & 8.89          & 8.08          \\
        Claude Code, vanilla (\texttt{GPT-5.4})                 & 8.20          & \underline{9.13}          & 7.67          & \textbf{10.00} & \underline{8.75}          \\
        \midrule
		ReasFlow \texttt{WritingAgent} (\texttt{GPT-5.1})       & \textbf{9.73} & \textbf{9.47} & \textbf{8.16} & \underline{9.87}          & \textbf{9.31} \\
		\bottomrule
	\end{tabular}
\end{table}

As shown in Table~\ref{tab:writer_scores}, the ReasFlow \texttt{WritingAgent} attains the highest overall score of \textbf{9.31}, leading the next best system by \textbf{0.56} absolute points and the median baseline by over \textbf{2.2} points. It is the only system that ranks first on three of the four dimensions simultaneously (I-Faithfulness, II-Coverage, III-Writing Quality); on IV-Submittability it is runner-up (9.87), losing only by a \textbf{0.13} margin to a short 11-page baseline that delivers a visually clean PDF but trails on content-level dimensions (II-Coverage and III-Writing Quality). The ranking differs substantially between dimensions, confirming that the four axes measure genuinely different aspects of manuscript assembly rather than a single latent quality; for instance, \textit{ReasLingo} (\texttt{GPT-5.4}) places third on I-Faithfulness but collapses to tenth on II-Coverage because its manuscript omits most of the core algorithm and experiment sections, while the \textit{Codex} family excels on I2-Algorithm Fidelity but loses on II-Coverage and IV-Submittability due to missing experimental integration and heavy compile-time overflows. These patterns suggest that assembling a publication-ready manuscript from heterogeneous research artifacts is not a narrow prompt-engineering task but requires coordinated planning, selection, polishing, and typesetting, which aligns with the two-tier design (Coordinator plus parallel \texttt{SubWriter}s) of the ReasFlow \texttt{WritingAgent}.

{\color{black}
\section{Conclusion and Future Work}

In this work, we introduced \textbf{ReasFlow}, an end-to-end autonomous multi-agent system for reasoning-centric scientific discovery in applied mathematics. ReasFlow addresses the distinctive challenges of theory-oriented research, including open-ended problem formulation, rigorous proof construction, and the effective use of implicit domain knowledge. It implements a collaborative paradigm in which the human researcher serves as the principal investigator, while the system carries out the labor-intensive stages of the research pipeline. Its effectiveness is driven by two complementary mechanisms: an internal verification loop that audits theoretical derivations and identifies critical errors before human inspection, and an automated knowledge-retrieval pipeline that proactively extracts domain insights and procedural heuristics into lightweight knowledge cards. Together, these mechanisms enable ReasFlow, through internal knowledge augmentation, to achieve reasoning performance comparable to that obtained with direct expert guidance.

Our evaluation comprises five case studies spanning federated learning, decentralized signal processing, manifold optimization, and LLM optimizers. Across these diverse applied mathematics settings, ReasFlow produced novel algorithms with rigorous convergence guarantees, strong empirical performance, and complete research manuscripts. Comparative evaluations show that ReasFlow consistently outperforms state-of-the-art publicly accessible AI research systems under carefully designed LLM-based review rubrics, while component-level analyses confirm the substantial contribution of knowledge augmentation. These findings establish ReasFlow as a practical and scalable framework for automating theory-driven research in applied mathematics.

Despite these promising results, several limitations remain. The current implementation of ReasFlow relies heavily on a built-in knowledge base tailored primarily to applied math and related areas. Extending the system to a broader range of theory-intensive disciplines will require automated knowledge-extraction pipelines capable of identifying and organizing domain-specific concepts, implicit insights, and methodological heuristics across disciplinary boundaries. Moreover, although the internal verification loop detects many common reasoning errors, it does not guarantee formal correctness. Integrating ReasFlow with interactive theorem provers and formal verification systems such as Lean, Coq, or Isabelle could provide stronger guarantees for critical theoretical claims. Finally, the system's capabilities remain constrained by the reasoning ability of its underlying large language models. As foundation models continue to advance, ReasFlow's modular, knowledge-augmented architecture can provide a robust foundation for increasingly autonomous and ambitious scientific discovery.

ReasFlow is publicly available via its GitHub repository and the ReasLab platform, which provides a collaborative workspace for researchers to pursue theoretical innovation with AI assistance. By bridging the gap between general-purpose language models and the specialized demands of reasoning-centric discovery, we hope that ReasFlow will accelerate theoretical research in applied mathematics and enable scientists to explore ideas that were previously constrained by the substantial effort required for rigorous derivation, experimental validation, and manuscript preparation.}

\bibliography{main}

@inproceedings{he2025subspace,
  title={Subspace Optimization for Large Language Models with Convergence Guarantees},
  author={He, Yutong and Li, Pengrui and Hu, Yipeng and Chen, Chuyan and Yuan, Kun},
  booktitle={International Conference on Machine Learning},
  pages={22468--22522},
  year={2025},
  organization={PMLR}
}

@article{he2025demuon,
  title={DeMuon: A Decentralized Muon for Matrix Optimization over Graphs},
  author={He, Chuan and Ren, Shuyi and Mao, Jingwei and Larsson, Erik G},
  journal={arXiv preprint arXiv:2510.01377},
  year={2025}
}

@article{zhang2025efficient,
  title={An Efficient Subspace Algorithm for Federated Learning on Heterogeneous Data},
  author={Zhang, Jiaojiao and Xu, Yuqi and Yuan, Kun},
  journal={arXiv preprint arXiv:2509.05213},
  year={2025}
}

@article{yang2021achieving,
  title={Achieving Linear Speedup with Partial Worker Participation in Non-IID Federated Learning},
  author={Yang, Haibo and Fang, Minghong and Liu, Jia},
  journal={Proceedings of ICLR},
  year={2021}
}

@article{romera2024mathematical,
  title={Mathematical discoveries from program search with large language models},
  author={Romera-Paredes, Bernardino and Barekatain, Mohammadamin and Novikov, Alexander and Balog, Matej and Kumar, M Pawan and Dupont, Emilien and Ruiz, Francisco JR and Ellenberg, Jordan S and Wang, Pengming and Fawzi, Omar and others},
  journal={Nature},
  volume={625},
  number={7995},
  pages={468--475},
  year={2024},
  publisher={Nature Publishing Group UK London}
}

@article{bubeck2025early,
  title={Early science acceleration experiments with {GPT}-5},
  author={Bubeck, S{\'e}bastien and Coester, Christian and Eldan, Ronen and Gowers, Timothy and Lee, Yin Tat and Lupsasca, Alexandru and Sawhney, Mehtaab and Scherrer, Robert and Sellke, Mark and Spears, Brian K and others},
  journal={arXiv preprint arXiv:2511.16072},
  year={2025}
}

@article{trinh2024solving,
  title={Solving olympiad geometry without human demonstrations},
  author={Trinh, Trieu H and Wu, Yuhuai and Le, Quoc V and He, He and Luong, Thang},
  journal={Nature},
  volume={625},
  number={7995},
  pages={476--482},
  year={2024},
  publisher={Nature Publishing Group UK London}
}

@article{huang2025winning,
  title={Winning gold at {IMO} 2025 with a model-agnostic verification-and-refinement pipeline},
  author={Huang, Yichen and Yang, Lin F},
  journal={arXiv preprint arXiv:2507.15855},
  year={2025}
}

@article{qiu2025physics,
  title={Physics supernova: {AI} agent matches elite gold medalists at {IPHO} 2025},
  author={Qiu, Jiahao and Shi, Jingzhe and Juan, Xinzhe and Zhao, Zelin and Geng, Jiayi and Liu, Shilong and Wang, Hongru and Wu, Sanfeng and Wang, Mengdi},
  journal={arXiv preprint arXiv:2509.01659},
  year={2025}
}

@article{lu2024ai,
  title={The {AI} scientist: Towards fully automated open-ended scientific discovery},
  author={Lu, Chris and Lu, Cong and Lange, Robert Tjarko and Foerster, Jakob and Clune, Jeff and Ha, David},
  journal={arXiv preprint arXiv:2408.06292},
  year={2024}
}

@article{yamada2025ai,
  title={The {AI} scientist-v2: Workshop-level automated scientific discovery via agentic tree search},
  author={Yamada, Yutaro and Lange, Robert Tjarko and Lu, Cong and Hu, Shengran and Lu, Chris and Foerster, Jakob and Clune, Jeff and Ha, David},
  journal={arXiv preprint arXiv:2504.08066},
  year={2025}
}

@article{weng2024cycleresearcher,
  title={Cycleresearcher: Improving automated research via automated review},
  author={Weng, Yixuan and Zhu, Minjun and Bao, Guangsheng and Zhang, Hongbo and Wang, Jindong and Zhang, Yue and Yang, Linyi},
  journal={arXiv preprint arXiv:2411.00816},
  year={2024}
}

@article{weng2025deepscientist,
  title={Deepscientist: Advancing frontier-pushing scientific findings progressively},
  author={Weng, Yixuan and Zhu, Minjun and Xie, Qiujie and Sun, Qiyao and Lin, Zhen and Liu, Sifan and Zhang, Yue},
  journal={arXiv preprint arXiv:2509.26603},
  year={2025}
}

@article{schmidgall2025agent,
  title={Agent laboratory: Using {LLM} agents as research assistants},
  author={Schmidgall, Samuel and Su, Yusheng and Wang, Ze and Sun, Ximeng and Wu, Jialian and Yu, Xiaodong and Liu, Jiang and Moor, Michael and Liu, Zicheng and Barsoum, Emad},
  journal={Findings of the Association for Computational Linguistics: EMNLP 2025},
  pages={5977--6043},
  year={2025},
  publisher={Association for Computational Linguistics}
}

@article{gottweis2025towards,
  title={Towards an {AI} co-scientist},
  author={Gottweis, Juraj and Weng, Wei-Hung and Daryin, Alexander and Tu, Tao and Palepu, Anil and Sirkovic, Petar and Myaskovsky, Artiom and Weissenberger, Felix and Rong, Keran and Tanno, Ryutaro and others},
  journal={arXiv preprint arXiv:2502.18864},
  year={2025}
}

@article{skarlinski2024language,
  title={Language agents achieve superhuman synthesis of scientific knowledge},
  author={Skarlinski, Michael D and Cox, Sam and Laurent, Jon M and Braza, James D and Hinks, Michaela and Hammerling, Michael J and Ponnapati, Manvitha and Rodriques, Samuel G and White, Andrew D},
  journal={arXiv preprint arXiv:2409.13740},
  year={2024}
}

@article{wang2024autosurvey,
  title={Autosurvey: Large language models can automatically write surveys},
  author={Wang, Yidong and Guo, Qi and Yao, Wenjin and Zhang, Hongbo and Zhang, Xin and Wu, Zhen and Zhang, Meishan and Dai, Xinyu and Zhang, Min and Wen, Qingsong and others},
  journal={Advances in neural information processing systems},
  volume={37},
  pages={115119--115145},
  year={2024}
}

@article{xu2026idea2story,
  title={Idea2Story: An Automated Pipeline for Transforming Research Concepts into Complete Scientific Narratives},
  author={Xu, Tengyue and Qian, Zhuoyang and Liu, Gaoge and Ling, Li and Zhang, Zhentao and Wu, Biao and Zhang, Shuo and Lu, Ke and Shi, Wei and Wang, Ziqi and others},
  journal={arXiv preprint arXiv:2601.20833},
  year={2026}
}

@article{asai2024openscholar,
  title={Openscholar: Synthesizing scientific literature with retrieval-augmented {LM}s},
  author={Asai, Akari and He, Jacqueline and Shao, Rulin and Shi, Weijia and Singh, Amanpreet and Chang, Joseph Chee and Lo, Kyle and Soldaini, Luca and Feldman, Sergey and D'arcy, Mike and others},
  journal={arXiv preprint arXiv:2411.14199},
  year={2024}
}

@article{hubert2025olympiad,
  title={Olympiad-level formal mathematical reasoning with reinforcement learning},
  author={Hubert, Thomas and Mehta, Rishi and Sartran, Laurent and Horv{\'a}th, Mikl{\'o}s Z and {\v{Z}}u{\v{z}}i{\'c}, Goran and Wieser, Eric and Huang, Aja and Schrittwieser, Julian and Schroecker, Yannick and Masoom, Hussain and others},
  journal={Nature},
  pages={1--3},
  year={2025},
  publisher={Nature Publishing Group UK London}
}

@article{zheng2021minif2f,
  title={Minif2f: a cross-system benchmark for formal olympiad-level mathematics},
  author={Zheng, Kunhao and Han, Jesse Michael and Polu, Stanislas},
  journal={arXiv preprint arXiv:2109.00110},
  year={2021}
}

@article{novikov2025alphaevolve,
  title={Alphaevolve: A coding agent for scientific and algorithmic discovery},
  author={Novikov, Alexander and V{\~u}, Ng{\^a}n and Eisenberger, Marvin and Dupont, Emilien and Huang, Po-Sen and Wagner, Adam Zsolt and Shirobokov, Sergey and Kozlovskii, Borislav and Ruiz, Francisco JR and Mehrabian, Abbas and others},
  journal={arXiv preprint arXiv:2506.13131},
  year={2025}
}

@article{li2025sciagent,
  title={SciAgent: A Unified Multi-Agent System for Generalistic Scientific Reasoning},
  author={Li, Xuchen and Wu, Ruitao and Liu, Xuanbo and Wang, Xukai and Hu, Jinbo and Bai, Zhixin and Zeng, Bohan and Liang, Hao and Chen, Leheng and Chen, Mingrui and others},
  journal={arXiv preprint arXiv:2511.08151},
  year={2025}
}

@article{li2025advancing,
  title={Advancing Mathematical Research via Human-{AI} Interactive Theorem Proving},
  author={Li, Chenyi and Lai, Zhijian and An, Dong and Hu, Jiang and Wen, Zaiwen},
  journal={arXiv preprint arXiv:2512.09443},
  year={2025}
}

@article{feng2026towards,
  title={Towards autonomous mathematics research},
  author={Feng, Tony and Trinh, Trieu H and Bingham, Garrett and Hwang, Dawsen and Chervonyi, Yuri and Jung, Junehyuk and Lee, Joonkyung and Pagano, Carlo and Kim, Sang-hyun and Pasqualotto, Federico and others},
  journal={arXiv preprint arXiv:2602.10177},
  year={2026}
}

@article{zheng2023judging,
  title={Judging {LLM}-as-a-judge with {MT}-{B}ench and chatbot arena},
  author={Zheng, Lianmin and Chiang, Wei-Lin and Sheng, Ying and Zhuang, Siyuan and Wu, Zhanghao and Zhuang, Yonghao and Lin, Zi and Li, Zhuohan and Li, Dacheng and Xing, Eric and others},
  journal={Advances in neural information processing systems},
  volume={36},
  pages={46595--46623},
  year={2023}
}

@article{chu2024pre,
  title={Pre: A peer review based large language model evaluator},
  author={Chu, Zhumin and Ai, Qingyao and Tu, Yiteng and Li, Haitao and Liu, Yiqun},
  journal={arXiv preprint arXiv:2401.15641},
  year={2024}
}

@inproceedings{idahl2025openreviewer,
  title={Openreviewer: A specialized large language model for generating critical scientific paper reviews},
  author={Idahl, Maximilian and Ahmadi, Zahra},
  booktitle={Proceedings of the 2025 Conference of the Nations of the Americas Chapter of the Association for Computational Linguistics: Human Language Technologies (System Demonstrations)},
  pages={550--562},
  year={2025}
}

@inproceedings{zhu2025deepreview,
  title={Deepreview: Improving {LLM}-based paper review with human-like deep thinking process},
  author={Zhu, Minjun and Weng, Yixuan and Yang, Linyi and Zhang, Yue},
  booktitle={Proceedings of the 63rd Annual Meeting of the Association for Computational Linguistics (Volume 1: Long Papers)},
  pages={29330--29355},
  year={2025}
}

@article{gao2025reviewagents,
  title={Reviewagents: Bridging the gap between human and {AI}-generated paper reviews},
  author={Gao, Xian and Ruan, Jiacheng and Zhang, Zongyun and Gao, Jingsheng and Liu, Ting and Fu, Yuzhuo},
  journal={arXiv preprint arXiv:2503.08506},
  year={2025}
}

@article{ye2024justice,
  title={Justice or prejudice? quantifying biases in {LLM}-as-a-judge},
  author={Ye, Jiayi and Wang, Yanbo and Huang, Yue and Chen, Dongping and Zhang, Qihui and Moniz, Nuno and Gao, Tian and Geyer, Werner and Huang, Chao and Chen, Pin-Yu and others},
  journal={arXiv preprint arXiv:2410.02736},
  year={2024}
}

@article{li2025llms,
  title={Llms cannot reliably judge (yet?): A comprehensive assessment on the robustness of {LLM}-as-a-judge},
  author={Li, Songze and Xu, Chuokun and Wang, Jiaying and Gong, Xueluan and Chen, Chen and Zhang, Jirui and Wang, Jun and Lam, Kwok-Yan and Ji, Shouling},
  journal={arXiv preprint arXiv:2506.09443},
  year={2025}
}

@article{ye2024we,
  title={Are we there yet? revealing the risks of utilizing large language models in scholarly peer review},
  author={Ye, Rui and Pang, Xianghe and Chai, Jingyi and Chen, Jiaao and Yin, Zhenfei and Xiang, Zhen and Dong, Xiaowen and Shao, Jing and Chen, Siheng},
  journal={arXiv preprint arXiv:2412.01708},
  year={2024}
}

@article{lou2024aaar,
  title={AAAR-1.0: Assessing {AI}'s Potential to Assist Research},
  author={Lou, Renze and Xu, Hanzi and Wang, Sijia and Du, Jiangshu and Kamoi, Ryo and Lu, Xiaoxin and Xie, Jian and Sun, Yuxuan and Zhang, Yusen and Ahn, Jihyun Janice and others},
  journal={arXiv preprint arXiv:2410.22394},
  year={2024}
}

@article{yang2026aris,
  title={ARIS: Autonomous Research via Adversarial Multi-Agent Collaboration},
  author={Yang, Ruofeng and Li, Yongcan and Li, Shuai},
  journal={arXiv preprint arXiv:2605.03042},
  year={2026}
}

@article{miao2025physmaster,
  title={PhysMaster: Building an Autonomous {AI} Physicist for Theoretical and Computational Physics Research},
  author={Miao, Tingjia and Dai, Jiawen and Liu, Jingkun and Tan, Jinxin and Zhang, Muhua and Jin, Wenkai and Du, Yuwen and Jin, Tian and Pang, Xianghe and Liu, Zexi and others},
  journal={arXiv preprint arXiv:2512.19799},
  year={2025}
}

@article{wang2026m2f,
  title={M2F: Automated Formalization of Mathematical Literature at Scale},
  author={Wang, Zichen and Ma, Wanli and Ming, Zhenyu and Zhang, Gong and Yuan, Kun and Wen, Zaiwen},
  journal={arXiv preprint arXiv:2602.17016},
  year={2026}
}

@article{ju2026automated,
  title={Automated Conjecture Resolution with Formal Verification},
  author={Ju, Haocheng and Gao, Guoxiong and Jiang, Jiedong and Wu, Bin and Sun, Zeming and Chen, Leheng and Wang, Yutong and Wang, Yuefeng and Wang, Zichen and He, Wanyi and others},
  journal={arXiv preprint arXiv:2604.03789},
  year={2026}
}

@article{chai2025scimaster,
  title={SciMaster: Towards General-Purpose Scientific {AI} Agents, Part I. X-Master as Foundation: Can We Lead on Humanity's Last Exam?},
  author={Chai, Jingyi and Tang, Shuo and Ye, Rui and Du, Yuwen and Zhu, Xinyu and Zhou, Mengcheng and Wang, Yanfeng and Zhang, Yuzhi and Zhang, Linfeng and Chen, Siheng and others},
  journal={arXiv preprint arXiv:2507.05241},
  year={2025}
}

@article{liu2025ml,
  title={Ml-master: Towards {AI}-for-{AI} via integration of exploration and reasoning},
  author={Liu, Zexi and Cai, Yuzhu and Zhu, Xinyu and Zheng, Yujie and Chen, Runkun and Wen, Ying and Wang, Yanfeng and Chen, Siheng and others},
  journal={arXiv preprint arXiv:2506.16499},
  year={2025}
}

@article{ma2025reliable,
  title={Reliable fine-grained evaluation of natural language math proofs},
  author={Ma, Wenjie and Cojocaru, Andrei and Kolhe, Neel and Louie, Bradley and Sharif, Robin Said and Zhang, Haihan and Zhuang, Vincent and Zaharia, Matei and Min, Sewon},
  journal={arXiv preprint arXiv:2510.13888},
  year={2025}
}

@article{an2026qed,
  title={QED: An Open-Source Multi-Agent System for Generating Mathematical Proofs on Open Problems},
  author={An, Chenyang and Ye, Qihao and Pan, Minghao and Zhang, Jiayaun},
  journal={arXiv preprint arXiv:2604.24021},
  year={2026}
}

@article{jakovetic2014fast,
  author    = {Du{\v{s}}an Jakoveti{\'c} and Jo{\~a}o Xavier and Jos{\'e} M. F. Moura},
  title     = {Fast Distributed Gradient Methods},
  journal   = {IEEE Transactions on Automatic Control},
  volume    = {59},
  number    = {5},
  pages     = {1131--1146},
  year      = {2014},
  doi       = {10.1109/TAC.2014.2298712}
}

@article{scaman2017optimal,
  author    = {Kevin Scaman and Francis Bach and S{\'e}bastien Bubeck and Yin Tat Lee and Laurent Massouli{\'e}},
  title     = {Optimal Algorithms for Smooth and Strongly Convex Distributed Optimization in Networks},
  journal   = {Proceedings of the 35th International Conference on Machine Learning},
  series    = {Proceedings of Machine Learning Research},
  volume    = {80},
  pages     = {3027--3036},
  year      = {2018}
}

@article{berthier2020accelerated,
  author    = {Rapha{\"e}l Berthier and Francis Bach and Pierre Gaillard},
  title     = {Accelerated Gossip in Networks of Given Dimension Using Jacobi Polynomial Iterations},
  journal   = {SIAM Journal on Mathematics of Data Science},
  volume    = {2},
  number    = {1},
  pages     = {24--47},
  year      = {2020},
  doi       = {10.1137/19M1244822}
}

@article{lin2025goedel,
  title={Goedel-prover: A frontier model for open-source automated theorem proving},
  author={Lin, Yong and Tang, Shange and Lyu, Bohan and Wu, Jiayun and Lin, Hongzhou and Yang, Kaiyu and Li, Jia and Xia, Mengzhou and Chen, Danqi and Arora, Sanjeev and others},
  journal={arXiv preprint arXiv:2502.07640},
  year={2025}
}

@article{lin2025goedelv2,
  title={Goedel-prover-v2: Scaling formal theorem proving with scaffolded data synthesis and self-correction},
  author={Lin, Yong and Tang, Shange and Lyu, Bohan and Yang, Ziran and Chung, Jui-Hui and Zhao, Haoyu and Jiang, Lai and Geng, Yihan and Ge, Jiawei and Sun, Jingruo and others},
  journal={arXiv preprint arXiv:2508.03613},
  year={2025}
}

@article{wang2025kimina,
  title={Kimina-prover preview: Towards large formal reasoning models with reinforcement learning},
  author={Wang, Haiming and Unsal, Mert and Lin, Xiaohan and Baksys, Mantas and Liu, Junqi and Santos, Marco Dos and Sung, Flood and Vinyes, Marina and Ying, Zhenzhe and Zhu, Zekai and others},
  journal={arXiv preprint arXiv:2504.11354},
  year={2025}
}

@article{chen2025seed,
  title={Seed-prover: Deep and broad reasoning for automated theorem proving},
  author={Chen, Luoxin and Gu, Jinming and Huang, Liankai and Huang, Wenhao and Jiang, Zhicheng and Jie, Allan and Jin, Xiaoran and Jin, Xing and Li, Chenggang and Ma, Kaijing and others},
  journal={arXiv preprint arXiv:2507.23726},
  year={2025}
}

@article{chen2025seedv1.5,
  title={Seed-prover 1.5: Mastering undergraduate-level theorem proving via learning from experience},
  author={Chen, Jiangjie and Chen, Wenxiang and Du, Jiacheng and Hu, Jinyi and Jiang, Zhicheng and Jie, Allan and Jin, Xiaoran and Jin, Xing and Li, Chenggang and Shi, Wenlei and others},
  journal={arXiv preprint arXiv:2512.17260},
  year={2025}
}

@inproceedings{xin2025bfs,
  title={Bfs-prover: Scalable best-first tree search for llm-based automatic theorem proving},
  author={Xin, Ran and Xi, Chenguang and Yang, Jie and Chen, Feng and Wu, Hang and Xiao, Xia and Sun, Yifan and Zheng, Shen and Ding, Ming},
  booktitle={Proceedings of the 63rd Annual Meeting of the Association for Computational Linguistics (Volume 1: Long Papers)},
  pages={32588--32599},
  year={2025}
}

@article{tsoukalas2024putnambench,
  title={Putnambench: Evaluating neural theorem-provers on the putnam mathematical competition},
  author={Tsoukalas, George and Lee, Jasper and Jennings, John and Xin, Jimmy and Ding, Michelle and Jennings, Michael and Thakur, Amitayush and Chaudhuri, Swarat},
  journal={Advances in Neural Information Processing Systems},
  volume={37},
  pages={11545--11569},
  year={2024}
}

@article{abouzaid2026first,
  title={First Proof},
  author={Abouzaid, Mohammed and Blumberg, Andrew J and Hairer, Martin and Kileel, Joe and Kolda, Tamara G and Nelson, Paul D and Spielman, Daniel and Srivastava, Nikhil and Ward, Rachel and Weinberger, Shmuel and others},
  journal={arXiv preprint arXiv:2602.05192},
  year={2026}
}

@article{phan2025humanity,
  title={Humanity's last exam},
  author={Phan, Long and Gatti, Alice and Han, Ziwen and Li, Nathaniel and Hu, Josephina and Zhang, Hugh and Zhang, Chen Bo Calvin and Shaaban, Mohamed and Ling, John and Shi, Sean and others},
  journal={arXiv preprint arXiv:2501.14249},
  year={2025}
}

@article{ren2025deepseek,
  title={Deepseek-prover-v2: Advancing formal mathematical reasoning via reinforcement learning for subgoal decomposition},
  author={Ren, ZZ and Shao, Zhihong and Song, Junxiao and Xin, Huajian and Wang, Haocheng and Zhao, Wanjia and Zhang, Liyue and Fu, Zhe and Zhu, Qihao and Yang, Dejian and others},
  journal={arXiv preprint arXiv:2504.21801},
  year={2025}
}

@article{xin2025scaling,
  title={Scaling up multi-turn off-policy rl and multi-agent tree search for llm step-provers},
  author={Xin, Ran and Zheng, Zeyu and Nie, Yanchen and Yuan, Kun and Xiao, Xia},
  journal={arXiv preprint arXiv:2509.06493},
  year={2025}
}

@article{wang2026longcat,
  title={LongCat-Flash-Prover: Advancing Native Formal Reasoning via Agentic Tool-Integrated Reinforcement Learning},
  author={Wang, Jianing and Zhang, Jianfei and Guo, Qi and Guo, Linsen and Li, Rumei and Zhang, Chao and Peng, Chong and Wang, Cunguang and Zhao, Dengchang and Shi, Jiarong and others},
  journal={arXiv preprint arXiv:2603.21065},
  year={2026}
}

@article{ju2026matlas,
  title={Matlas: A Semantic Search Engine for Mathematics},
  author={Ju, Haocheng and Chen, Leheng and Wu, Peihao and Dai, Bryan and Dong, Bin},
  journal={arXiv preprint arXiv:2604.17484},
  year={2026}
}

@article{he2026fedslop,
  title={FedSLoP: Memory-Efficient Federated Learning with Low-Rank Gradient Projection},
  author={He, Yutong and Huang, Zhengyang and Geng, Jiahe},
  journal={arXiv preprint arXiv:2604.24012},
  year={2026},
  note={The unpolished ReasFlow-generated manuscript, prior to human-expert polishing, is available at \url{https://blog.reaslab.io/papers/[blog]FedSLoP.pdf}.}
}

@article{zhu2026subspace,
  title={Subspace Optimization for Efficient Federated Learning under Heterogeneous Data},
  author={Zhu, Shuchen and Huang, Zhengyang and Xu, Yuqi and Li, Peijin},
  journal={arXiv preprint arXiv:2604.25467},
  year={2026},
  note={The unpolished ReasFlow-generated manuscript, prior to human-expert polishing, is available at \url{https://blog.reaslab.io/papers/[blog]Subspace-Scaffold.pdf}.}
}

@article{zhang2026suda,
  title={SUDA-Muon: Structural Design Principles and Boundaries for Fully Decentralized Muon},
  author={Zhang, Hengrui and Kong, Boao and Geng, Jiahe and Huang, Zhengyang},
  journal={arXiv preprint arXiv:2604.23980},
  year={2026},
  note={The unpolished ReasFlow-generated manuscript, prior to human-expert polishing, is available at \url{https://blog.reaslab.io/papers/[blog]Suda-Muon.pdf}.}
}

@article{li2026retraction,
  title={A Retraction-Free EXTRA Method for Decentralized Optimization on the Stiefel Manifold},
  author={Li, Shu and Hu, Jiang},
  journal={arXiv preprint arXiv:2604.23754},
  year={2026},
  note={The unpolished ReasFlow-generated manuscript, prior to human-expert polishing, is available at \url{https://blog.reaslab.io/papers/[blog]Retractionfree-EXTRA.pdf}.}
}

@article{chen2026mlr,
  title={Mlr-bench: Evaluating ai agents on open-ended machine learning research},
  author={Chen, Hui and Xiong, Miao and Lu, Yujie and Han, Wei and Deng, Ailin and He, Yufei and Wu, Jiaying and Li, Yibo and Liu, Yue and Hooi, Bryan},
  journal={Advances in Neural Information Processing Systems},
  volume={38},
  year={2026}
}

@article{sun2026accelerated,
      title={Accelerated Decentralized Stochastic Gradient Descent for Strongly Convex Optimization}, 
      author={Sun, Ming and Yuan, Kun},
      year={2026},
      journal={arXiv preprint arXiv:2606.07496},
      note={The unpolished ReasFlow-generated manuscript, prior to human-expert polishing, is available at \url{https://blog.reaslab.io/papers/[blog]MC-ADSGD.pdf}.}
}
\bibliographystyle{icml2026}
\newpage
\appendix
\onecolumn

\section{Detailed Agent Specifications}\label{app:agent_specs}
This appendix catalogs the list of tools of each specialized agent in ReasFlow, as referenced in Sec.~\ref{sec:agent_spec}. 

\subsection{Survey Agent}
\textbf{Tools:}
\begin{enumerate}
	\item \texttt{BasicTools}: Read, write, edit files; list directory contents.
	\item \texttt{OutlineWriterTool}: Writes a structured survey outline.
	\item \texttt{SurveyWriterTool}: Queries the built-in paper database and compiles a survey report.
	\item \texttt{RelatedWorkWriterTool}: Drafts the \textit{Related Work} section with \texttt{bibtex} entries.
	\item \texttt{LiteratureSearchTool}: Searches for papers beyond the internal database (e.g., via Semantic Scholar).
	\item \texttt{PaperLibrarySearchTool}: Searches the built-in FAISS vector database for semantically relevant knowledge cards.
	\item \texttt{LiteratureDetailTools}: Retrieves individual paper metadata, citation networks, and reference lists.
	\item \texttt{ReviewTool}: Evaluates the drafted \textit{Related Work} along four dimensions (format integrity, citation relevance, key reference coverage, instruction following) and returns a scored report.
	\item \texttt{BibtexFormattingSkill}: Provides instructions for generating correct \texttt{bibtex} entries.
\end{enumerate}

\subsection{Algorithm Agent}
\textbf{Tools:}
\begin{enumerate}
	\item \texttt{BasicTools}
	\item \texttt{KnowledgeCardSearch}: Searches knowledge cards for relevant algorithmic structures.
	\item \texttt{LatexWriterAgent}: Sub-agent for iterative pseudocode and report generation in \LaTeX.
	\item \texttt{TerminalCommandTool}: Executes shell commands.
	\item \texttt{SmartPlotTool}: Generates diagnostic figures (e.g., loss curves) from logs.
	\item \texttt{ImageReaderTool}: Verifies visual results via a multimodal LLM.
\end{enumerate}

\subsection{Prover Agent}
\textbf{Tools:}
\begin{enumerate}
	\item \texttt{BasicTools}
	\item \texttt{RefDownloaderSkill}: Instructs the agent on downloading LaTeX source from arXiv.
	\item \texttt{MathContentSearch}: Locates relevant mathematical content across references.
	\item \texttt{CardGeneratorSkill}: Extracts proof techniques and summarizes them into knowledge cards.
	\item \texttt{KnowledgeSearchTool}: Searches for related knowledge cards.
	\item \texttt{CardRetrieverSkill}: Instructs the agent on iterative card retrieval.
	\item \texttt{LemmaProverSkill}: Manages sub-agent sessions for lemma proving.
	\item \texttt{LemmaProverAgent}: Sub-agent that generates lemma proofs.
	\item \texttt{LemmaVerifierAgent}: Sub-agent that checks correctness of drafted proofs.
\end{enumerate}

\subsection{Experiment Agent}
\textbf{Tools:}
\begin{enumerate}
	\item \texttt{BasicTools}
	\item \texttt{KnowledgeCardSearch}: Retrieves dataset/task recommendations and baseline pseudocode.
	\item \texttt{TerminalCommandTool}
	\item \texttt{AutoTuningTool}: Performs automatic hyperparameter optimization.
	\item \texttt{SmartPlotTool}: Generates publication-quality figures.
	\item \texttt{PlotAnalysisTool}: Analyzes figures for clarity and anomalies.
	\item \texttt{LatexWriterAgent}: Assists in composing the final experiment report.
\end{enumerate}

\subsection{Introduction Agent}
\textbf{Tools:}
\begin{enumerate}
	\item \texttt{BasicTools}
    \item \texttt{ExtractionTools}: Tools including \texttt{ExtractSurveyInfo}, \texttt{ExtractMethodInfo}, \texttt{ExtractExperimentInfo}, \texttt{ExtractTheoryInfo}, and \texttt{OrganizeExtractedInfo} that generate structured section summaries.
	\item \texttt{IntroductionWriterTool}: Writes the \textit{Introduction} section in LaTeX with inline hallucination annotations flagging unsupported claims. Supports three writing styles (ML, Math, Default).
    \item \texttt{LiteratureSearchTool}:  A suite of tools including \texttt{LiteratureGetPaper}, \texttt{LiteratureGetCitations} and \texttt{LiteratureGetReferences} that retrieve information via Google Scholar and Semantic Scholar.
	\item \texttt{BibtexFormattingSkill}
    \item \texttt{EvalIntroduction}: Evaluates the generated introduction on coverage, faithfulness and context quality.
    \item \texttt{RefineIntroduction}: Performs targeted local refinement based on the evaluation report.
\end{enumerate}

\subsection{Writing Agent}

The \texttt{WritingAgent} uses a two-tier Coordinator + SubWriter architecture.

\medskip
\noindent\textbf{Coordinator Tools:}
\begin{enumerate}
	\item \texttt{BasicTools}
	\item \texttt{TerminalCommandTool}
	\item \texttt{LatexCompileTool}: Compiles the LaTeX project and returns structured logs.
	\item \texttt{ScreenshotAnalysisTool}: Renders PDF pages and inspects them via a vision LLM.
	\item \texttt{DispatchTools} (\texttt{DispatchTask}, \texttt{wait\_for\_next}): Asynchronously launches SubWriter instances and collects their results.
	\item \texttt{ReviewCompletenessTool}: Runs 4 parallel LLM checks (structure, reference verification, visual layout, compilation) + asset citation coverage, returns a fix-task list.
	\item \texttt{ReviewWritingQualityTool}: Multi-model parallel writing review + meta-review aggregation with venue-aware criteria.
\end{enumerate}

\noindent\textbf{SubWriter Tools:}
\begin{enumerate}
	\item \texttt{BasicTools}, \texttt{TerminalCommandTool}, \texttt{LatexCompileTool}, \texttt{ScreenshotAnalysisTool} (same as Coordinator)
	\item \texttt{GenerateMethodFigureTool}: Multi-agent image pipeline (Retriever $\rightarrow$ Planner $\rightarrow$ Stylist $\rightarrow$ Visualizer $\rightarrow$ Critic) for structural diagrams.
\end{enumerate}

\section{Experimental Details}\label{app:exp-detail}
\subsection{Experimental Details for Sec.~\ref{subsec:exp-paper}}
\subsubsection{Generation Details}
In this subsection, we clarify the prompt used to generate the papers for comparison, which is shown in Fig.~\ref{fig:prompt-paper1}.

\begin{center}
	\begin{LLMBox}[width=\textwidth, fontupper=\small]{Prompt for Full Paper Generation}
		FedAvg is a classic distributed optimization algorithm in federated learning. Recently, subspace optimization methods such as GoLore have sparked a wave of research on memory-efficient large-scale distributed algorithms, which reduce memory usage by projecting gradients onto a low-rank subspace, allowing optimizer states (e.g., the first- and second-order moments in Adam) to be stored in a low-rank memory-saving version. Consider designing the algorithm SFedAvg-GoLore, which introduces the subspace technique from GoLore into FedAvg. It is required to consider only local momentum, without global momentum, and to switch the subspace and reset the momentum to zero after each communication round. Please conduct academic research and writing focusing on SFedAvg-GoLore.\\

		References:\\
		$-$ $<$https://arxiv.org/abs/2101.11203$>$: Proof of FedAvg\\
		$-$ $<$https://arxiv.org/abs/2410.11289$>$: Convergence of subspace algorithms
	\end{LLMBox}
	\captionof{figure}{Prompt for automatic paper generation in Sec.~\ref{subsec:exp-paper}.}
	\label{fig:prompt-paper1}
\end{center}

\subsubsection{Evaluation Details}
In this subsection, we provide detailed evaluation metrics and results for evaluating the generated papers in Sec.~\ref{subsec:exp-paper}. As discussed, we design nine critical scholarly aspects. For each aspect, we design finer dimensions for assessment. Below we elaborate the detailed prompts for each aspect.

\noindent\textbf{A. Mathematical rigor. }For mathematical rigor, we consider five dimensions including: (i) academic formalism and formatting (10 points), (ii) target alignment and relevance (10 points), (iii) consistency of assumptions and notation (10 points), (iv) logical deduction and correctness (50 points), and (v) theoretical significance and strength (20 points). The prompt is as listed in Fig.~\ref{fig:prompt-paper-A}. Corresponding evaluation results are listed in Table~\ref{tab:paper_A_scores_detail}.

\begin{center}
	\begin{LLMBox}[width=\textwidth, fontupper=\small]{System prompt for evaluating \textit{Mathematical rigor}.}
		\begin{markdown}
You are the **Mathematical Rigor Reviewer**. Evaluate proofs on five dimensions (total 100 pts). Any fatal logical gap can invalidate the proof.<br>

Dimension rubric
- Dim1 Academic Formalism \\& Formatting (10): Standard proof structure (Lemma/Theorem/Proof), proper LaTeX/math. If wall-of-text with no math or no proof framework, give 0 here and strongly penalize total.
- Dim2 Target Alignment \\& Relevance (10): Conclusions, loss functions, and update steps align exactly with the stated problem/setting. If irrelevant (e.g., proving SGD when FL requested, assuming convexity in non-convex setting), set this to 0 and cap total $\le$ 30.
- Dim3 Consistency of Assumptions \\& Notation (10): Only provided assumptions used; notation defined and consistent. Each undeclared strong assumption deduct 5; undefined/inconsistent symbols deduct 1–3 each.
- Dim4 Logical Deduction \\& Correctness (50, core): Line-by-line check.
* Trivial skips: standard algebra/inequalities → no deduction.
* Suspicious skips: non-trivial jumps or “it is easy to see/similarly” without steps → deduct 5–10 each; if many, Dim4 $\le$ 20.
* Explicit errors/fake proofs: wrong inequality direction, illegal expectation moves, dimension mismatch, fabricated steps → any single one sets Dim4=0 and total $\le$ 40.
- Dim5 Theoretical Significance \\& Strength (20): Strongest conclusions with weakest assumptions. Reward tight rates/general cases; penalize “trivializing” extra assumptions that make the problem easy.

Scoring rules
- Use 0–100 mapped by weights above (10/10/10/50/20). Apply caps noted in Dim2/Dim4 explicit errors.
- Be evidence-based; cite specific formulas/steps from the provided text only.

Output
- Return **JSON only** per the given schema (sub\\_scores + summary). Do not add fields.
\end{markdown}
	\end{LLMBox}
	\captionof{figure}{System prompt for evaluating mathematical rigor.}
	\label{fig:prompt-paper-A}
\end{center}

\begin{table}[htbp]
	\centering
	\caption{Detailed Scores for \textbf{mathematical rigor}. S1, S2, S3 corresponds to evaluations performed with \texttt{GPT-5.1}, \texttt{GPT-5.4} and \texttt{Gemini-3-Pro} as the reviewer's LLM backend, respectively. Full scores for each dimension are 100. The final score is the weighted average of all dimensions. \textbf{Bold}: best; \underline{underlined}: second best.}
	\label{tab:paper_A_scores_detail}
	\newcolumntype{Z}{>{\centering\arraybackslash}p{1.4em}}
	\resizebox{\textwidth}{!}{
		\begin{tabular}{l|ZZZ|ZZZ|ZZZ|ZZZ|ZZZ|ZZZ}
			\toprule
			\multirow{2}{*}{AI Agent}    & \multicolumn{3}{c|}{Dim 1 Score} & \multicolumn{3}{c|}{Dim 2 Score} & \multicolumn{3}{c|}{Dim 3 Score} & \multicolumn{3}{c|}{Dim 4 Score} & \multicolumn{3}{c|}{Dim 5 Score} & \multicolumn{3}{c}{Final Score}                                                                    \\
			\cmidrule(lr){2-4} \cmidrule(lr){5-7} \cmidrule(lr){8-10} \cmidrule(lr){11-13} \cmidrule(lr){14-16} \cmidrule(lr){17-19}
			& S1 & S2 & S3 & S1  & S2 & S3 & S1 & S2 & S3 & S1 & S2 & S3 & S1 & S2 & S3 & S1 & S2 & S3 \\
			\midrule
			ChatGPT (\texttt{GPT-5.3}) & 70 & 50 & 10 & \underline{90} & 75 & 80 & 82 & 45 & 70 & 60 & 15 & 0 & 65 & 40 & 20 & 67.2 & 32.5 & 20 \\
			ChatGPT Pro (\texttt{GPT-5.4-Pro}) & 90 & \underline{88} & 50 & \underline{90} & \underline{82} & \underline{90} & \textbf{92} & 76 & \underline{95}  & \underline{85} & 28 & 15 & 78 & 45 & 40 & 85.3 & 47.6 & 39 \\
			Gemini (\texttt{Gemini-3.1-Pro}) & 70 & 35 & 0 & 85 & 70 & 80 & 80 & 35 & 80  & 60 & 10 & 0  & 65 & 35 & 0  & 66.5 & 26 & 16 \\
			Claude Pro (\texttt{Claude Opus 4.6}) & \textbf{95} & 78 & \underline{95} & \textbf{95} & 72 & \textbf{95} & 82 & 20 & \underline{95} & \textbf{88} & 0  & \underline{90} & \textbf{90} & 25 & \textbf{90} & \textbf{89.2} & 22 & \underline{91.5} \\
			\texttt{CycleResearcher-ML-12B} & 50 & 45 & 20 & 70 & 20 & 10 & 30 & 10 & 0 & 20 & 0  & 0 & 40 & 15 & 0 & 33 & 10.5 & 3 \\
			DeepScientist (\texttt{GPT-5.4}) & 85 & 86 & 0 & \underline{90} & \textbf{88} & 80 & 85 & \underline{78} & 80 & 80 & \underline{42} & 0 & 70 & 22 & 0 & 80 & 50.6 & 16 \\
			AI Scientist-v2 (\texttt{GPT-5.4}) & 60 & 15 & 0 & 85 & 55 & 60 & 75 & 45 & 30 & 40 & 5 & 0 & 60 & 10 & 0 & 54 & 16 & 9 \\
			ARIS (\texttt{GPT-5.4}) & 75 & 85 & 20 & \underline{90} & \underline{82} & \underline{90} & 80 & \underline{78} & 90  & 55 & 35 & 0  & 70 & \underline{55} & 10 & 66 & \underline{53} & 22 \\
            \midrule
			ReasFlow (\texttt{GPT-5.4}) & \underline{92} & \textbf{90} & \textbf{100} & \underline{90} & \textbf{88} & \textbf{95} & \underline{88} & \textbf{82} & \textbf{100} & \underline{85} & \textbf{76} & \textbf{95} & \underline{80} & \textbf{68} & \underline{85} & \underline{85.5} & \textbf{77.6} & \textbf{94} \\
			\bottomrule
		\end{tabular}
	}
\end{table}

\noindent\textbf{B. Originality and significance. }For originality and significance, we consider three dimensions including: (i) method novelty (40 points), (ii) problem importance (30 points), and (iii) differentiation (30 points). The prompt is as listed in Fig.~\ref{fig:prompt-paper-B}. Corresponding evaluation results are listed in Table~\ref{tab:paper_B_scores_detail}.

\begin{center}
	\begin{LLMBox}[width=\textwidth, fontupper=\small]{System prompt for evaluating \textit{Originality and significance}.}
		\begin{markdown}
You are the **Originality \\& Significance Reviewer**.

Dimension rubric
- Dim1 Method Novelty (40): substantive technical innovation vs trivial tweak; clarity of the new idea.
- Dim2 Problem Importance (30): relevance and impact of the problem setting in math/application.
- Dim3 Differentiation (30): explicit, evidence-backed comparison to closest prior work; stated improvements.

Evaluation steps
1) Extract the core idea/algorithm; enumerate what is genuinely new.
2) Align against closest prior art listed; check material differences and evidence.
3) Judge importance of the problem/benchmark/context and stated impact.

Scoring rules
- Inflated novelty with no evidence → heavy deduct Dim1.
- Weak/contrived problem → deduct Dim2.
- Missing/vague comparison → deduct Dim3 (cap total at 40 if over-claiming).
- Evidence only from provided text.

Output
- JSON only with the provided schema.
\end{markdown}
	\end{LLMBox}
	\captionof{figure}{System prompt for evaluating originality and significance.}
	\label{fig:prompt-paper-B}
\end{center}

\begin{table}[htbp]
	\centering
	\caption{Detailed Scores for \textbf{originality and significance}. S1, S2, S3 corresponds to evaluations performed with \texttt{GPT-5.1}, \texttt{GPT-5.4} and \texttt{Gemini-3-Pro} as the reviewer's LLM backend, respectively. Full scores for each dimension are 100. The final score is the weighted average of all dimensions. \textbf{Bold}: best; \underline{underlined}: second best.}
	\label{tab:paper_B_scores_detail}
	\newcolumntype{Z}{>{\centering\arraybackslash}p{1.4em}}
	\begin{tabular}{l|ZZZ|ZZZ|ZZZ|ZZZ}
		\toprule
		\multirow{2}{*}{AI Agent} & \multicolumn{3}{c|}{Dim 1 Score} & \multicolumn{3}{c|}{Dim 2 Score} & \multicolumn{3}{c|}{Dim 3 Score} & \multicolumn{3}{c}{Final Score} \\
		\cmidrule(lr){2-4} \cmidrule(lr){5-7} \cmidrule(lr){8-10} \cmidrule(lr){11-13} & S1 & S2 & S3 & S1 & S2 & S3 & S1 & S2 & S3 & S1 & S2   & S3 \\
		\midrule
		ChatGPT (\texttt{GPT-5.3}) & 70 & 45 & 50 & \underline{85} & 75 & \underline{80} & 45 & 30 & 20 & 67 & 49.5 & 50 \\
		ChatGPT Pro (\texttt{GPT-5.4-Pro}) & 75 & 52 & 45 & \underline{85} & 74 & \underline{80} & 80 & 58 & 75 & 79.5 & 60.4 & 64.5 \\
		Gemini (\texttt{Gemini-3.1-Pro}) & 65 & 45 & \underline{60}  & \underline{85} & 75 & \underline{80} & 55 & 30 & \underline{80} & 68 & 49.5 & \underline{72}   \\
		Claude Pro (\texttt{Claude Opus 4.6}) & \textbf{80} & 45  & \textbf{85} & \textbf{90} & 75 & \textbf{90} & \underline{85} & 54 & \textbf{90} & \textbf{84.5} & 56.7 & \textbf{88} \\
		\texttt{CycleResearcher-ML-12B} & 45 & 20 & 20 & 75 & 70 & 60 & 25 & 10 & 10 & 48 & 32 & 29 \\
		DeepScientist (\texttt{GPT-5.4}) & 60 & 45 & 10  & 80 & 72 & 70 & 75 & 68 & \underline{80} & 70.5 & 60 & 49   \\
		AI Scientist-v2 (\texttt{GPT-5.4}) & 75 & 55 & 40 & 80 & 72 & \underline{80} & 65 & 50 & 40 & 73.5 & 58.6 & 52 \\
		ARIS (\texttt{GPT-5.4}) & \underline{78} & \textbf{72} & \underline{60} & \underline{85} & \textbf{78} & \underline{80} & 80 & \textbf{80} & \underline{80} & 80.7 & \textbf{76.2} & \underline{72} \\
        \midrule
		ReasFlow (\texttt{GPT-5.4}) & \underline{78} & \underline{58} & \textbf{85} & \underline{85} & \underline{76} & \textbf{90} & \textbf{90} & \underline{72} & \textbf{90} & \underline{83.7} & \underline{67.6} & \textbf{88} \\
		\bottomrule
	\end{tabular}
\end{table}

\noindent\textbf{C. Numerical experiments. }For numerical experiments, we consider five dimensions including: (i) sufficiency (25 points), (ii) baseline fairness (25 points), (iii) parameter specification (15 points), (iv) reproducibility (20 points), and (v) statistical reliability (15 points). The prompt is as listed in Fig.~\ref{fig:prompt-paper-C}. Corresponding evaluation results are listed in Table~\ref{tab:paper_C_scores_detail}.

\begin{center}
	\begin{LLMBox}[width=\textwidth, fontupper=\small]{System prompt for evaluating \textit{Numerical experiments}.}
		\begin{markdown}
You are the **Numerical Experiments Reviewer**.

Dimension rubric
- Dim1 Sufficiency (25): coverage of typical + edge cases; multiple difficulty levels.
- Dim2 Baseline Fairness (25): fair comparison vs SOTA/classical baselines with matched settings/metrics.
- Dim3 Parameter Specification (15): all key params (grid/step/tol/seeds) stated.
- Dim4 Reproducibility (20): rerun detail—code/links/splits/seeds/ablations.
- Dim5 Statistical Reliability (15): mean/std over multiple runs when stochastic; variance reported.

Evaluation steps
1) List tasks/datasets/cases; judge diversity and relevance.
2) Inspect baselines and fairness (same splits, metrics, tuning budgets).
3) Collect hyper/solver params; note missing ones.
4) Check for code/links/seeds/ablations; assess rerun feasibility.
5) For stochastic settings, verify repeats and variance reporting.

Scoring rules
- Single easy case or missing baselines → large deduct Dim1/Dim2.
- Missing key params/seeds → deduct Dim3/Dim4.
- No repeats in stochastic setting → deduct Dim5 (or 70 “not applicable” if deterministic).
- Fabricated-looking tables or impossible gains → cap total at 30 pending credibility.

Output
- JSON only with given schema.
\end{markdown}
	\end{LLMBox}
	\captionof{figure}{System prompt for evaluating numerical experiments.}
	\label{fig:prompt-paper-C}
\end{center}

\begin{table}[htbp]
	\centering
	\caption{Detailed Scores for \textbf{numerical experiments}. S1, S2, S3 corresponds to evaluations performed with \texttt{GPT-5.1}, \texttt{GPT-5.4} and \texttt{Gemini-3-Pro} as the reviewer's LLM backend, respectively. Full scores for each dimension are 100. The final score is the weighted average of all dimensions. \textbf{Bold}: best; \underline{underlined}: second best.}
	\label{tab:paper_C_scores_detail}
	\newcolumntype{Z}{>{\centering\arraybackslash}p{1.4em}}
	\resizebox{\textwidth}{!}{
		\begin{tabular}{l|ZZZ|ZZZ|ZZZ|ZZZ|ZZZ|ZZZ}
			\toprule
			\multirow{2}{*}{AI Agent}    & \multicolumn{3}{c|}{Dim 1 Score} & \multicolumn{3}{c|}{Dim 2 Score} & \multicolumn{3}{c|}{Dim 3 Score} & \multicolumn{3}{c|}{Dim 4 Score} & \multicolumn{3}{c|}{Dim 5 Score} & \multicolumn{3}{c}{Final Score} \\
			\cmidrule(lr){2-4} \cmidrule(lr){5-7} \cmidrule(lr){8-10} \cmidrule(lr){11-13} \cmidrule(lr){14-16} \cmidrule(lr){17-19} & S1 & S2 & S3 & S1 & S2 & S3 & S1 & S2 & S3 & S1 & S2 & S3 & S1 & S2 & S3 & S1 & S2 & S3 \\
			\midrule
			ChatGPT (\texttt{GPT-5.3}) & 5 & 10 & 0 & 0 & 5 & 0 & 30 & 10 & 0 & 10 & 10 & 0 & 20 & 5 & \underline{0} & 10.8 & 8 & 0 \\
			ChatGPT Pro (\texttt{GPT-5.4-Pro}) & 5 & 5 & 0 & 0 & 0 & 0 & 30 & 10 & 0 & 20 & 5 & 0 & 10 & 0 & \underline{0} & 11.3 & 3.8 & 0 \\
			Gemini (\texttt{Gemini-3.1-Pro}) & 5 & 0 & 0 & 0 & 0 & 0 & 0 & 10 & 0 & 20 & 5 & 0 & 0 & 0 & \underline{0} & 5.3 & 2.5 & 0 \\
			Claude Pro (\texttt{Claude Opus 4.6}) & 5 & 0 & 0 & 0 & 0 & 0 & 25 & 5 & 0 & \underline{30} & 0 & 0  & 0 & 0 & \underline{0} & 11 & 0.8 & 0 \\
			\texttt{CycleResearcher-ML-12B} & 15 & 15 & 0 & 20 & 10 & 0 & 20 & 10 & 0 & 15 & 5 & 0  & 10 & 0 & \underline{0} & 16.3 & 8.8 & 0 \\
			DeepScientist (\texttt{GPT-5.4}) & 55 & 28 & 10 & \textbf{80} & \textbf{63} & \underline{60} & \textbf{70} & \textbf{74} & \textbf{90} & \textbf{60} & \textbf{42} & \underline{20} & \underline{30} & \underline{18} & \underline{0} & \underline{60.8} & \underline{45} & \underline{35} \\
			AI Scientist-v2 (\texttt{GPT-5.4}) & \underline{60} & \underline{30} & \underline{30} & 65 & 45 & 40 & 30 & 15 & 0 & \underline{30} & 10 & 0 & 25 & 10 & \underline{0} & 45.5 & 24.5 & 17.5 \\
			ARIS (\texttt{GPT-5.4}) & 10 & 15 & 0 & 15 & 25 & 0 & 30 & 30 & 0 & 20 & 15 & 0 & 10 & 10 & \underline{0} & 16.3 & 19 & 0 \\
            \midrule
			ReasFlow (\texttt{GPT-5.4}) & \textbf{70} & \textbf{35} & \textbf{70} & \underline{72} & \underline{52} & \textbf{85} & \underline{65} & \underline{66} & \underline{75} & \textbf{60} & \underline{38} & \textbf{60} & \textbf{80} & \textbf{78} & \textbf{90} & \textbf{69.3} & \textbf{51} & \textbf{75.5} \\
			\bottomrule
		\end{tabular}
	}
\end{table}

\noindent\textbf{D. Theory-experiment consistency. }For originality and significance, we consider three dimensions including: (i) convergence order verification (40 points), (ii) error bound verification (35 points), and (iii) discrepancy explanation (25 points). The prompt is as listed in Fig.~\ref{fig:prompt-paper-D}. Corresponding evaluation results are listed in Table~\ref{tab:paper_D_scores_detail}.

\begin{center}
	\begin{LLMBox}[width=\textwidth, fontupper=\small]{System prompt for evaluating \textit{Theory-experiment consistency}.}
		\begin{markdown}
You are the **Theory–Experiment Consistency Reviewer**.

Dimension rubric
- Dim1 Convergence Order Verification (40): theoretical $O(h^p)$ vs empirical rates.
- Dim2 Error Bound Verification (35): theoretical error bounds observed in results.
- Dim3 Discrepancy Explanation (25): reasonable, specific explanations for mismatches.

Evaluation steps
1) Extract theoretical rates/bounds and required conditions.
2) Check convergence tables/plots for matching slopes; note mismatches.
3) If mismatches, assess explanations for adequacy.

Scoring rules
- Missing empirical verification → deduct Dim1/Dim2.
- Unexplained discrepancies → heavy deduct Dim3; may cap total at 40.
- If no theory, but empirical convergence shown, score Dim1/Dim2 on evidence and note absence of theory.

Output
- JSON only per schema.
\end{markdown}
	\end{LLMBox}
	\captionof{figure}{System prompt for evaluating theory-experiment consistency.}
	\label{fig:prompt-paper-D}
\end{center}

\begin{table}[htbp]
	\centering
	\caption{Detailed Scores for \textbf{theory-experiment consistency}. S1, S2, S3 corresponds to evaluations performed with \texttt{GPT-5.1}, \texttt{GPT-5.4} and \texttt{Gemini-3-Pro} as the reviewer's LLM backend, respectively. Full scores for each dimension are 100. The final score is the weighted average of all dimensions. \textbf{Bold}: best; \underline{underlined}: second best.}
	\label{tab:paper_D_scores_detail}
	\newcolumntype{Z}{>{\centering\arraybackslash}p{1.4em}}
	\begin{tabular}{l|ZZZ|ZZZ|ZZZ|ZZZ}
		\toprule
		\multirow{2}{*}{AI Agent} & \multicolumn{3}{c|}{Dim 1 Score} & \multicolumn{3}{c|}{Dim 2 Score} & \multicolumn{3}{c|}{Dim 3 Score} & \multicolumn{3}{c}{Final Score} \\
		\cmidrule(lr){2-4} \cmidrule(lr){5-7} \cmidrule(lr){8-10} \cmidrule(lr){11-13}& S1 & S2 & S3 & S1 & S2 & S3 & S1 & S2 & S3 & S1 & S2 & S3 \\
		\midrule
		ChatGPT (\texttt{GPT-5.3}) & 10 & 5 & \underline{0} & 10 & 5  & \underline{0} & 40 & 10 & \underline{0} & 17.5 & 6.3 & \underline{0} \\
		ChatGPT Pro (\texttt{GPT-5.4-Pro}) & 10 & 5 & \underline{0}   & 10 & 5 & \underline{0} & 15 & 10 & \underline{0} & 11.3 & 6.3 & \underline{0} \\
		Gemini (\texttt{Gemini-3.1-Pro}) & 5 & 0 & \underline{0} & 0 & 0  & \underline{0} & 40 & 5 & \underline{0} & 12 & 1.3 & \underline{0} \\
		Claude Pro (\texttt{Claude Opus 4.6}) & 10 & 0 & \underline{0} & 10 & 0 & \underline{0} & 40 & 9 & \underline{0} & 17.5 & 2.3 & \underline{0} \\
		\texttt{CycleResearcher-ML-12B} & 15 & 5 & \underline{0} & 10 & 5 & \underline{0} & 20 & 0 & \underline{0} & 14.5 & 3.8 & \underline{0} \\
		DeepScientist (\texttt{GPT-5.4}) & 20 & \underline{15} & \underline{0} & 20 & \underline{10} & \underline{0} & \underline{60} & \textbf{62} & \underline{0} & 30 & \underline{25} & \underline{0} \\
		AI Scientist-v2 (\texttt{GPT-5.4}) & \underline{30} & 5 & \underline{0}  & \underline{25} & 5 & \underline{0} & 50 & 10 & \underline{0} & \underline{33.3} & 6.3 & \underline{0} \\
		ARIS (\texttt{GPT-5.4}) & 15 & 10 & \underline{0} & 15 & \underline{10} & \underline{0} & 40 & 15 & \underline{0} & 21.3 & 11.3 & \underline{0} \\
        \midrule
		ReasFlow (\texttt{GPT-5.4}) & \textbf{50} & \textbf{18} & \textbf{70} & \textbf{50} & \textbf{15} & \textbf{60} & \textbf{85} & \underline{58} & \textbf{90} & \textbf{58.8} & \textbf{27} & \textbf{71.5} \\
		\bottomrule
	\end{tabular}
\end{table}

\noindent\textbf{E. Writing and presentation. }For numerical experiments, we consider five dimensions including: (i) structure completeness (20 points), (ii) abstract quality (15 points), (iii) notation consistency (25 points), (iv) language quality (20 points), and (v) definitions and notation (20 points). The prompt is as listed in Fig.~\ref{fig:prompt-paper-E}. Corresponding evaluation results are listed in Table~\ref{tab:paper_E_scores_detail}.

\begin{center}
	\begin{LLMBox}[width=\textwidth, fontupper=\small]{System prompt for evaluating \textit{Writing and presentation}.}
		\begin{markdown}
You are the **Writing \\& Presentation Reviewer**.

Dimension rubric
- Dim1 Structure Completeness (20): standard flow Intro → Method → Analysis → Experiments → Conclusion.
- Dim2 Abstract Quality (15): accurate summary of problem, method, main results.
- Dim3 Notation Consistency (25): symbols defined before use; consistent.
- Dim4 Language Quality (20): grammar/spelling; clarity.
- Dim5 Definitions \\& Notation (20): key terms/notation clearly defined before first use.

Evaluation steps
1) Check abstract and section flow; note missing parts.
2) Track definitions/notation; spot inconsistencies/overloading.
3) Note language issues that impede understanding.

Scoring rules
- Missing sections/undefined notation → deduct Dim1/Dim3/Dim5.
- Ambiguous or overloaded symbols → deduct Dim3/Dim5.
- Poor grammar hurting clarity → deduct Dim4.

Output
- JSON only per schema.
\end{markdown}
	\end{LLMBox}
	\captionof{figure}{System prompt for evaluating writing and presentation.}
	\label{fig:prompt-paper-E}
\end{center}

\begin{table}[htbp]
	\centering
	\caption{Detailed Scores for \textbf{writing and presentation}. S1, S2, S3 corresponds to evaluations performed with \texttt{GPT-5.1}, \texttt{GPT-5.4} and \texttt{Gemini-3-Pro} as the reviewer's LLM backend, respectively. Full scores for each dimension are 100. The final score is the weighted average of all dimensions. \textbf{Bold}: best; \underline{underlined}: second best.}
	\label{tab:paper_E_scores_detail}
	\newcolumntype{Z}{>{\centering\arraybackslash}p{1.4em}}
	\resizebox{\textwidth}{!}{
		\begin{tabular}{l|ZZZ|ZZZ|ZZZ|ZZZ|ZZZ|ZZZ}
			\toprule
			\multirow{2}{*}{AI Agent} & \multicolumn{3}{c|}{Dim 1 Score} & \multicolumn{3}{c|}{Dim 2 Score} & \multicolumn{3}{c|}{Dim 3 Score} & \multicolumn{3}{c|}{Dim 4 Score} & \multicolumn{3}{c|}{Dim 5 Score} & \multicolumn{3}{c}{Final Score} \\
			\cmidrule(lr){2-4} \cmidrule(lr){5-7} \cmidrule(lr){8-10} \cmidrule(lr){11-13} \cmidrule(lr){14-16} \cmidrule(lr){17-19} & S1 & S2 & S3 & S1 & S2 & S3 & S1 & S2 & S3  & S1 & S2 & S3  & S1 & S2 & S3  & S1 & S2   & S3 \\
			\midrule
			ChatGPT (\texttt{GPT-5.3}) & 75 & 65 & 40 & 80 & 80 & 85 & 80 & 50 & 80  & 85 & 80 & 90  & 80 & 55 & 70 & 80 & 64.5 & 72.8 \\
			ChatGPT Pro (\texttt{GPT-5.4-Pro}) & 85 & 88 & 40 & \underline{90} & 84 & 85 & \textbf{90} & 76 & \underline{95} & 92 & 86 & 90 & \textbf{90} & 78 & 90 & \underline{89.4} & 82 & 80.5 \\
			Gemini (\texttt{Gemini-3.1-Pro}) & 75 & 70 & 40 & 80 & 75 & 80 & 80 & 60 & 80  & 85 & 85 & 90 & 80 & 55 & 90 & 80 & 68.3 & 76 \\
			Claude Pro (\texttt{Claude Opus 4.6}) & \textbf{95} & \underline{90} & 60 & \underline{90} & 82 & \underline{95} & 80 & 28 & \underline{95} & 92 & 78 & \underline{95} & \underline{88} & 42 & \underline{95} & 88.5 & 61.3 & \underline{88} \\
			\texttt{CycleResearcher-ML-12B} & 65 & 75 & 30 & 60 & 45 & 60 & 35 & 15 & 10 & 75 & 55 & 50 & 35 & 15 & 10 & 52.8 & 39.5 & 29.5 \\
			DeepScientist (\texttt{GPT-5.4}) & \textbf{95} & \underline{90} & 60 & \underline{90} & \textbf{90} & \underline{95} & \textbf{90} & \textbf{82} & \underline{95} & \underline{95} & \underline{88} & \underline{95} & \textbf{90} & \underline{80} & \underline{95} & \textbf{92} & \underline{85.6} & \underline{88} \\
			AI Scientist-v2 (\texttt{GPT-5.4}) & 85 & 55 & 50 & 80 & 65 & 40 & 80 & 45 & 30  & 85 & 70 & 70 & 75 & 40 & 20 & 81 & 54 & 41.5 \\
			ARIS (\texttt{GPT-5.4}) & \underline{90} & \textbf{92} & \underline{70} & \textbf{92} & \textbf{90} & 90 & 85 & \underline{78} & 90 & 92 & \textbf{90} & 90 & 85 & \textbf{82} & 90 & 88.5 & \textbf{85.8} & 86 \\
            \midrule
			ReasFlow (\texttt{GPT-5.4}) & \textbf{95} & \underline{90} & \textbf{100} & \textbf{92} & \underline{88} & \textbf{100} & \underline{88} & \textbf{82} & \textbf{100} & \textbf{96} & 84 & \textbf{100} & \textbf{90} & \textbf{82} & \textbf{100} & \textbf{92} & 84.9 & \textbf{100} \\
			\bottomrule
		\end{tabular}
	}
\end{table}

\noindent\textbf{F. References. }For originality and significance, we consider three dimensions including: (i) recency (30 points), (ii) key reference coverage (45 points), and (iii) citation format (25 points). The prompt is as listed in Fig.~\ref{fig:prompt-paper-F}. Corresponding evaluation results are listed in Table~\ref{tab:paper_F_scores_detail}.

\begin{center}
	\begin{LLMBox}[width=\textwidth, fontupper=\small]{System prompt for evaluating \textit{References}.}
		\begin{markdown}
You are the **References Reviewer**.

Dimension rubric
- Dim1 Recency (30): fraction of refs in last 3 years (target $\ge$30\%).
- Dim2 Key Reference Coverage (45): foundational/closest works cited.
- Dim3 Citation Format (25): consistent formatting; DOIs where available.

Evaluation steps
1) Estimate year distribution; compute recent fraction.
2) Check coverage of foundational/closest works mentioned.
3) Inspect formatting consistency and DOIs.

Scoring rules
- Recent fraction $<$15\% → strong deduct Dim1; 15–30\% moderate deduct.
- Missing obvious key works → deduct Dim2; multiple absences may cap total at 40.
- Messy formatting/absent DOIs → deduct Dim3.

Output
- JSON only per schema.
\end{markdown}
	\end{LLMBox}
	\captionof{figure}{System prompt for evaluating references.}
	\label{fig:prompt-paper-F}
\end{center}

\begin{table}[htbp]
	\centering
	\caption{Detailed Scores for \textbf{references}. S1, S2, S3 corresponds to evaluations performed with \texttt{GPT-5.1}, \texttt{GPT-5.4} and \texttt{Gemini-3-Pro} as the reviewer's LLM backend, respectively. Full scores for each dimension are 100. The final score is the weighted average of all dimensions. \textbf{Bold}: best; \underline{underlined}: second best.}
	\label{tab:paper_F_scores_detail}
	\newcolumntype{Z}{>{\centering\arraybackslash}p{1.4em}}
	\begin{tabular}{l|ZZZ|ZZZ|ZZZ|ZZZ}
		\toprule
		\multirow{2}{*}{AI Agent} & \multicolumn{3}{c|}{Dim 1 Score} & \multicolumn{3}{c|}{Dim 2 Score} & \multicolumn{3}{c|}{Dim 3 Score} & \multicolumn{3}{c}{Final Score} \\
		\cmidrule(lr){2-4} \cmidrule(lr){5-7} \cmidrule(lr){8-10} \cmidrule(lr){11-13} & S1 & S2 & S3 & S1 & S2 & S3 & S1 & S2 & S3 & S1 & S2   & S3 \\
		\midrule
		ChatGPT (\texttt{GPT-5.3}) & 20 & 0 & 0 & 25 & 10 & 0 & 20 & 0 & 0 & 22.3 & 4.5 & 0 \\
		ChatGPT Pro (\texttt{GPT-5.4-Pro}) & 85 & \underline{82} & \underline{90} & 80 & 55 & 40 & 80 & \underline{72} & 70 & 81.5 & 67.4 & 62.5 \\
		Gemini (\texttt{Gemini-3.1-Pro}) & \underline{90} & 70 & \textbf{100}  & 40 & 20 & 40 & 80 & 60 & 80 & 65 & 45 & 68 \\
		Claude Pro (\texttt{Claude Opus 4.6}) & \underline{90} & \textbf{84} & \underline{90} & \underline{90} & 72 & \underline{95} & 80 & \underline{72} & \underline{85} & 87.5 & \underline{75.6} & \underline{91} \\
		\texttt{CycleResearcher-ML-12B} & 10 & 0 & 0 & 5 & 0 & 0 & 5 & 0 & 0 & 6.5 & 0 & 0 \\
		DeepScientist (\texttt{GPT-5.4}) & \underline{90} & \underline{82} & \textbf{100} & 85 & 68 & 90 & 85 & \underline{72} & 70 & 86.5 & 73.2 & 88 \\
		AI Scientist-v2 (\texttt{GPT-5.4}) & 85 & 75 & \textbf{100} & 80 & 45 & 80  & 80 & 60 & 50 & 81.5 & 57.8 & 78.5 \\
		ARIS (\texttt{GPT-5.4}) & 85 & 72 & 70 & 88 & \textbf{82} & 90 & \textbf{90} & \textbf{78} & \textbf{90} & \underline{87.6} & \textbf{78} & 84 \\
        \midrule
		ReasFlow (\texttt{GPT-5.4}) & \textbf{95} & 74 & \textbf{100} & \textbf{92} & \underline{76} & \textbf{100} & \underline{88} & 44 & \textbf{90} & \textbf{91.9} & 67.4 & \textbf{97.5} \\
		\bottomrule
	\end{tabular}
\end{table}

\noindent\textbf{G. Internal consistency. }For originality and significance, we consider three dimensions including: (i) numerical consistency (30 points), (ii) symbol consistency (30 points), and (iii) claim-evidence consistency (40 points). The prompt is as listed in Fig.~\ref{fig:prompt-paper-G}. Corresponding evaluation results are listed in Table~\ref{tab:paper_G_scores_detail}.

\begin{center}
	\begin{LLMBox}[width=\textwidth, fontupper=\small]{System prompt for evaluating \textit{Internal consistency}.}
		\begin{markdown}
You are the **Internal Consistency Reviewer**.

Dimension rubric
- Dim1 Numerical Consistency (30): numbers match across abstract, main text, tables.
- Dim2 Symbol Consistency (30): variables use the same symbols throughout.
- Dim3 Claim–Evidence Consistency (40): every claim has theorem/proof/experiment support.

Evaluation steps
1) Cross-check key metrics/figures across sections.
2) Track symbol definitions vs later usage.
3) For each major claim, locate supporting evidence; flag missing links.

Scoring rules
- Conflicting numbers/symbols → heavy deduction Dim1/Dim2.
- Unsupported claims → deduct Dim3; repeated issues may cap total at 40.

Output
- JSON only per schema.
\end{markdown}
	\end{LLMBox}
	\captionof{figure}{System prompt for evaluating internal consistency.}
	\label{fig:prompt-paper-G}
\end{center}

\begin{table}[htbp]
	\centering
	\caption{Detailed Scores for \textbf{internal consistency}. S1, S2, S3 corresponds to evaluations performed with \texttt{GPT-5.1}, \texttt{GPT-5.4} and \texttt{Gemini-3-Pro} as the reviewer's LLM backend, respectively. Full scores for each dimension are 100. The final score is the weighted average of all dimensions. \textbf{Bold}: best; \underline{underlined}: second best.}
	\label{tab:paper_G_scores_detail}
	\newcolumntype{Z}{>{\centering\arraybackslash}p{1.4em}}
	\begin{tabular}{l|ZZZ|ZZZ|ZZZ|ZZZ}
		\toprule
		\multirow{2}{*}{AI Agent}    & \multicolumn{3}{c|}{Dim 1 Score} & \multicolumn{3}{c|}{Dim 2 Score} & \multicolumn{3}{c|}{Dim 3 Score} & \multicolumn{3}{c}{Final Score} \\
		\cmidrule(lr){2-4} \cmidrule(lr){5-7} \cmidrule(lr){8-10} \cmidrule(lr){11-13} & S1 & S2 & S3 & S1 & S2 & S3 & S1 & S2 & S3 & S1 & S2 & S3 \\
		\midrule
		ChatGPT (\texttt{GPT-5.3}) & \textbf{90} & 25 & 80 & 85 & 45 & 90 & 60 & 20 & 20 & 76.5 & 29 & 59 \\
		ChatGPT Pro (\texttt{GPT-5.4-Pro}) & \textbf{90} & 70 & 85 & \textbf{92} & 78 & \underline{95} & \underline{88} & 60 & 70 & \underline{89.8} & 68.4 & 82 \\
		Gemini (\texttt{Gemini-3.1-Pro}) & \underline{85} & 65 & 60 & 85 & 60 & 90 & 55 & 20 & 25 & 73 & 45.5 & 55 \\
		Claude Pro (\texttt{Claude Opus 4.6}) & \textbf{90} & 55  & \underline{95} & 80 & 20 & \underline{95} & 85 & 33 & \underline{95} & 85 & 35.7 & \underline{95} \\
		\texttt{CycleResearcher-ML-12B} & 60 & 25 & 20   & 35 & 10 & 10  & 25 & 5 & 0 & 38.5 & 12.5 & 9    \\
		DeepScientist (\texttt{GPT-5.4}) & \textbf{90} & \textbf{86} & \textbf{100} & \underline{90} & \textbf{82} & \textbf{100} & \underline{88} & 76 & \textbf{100} & 89.2 & \textbf{80.8} & \textbf{100} \\
		AI Scientist-v2 (\texttt{GPT-5.4}) & 75 & 55 & 80 & 85 & 35 & 60 & 60 & 40 & 10 & 72 & 43 & 46   \\
		ARIS (\texttt{GPT-5.4}) & \underline{85} & \underline{78} & \textbf{100} & 80 & \underline{80} & \underline{95} & \underline{88} & \underline{80} & 0 & 84.7 & 79.4 & 58.5 \\
        \midrule
		ReasFlow (\texttt{GPT-5.4}) & \textbf{90} & \underline{78} & \textbf{100} & 88 & \underline{80} & \textbf{100} & \textbf{92} & \textbf{82} & \textbf{100} & \textbf{90.2} & \underline{80.2} & \textbf{100} \\
		\bottomrule
	\end{tabular}
\end{table}

\noindent\textbf{H. Visual quality. }For numerical experiments, we consider five dimensions including: (i) figure quality (25 points), (ii) equation typesetting (20 points), (iii) page layout (20 points), (iv) captions (15 points), and (v) grayscale readability (20 points). The prompt is as listed in Fig.~\ref{fig:prompt-paper-H}. Corresponding evaluation results are listed in Table~\ref{tab:paper_H_scores_detail}.

\begin{center}
	\begin{LLMBox}[width=\textwidth, fontupper=\small]{System prompt for evaluating \textit{Visual quality}.}
		\begin{markdown}
You are the **Visual Quality Reviewer**.

Dimension rubric
- Dim1 Figure Quality (25): resolution, readable fonts, labeled axes with units.
- Dim2 Equation Typesetting (20): correct rendering, no overflow/broken symbols.
- Dim3 Page Layout (20): consistent margins/spacing/fonts per template.
- Dim4 Captions (15): numbered captions for all figures/tables.
- Dim5 Grayscale Readability (20): distinguishable when printed in B/W.

Evaluation steps
1) Scan pages for figures/tables; check captions and readability.
2) Inspect equations for rendering issues.
3) Note layout consistency and grayscale legibility.

Scoring rules
- Blurry/uncaptioned figures → deduct Dim1/Dim4.
- Broken equations → deduct Dim2.
- Layout issues → deduct Dim3; poor grayscale → deduct Dim5.

Output
- JSON only per schema.
\end{markdown}
	\end{LLMBox}
	\captionof{figure}{System prompt for evaluating visual quality.}
	\label{fig:prompt-paper-H}
\end{center}

\begin{table}[htbp]
	\centering
	\caption{Detailed Scores for \textbf{visual quality}. S1, S2, S3 corresponds to evaluations performed with \texttt{GPT-5.1}, \texttt{GPT-5.4} and \texttt{Gemini-3-Pro} as the reviewer's LLM backend, respectively. Full scores for each dimension are 100. The final score is the weighted average of all dimensions. \textbf{Bold}: best; \underline{underlined}: second best.}
	\label{tab:paper_H_scores_detail}
	\newcolumntype{Z}{>{\centering\arraybackslash}p{1.4em}}
	\resizebox{\textwidth}{!}{
		\begin{tabular}{l|ZZZ|ZZZ|ZZZ|ZZZ|ZZZ|ZZZ}
			\toprule
			\multirow{2}{*}{AI Agent}    & \multicolumn{3}{c|}{Dim 1 Score} & \multicolumn{3}{c|}{Dim 2 Score} & \multicolumn{3}{c|}{Dim 3 Score} & \multicolumn{3}{c|}{Dim 4 Score} & \multicolumn{3}{c|}{Dim 5 Score} & \multicolumn{3}{c}{Final Score} \\
			\cmidrule(lr){2-4} \cmidrule(lr){5-7} \cmidrule(lr){8-10} \cmidrule(lr){11-13} \cmidrule(lr){14-16} \cmidrule(lr){17-19} & S1 & S2 & S3 & S1 & S2 & S3 & S1 & S2 & S3  & S1 & S2 & S3 & S1 & S2 & S3 & S1 & S2 & S3 \\
			\midrule
			ChatGPT (\texttt{GPT-5.3}) & \underline{80} & 35 & 0 & 90 & 75 & 90 & 85 & 80 & 85 & \underline{85} & 60 & 80  & \underline{90} & \textbf{85} & \textbf{100} & 85.8 & 65.8 & 67 \\
			ChatGPT Pro (\texttt{GPT-5.4-Pro}) & \underline{80} & 60  & 0 & \underline{92} & \textbf{85} & \underline{95} & \underline{90} & \underline{82} & 90 & \underline{85} & 75 & 80 & \textbf{95} & 70 & \textbf{100} & \underline{88.2} & \underline{73.7} & 69 \\
			Gemini (\texttt{Gemini-3.1-Pro}) & 60 & 20 & 0 & 85  & 75 & 90 & 85 & 70 & 90 & 60 & 10 & 0 & \underline{90} & 50 & \textbf{100} & 76 & 45.5 & 56 \\
			Claude Pro (\texttt{Claude Opus 4.6}) & 75 & 55 & 0 & 90 & 80 & \underline{95} & 85 & \underline{82} & 90  & \underline{85} & 68 & 80 & 80 & \textbf{85} & \textbf{100} & 82.5 & 73.4 & 69   \\
			\texttt{CycleResearcher-ML-12B} & 50 & 25 & 0 & 70 & 40 & 20 & 75 & 65 & 50 & 70 & 50 & 10 & 80 & 50 & 80 & 68 & 44.8 & 31.5 \\
			DeepScientist (\texttt{GPT-5.4}) & \underline{80} & 50 & 0 & \textbf{95} & 82 & \underline{95} & \underline{90} & \textbf{84} & \underline{95} & \underline{85} & 78 & 40 & 85 & \underline{80} & \textbf{100} & 86.8 & 73.4 & 64 \\
			AI Scientist-v2 (\texttt{GPT-5.4}) & \underline{80} & \underline{65} & \textbf{90} & 90 & 55 & 90 & 85 & 75 & 90 & \textbf{90} & \underline{80} & \underline{90} & 70 & 40 & 70 & 82.5 & 62.3 & \underline{86} \\
			ARIS (\texttt{GPT-5.4}) & 70 & 35 & 0 & 90 & \underline{84} & 90 & 85 & \underline{82} & 90 & 80 & 72 & 0 & 80 & \underline{80} & \textbf{100} & 80.5 & 68.8 & 56 \\
            \midrule
			ReasFlow (\texttt{GPT-5.4}) & \textbf{90} & \textbf{76} & \underline{85} & \textbf{95} & 82 & \textbf{100} & \textbf{92} & 80 & \textbf{100} & \textbf{90} & \textbf{84} & \textbf{100} & 85 & 52 & \underline{85} & \textbf{90.4} & \textbf{74.4} & \textbf{93.3} \\
			\bottomrule
		\end{tabular}
	}
\end{table}

\noindent\textbf{I. AI and ethics compliance. }For originality and significance, we consider three dimensions including: (i) GenAI disclosure (30 points), (ii) data authenticity (45 points), and (iii) originality declaration (25 points). The prompt is as listed in Fig.~\ref{fig:prompt-paper-I}. Corresponding evaluation results are listed in Table~\ref{tab:paper_I_scores_detail}.

\begin{center}
	\begin{LLMBox}[width=\textwidth, fontupper=\small]{System prompt for evaluating \textit{AI and ethics compliance}.}
		\begin{markdown}
You are the **AI \\& Ethics Compliance Reviewer**.

Dimension rubric
- Dim1 GenAI Disclosure (30): if AI tools used, disclosure present; if no evidence of use, disclosure not required.
- Dim2 Data Authenticity (45): signs of fabricated/manipulated data (too-perfect results, repeated patterns).
- Dim3 Originality Declaration (25): statements about no simultaneous submission / no plagiarism.

Evaluation steps
1) Search for AI/tool acknowledgments; if AI evident but undisclosed → penalize.
2) Inspect experimental results for suspicious perfection; note any red flags.
3) Look for originality/ethics statements; mark absence.

Scoring rules
- No AI use evidence \\& no disclosure needed → around 80 for Dim1; AI evident but undisclosed → severe deduction ($\sim$20).
- Suspicious data → deduct Dim2; genuine-looking data → around 85.
- Missing originality declaration → deduct Dim3 (common omission $\sim$50).

Output
- JSON only per schema.
\end{markdown}
	\end{LLMBox}
	\captionof{figure}{System prompt for evaluating AI and ethics compliance.}
	\label{fig:prompt-paper-I}
\end{center}

\begin{table}[htbp]
	\centering
	\caption{Detailed Scores for \textbf{AI and ethics compliance}. S1, S2, S3 corresponds to evaluations performed with \texttt{GPT-5.1}, \texttt{GPT-5.4} and \texttt{Gemini-3-Pro} as the reviewer's LLM backend, respectively. Full scores for each dimension are 100. The final score is the weighted average of all dimensions. \textbf{Bold}: best; \underline{underlined}: second best.}
	\label{tab:paper_I_scores_detail}
	\newcolumntype{Z}{>{\centering\arraybackslash}p{1.4em}}
	\begin{tabular}{l|ZZZ|ZZZ|ZZZ|ZZZ}
		\toprule
		\multirow{2}{*}{AI Agent}    & \multicolumn{3}{c|}{Dim 1 Score} & \multicolumn{3}{c|}{Dim 2 Score} & \multicolumn{3}{c|}{Dim 3 Score} & \multicolumn{3}{c}{Final Score} \\
		\cmidrule(lr){2-4} \cmidrule(lr){5-7} \cmidrule(lr){8-10} \cmidrule(lr){11-13} & S1 & S2 & S3 & S1 & S2 & S3 & S1 & S2 & S3 & S1 & S2   & S3 \\
		\midrule
		ChatGPT (\texttt{GPT-5.3}) & \textbf{80} & \textbf{80} & \textbf{80} & \underline{85} & 60 & \underline{85} & \underline{50} & 40 & \underline{50} & 74.8 & 61 & 74.8 \\
		ChatGPT Pro (\texttt{GPT-5.4-Pro}) & \textbf{80} & \textbf{80} & \textbf{80} & \textbf{90} & \underline{78} & \underline{85} & \underline{50} & \underline{45} & \underline{50} & \underline{77} & \underline{70.4} & 74.8 \\
		Gemini (\texttt{Gemini-3.1-Pro}) & \textbf{80} & \textbf{80} & \textbf{80} & \underline{85} & 65 & 80 & \underline{50} & \underline{45} & \textbf{60} & 74.8 & 64.5 & \underline{75}   \\
		Claude Pro (\texttt{Claude Opus 4.6}) & \textbf{80} & \textbf{80} & \textbf{80} & \textbf{90} & 65 & \underline{85} & \underline{50} & 41 & \underline{50} & \underline{77} & 63.5 & 74.8 \\
		\texttt{CycleResearcher-ML-12B} & \textbf{80} & \textbf{80} & 0 & 75 & 45 & 0 & \underline{50} & 40 & 0 & 70.3 & 54.3 & 0 \\
		DeepScientist (\texttt{GPT-5.4}) & \textbf{80} & \textbf{80} & \textbf{80} & \textbf{90} & \underline{78} & \underline{85} & \underline{50} & \underline{45} & \underline{50} & \underline{77} & \underline{70.4} & 74.8 \\
		AI Scientist-v2 (\texttt{GPT-5.4}) & \textbf{80} & \textbf{80} & \underline{20} & \underline{85} & 55 & 20 & \underline{50} & 35 & \underline{50} & 74.8 & 57.5 & 27.5 \\
		ARIS (\texttt{GPT-5.4}) & \textbf{80} & \textbf{80} & \textbf{80} & \textbf{90} & 72 & \underline{85} & \underline{50} & \textbf{50} & \underline{50} & \underline{77} & 68.9 & 74.8 \\
        \midrule
		ReasFlow (\texttt{GPT-5.4}) & \textbf{80} & \textbf{80} & \textbf{80} & \textbf{90} & \textbf{84} & \textbf{100} & \textbf{60} & 40 & \underline{50} & \textbf{79.5} & \textbf{71.8} & \textbf{81.5} \\
		\bottomrule
	\end{tabular}
\end{table}

\noindent\textbf{Overall score. }The overall score of the evaluated paper is computed via the following weighted average formula:
\begin{align*}
	S_{\text{full}} = 0.25\cdot S_A + 0.15\cdot S_B+0.2\cdot S_C+0.1\cdot S_D+0.1\cdot S_E + 0.05\cdot S_F+0.05\cdot S_G+0.05\cdot S_H+0.05\cdot S_I,
\end{align*}
where $S_A,S_B,\cdots,S_I$ denote the scores of the nine scholarly aspects, respectively.

\subsection{Experimental Details for Sec.~\ref{subsubsec:exp-survey}}\label{app:exp-survey}

To evaluate the \textit{Related Work} section produced by the \texttt{SurveyAgent}, we compare ReasFlow's agent against a base LLM baseline across three survey tasks (\paperone, \papertwo, and \paperthree). Each task is generated with four different LLM backbones (\texttt{DeepSeek-v3.2}, \texttt{GPT-5.1}, \texttt{GPT-5.4}, and \texttt{Claude Sonnet~4.6}), and independently scored by three evaluation models for robustness. The evaluation employs two dimensions, each scored on a 1-10 scale:

\begin{itemize}
	\item \textbf{Content Accuracy (Dim1)}: assesses the factual correctness of cited paper descriptions. For each article, 15 citations are sampled and verified against a 339-paper FAISS vector database. Papers with valid arXiv IDs are verified via full-text extraction; others are matched through semantic search. Each citation receives an accuracy level from 0 (fabricated) to 3 (essentially correct).
	\item \textbf{Citation Relevance (Dim2)}: evaluates whether citations align with the survey plan and are topically appropriate. The score combines key reference coverage, per-citation relevance ratings (0--2), citation count bonuses, and off-topic penalties.
\end{itemize}

\subsubsection{Generation Details}

Each of the three tasks is associated with a \texttt{survey\_plan.md} file that specifies the research topic, required structure, and a list of key references. This plan is provided verbatim to both ReasFlow and the baseline as the sole input.

\noindent\textbf{ReasFlow (\texttt{SurveyAgent}).} The \texttt{SurveyAgent} is invoked through its CLI in fully autonomous mode. The agent reads the survey plan and executes a structured three-step workflow: (i) \texttt{SurveyWriteOutline} generates a section outline; (ii) \texttt{SurveyWriteSurvey} compiles a comprehensive survey report by querying both the built-in FAISS paper database (339 knowledge cards) and external sources via the Semantic Scholar API; and (iii) \texttt{SurveyWriteRelatedWork} produces the final \texttt{related\_works.tex} and \texttt{references.bib}. The agent has access to the full tool suite described in Appendix~\ref{app:agent_specs}, including semantic paper search, citation network traversal, and an automated review tool that evaluates the draft across four quality dimensions and triggers iterative revision.

\noindent\textbf{Baseline (Base LLM).} The baseline uses an identical LLM backbone but operates through a minimal agent that lacks the structured survey tools. It receives the same survey plan and is instructed to produce the same output files using only basic literature search tools (\texttt{LiteratureSearch}, \texttt{LiteratureGetReferences}, \texttt{LiteratureGetCitations}) and file-writing capabilities. No structured outline, multi-step survey compilation, or automated review is available; the baseline generates the final output in a single agent pass.

\noindent\textbf{Variant matrix and orchestration.} Each of the two agents is paired with four LLM backbones, \texttt{DeepSeek-v3.2}, \texttt{GPT-5.1}, \texttt{GPT-5.4}, and \texttt{Claude Sonnet~4.6}, yielding $2 \times 4 = 8$ generation variants per task and $8 \times 3 = 24$ total generation runs. Each (task, variant) pair executes in an isolated workspace directory to prevent cross-contamination. The entire pipeline is driven by a batch script that reads a YAML configuration specifying the task list, variant definitions (agent type, LLM backbone, and mode), and evaluation model list. For each (task, variant), the script (i) creates an isolated workspace, (ii) writes the survey plan into the workspace, (iii) updates the agent's model configuration to the target backbone, (iv) invokes the corresponding agent CLI, and (v) copies the resulting \texttt{related\_works.tex} and \texttt{references.bib} to the shared evaluation directory. Generation runs that do not produce valid output files are logged as failures and excluded from evaluation.

\subsubsection{Evaluation Details}

The evaluation pipeline proceeds in two phases. First, for \textbf{Content Accuracy}, each of the 15 sampled citations undergoes a multi-step verification: (i)~if the citation has a valid arXiv ID, the full PDF is downloaded and its first 3\,000 characters are extracted as evidence; (ii)~otherwise, we query Semantic Scholar with the title and authors, and an LLM disambiguates the best match from the candidate list; (iii)~the evaluator LLM receives the paper's actual content alongside the description in the \textit{Related Work} and assigns an accuracy level on a 0--3 scale. The accuracy levels are then mapped to a normalized score: level 3 $\to$ 1.0, level 2 $\to$ 0.65, level 1 $\to$ 0.15, level 0 $\to$ 0.0, and rescaled to $[1, 10]$.

Second, for \textbf{Citation Relevance}, the pipeline (i)~checks all key references specified in the survey plan for coverage; (ii)~scores each sampled citation's relevance (0--2 scale: strongly relevant, weakly relevant, or irrelevant) against the plan; and (iii)~computes the final score via:
\begin{equation*}
	\text{Dim2} = \max\Bigl(1,\; \min\bigl(10,\; \underbrace{r_\text{cov} \times 7}_{\text{key coverage}} + \underbrace{r_\text{rel} \times 2}_{\text{quality bonus}} + \underbrace{\min(3, n_\text{cite}/40 \times 3)}_{\text{richness bonus}} - \underbrace{\max(0, (r_\text{irr} - 0.15)/0.85 \times 2.5)}_{\text{off-topic penalty}}\bigr)\Bigr),
\end{equation*}
where $r_\text{cov}$ is the key reference coverage rate, $r_\text{rel}$ is the fraction of strongly relevant citations, $n_\text{cite}$ is the total citation count, and $r_\text{irr}$ is the fraction of irrelevant citations. Fabricated papers detected in Dim1 are automatically counted as irrelevant.

The complete prompts for the two evaluation dimensions are presented in Fig.~\ref{fig:prompt-survey-dim1} and Fig.~\ref{fig:prompt-survey-dim2}.

\begin{center}
	\begin{LLMBox}[width=\textwidth, fontupper=\small]{Prompt for content accuracy verification (Dim1)}
		\begin{markdown}
You are an academic integrity reviewer. Judge whether the description in the Related Works accurately represents the cited paper.

**Important Note: Multi-Paper Joint Citations**

If the citation appears together with other papers (e.g., cite a, b, c), the description may be a joint summary of the entire group. In this case, only judge whether **this paper supports some reasonable part** of the description. If another paper in the group covers a portion the current paper does not address, do not penalize the current paper for that.

**Source Paper Information**
- Title: [paperTitle]
- Authors: [paperAuthors]
- Year: [paperYear]
- Paper excerpt (first 3000 chars): [paperContent]

**Description in Related Works**
	[description]

**Task**
Check whether the description matches the paper's actual content:
1. Is the paper's main method/contribution accurately described?
2. Are there exaggerations, distortions, or clearly incorrect statements?
3. Are key data/conclusions wrong?
4. For multi-paper citations, only judge if this paper supports some reasonable part.

Output JSON: accuracyLevel (0/1/2/3), hallucinationType (null / "misrepresentation" / "exaggeration" / "wrongMetadata"), details ("explanation or null if accurate").

accuracyLevel rubric:
- 3 = Essentially correct: consistent with paper content, no factual errors
- 2 = Minor deviation: does not affect reader understanding
- 1 = Substantial error in method/contribution: would mislead readers
- 0 = Completely fabricated or entirely irrelevant to this paper
\end{markdown}
	\end{LLMBox}
	\captionof{figure}{Prompt for verifying content accuracy of individual citations in the Related Works (Dim1).}
	\label{fig:prompt-survey-dim1}
\end{center}

\begin{center}
	\begin{LLMBox}[width=\textwidth, fontupper=\small]{Prompt for citation relevance scoring (Dim2)}
		\begin{markdown}
You are an academic paper reviewer evaluating the relevance of a citation in a Related Works section.

**Survey Plan**
	[plan]

**Citation Under Evaluation**
- Title: [title]
- Authors: [authors]
- Year: [year]
- Venue: [venue]

**Scoring Rubric (0--2)**

Judge relevance strictly against the Survey Plan:

**2 = Strongly relevant**: satisfies any of the following:
- Explicitly listed in the Plan's "Key References"
- A directly representative work of a core research direction described in the Plan
- A background work that directly motivates this research, clearly mentioned or highly aligned with the Plan

**1 = Weakly relevant**: satisfies any of the following:
- An adjacent technique to the Plan's core direction, with some overlap but not central
- Broader context for the research background; reasonable to mention but not emphasized by the Plan

**0 = Irrelevant**:
- No apparent connection to any core direction in the Plan
- Clearly belongs to a different research area

Output JSON: relevance (0--2), reason ("brief explanation").
\end{markdown}
	\end{LLMBox}
	\captionof{figure}{Prompt for scoring per-citation relevance against the survey plan (Dim2).}
	\label{fig:prompt-survey-dim2}
\end{center}

\noindent The total score is $\text{Dim1} + \text{Dim2}$ (out of 20). Results are summarized in Table~\ref{tab:survey_scores_detail}. To assess evaluation robustness, we employ three independent evaluation models (\texttt{GPT-5.1}, \texttt{Sonnet~4.6}, and \texttt{DeepSeek-v3.2}) and report per-judger total scores for each generation backbone. The \texttt{SurveyAgent} achieves an overall advantage of \textbf{+3.32 points} over the base model. The improvement is predominantly driven by Content Accuracy (Dim1: +2.97), as the agent's citations originate from the verified paper library and thus achieve high retrieval hit rates. Across all 36 paired comparisons (4 backbones $\times$ 3 tasks $\times$ 3 evaluation models), the \texttt{SurveyAgent} wins in \textbf{29 cases (80.6\% win rate)}.

\begin{table}[htbp]
	\centering
	\caption{Per-task evaluation of the \texttt{SurveyAgent} vs.\ the base LLM across four generation backbones and three evaluation models. Each cell reports the total score (Content Accuracy $+$ Citation Relevance, out of 20). For each evaluation model, the best ReasFlow score and the best advantage are highlighted in \textbf{bold}.}
	\label{tab:survey_scores_detail}
	\resizebox{\textwidth}{!}{
		\begin{tabular}{ll|ccc|ccc|ccc|ccc}
			\toprule
			\multirow{2}{*}{Task}                & \multirow{2}{*}{Backbone} & \multicolumn{3}{c|}{Judger: \texttt{GPT-5.1}} & \multicolumn{3}{c|}{Judger: \texttt{Sonnet 4.6}} & \multicolumn{3}{c|}{Judger: \texttt{DeepSeek-v3.2}} & \multicolumn{3}{c}{Average}                                                                                                                         \\
			\cmidrule(lr){3-5} \cmidrule(lr){6-8} \cmidrule(lr){9-11} \cmidrule(lr){12-14}
			                                     &                           & Base                                 & RF                                      & Adv.                                  & Base                        & RF             & Adv.            & Base  & RF             & Adv.           & Base  & RF             & Adv.            \\
			\midrule
			\multirow{5}{*}{\shortstack[l]{\paperone}}			                                     & \texttt{DeepSeek-v3.2}                  & 11.70                                & 18.34                                   & +6.64                                 & 14.39                       & \textbf{18.25} & +3.86           & 11.87 & 17.19          & +5.32          & 12.65 & 17.93          & +5.27           \\
			                                     & \texttt{GPT-5.1}                   & 19.27                                & \textbf{18.83}                          & $-0.44$                               & 19.32                       & 18.08          & $-1.24$         & 19.00 & 16.44          & $-2.56$        & 19.20 & 17.78          & $-1.41$         \\
			                                     & \texttt{GPT-5.4}                  & 8.99                                 & 17.09                                   & +8.10                                 & 9.26                        & 17.01          & +7.75           & 8.99  & 16.15          & +7.16          & 9.08  & 16.75          & +7.67           \\
			                                     & \texttt{Sonnet 4.6}                & 10.86                                & 17.77                                   & +6.91                                 & 11.31                       & 17.87          & +6.56           & 11.04 & 16.07          & +5.03          & 11.07 & 17.24          & +6.17           \\
			\cmidrule(lr){2-14}
			                                     & Average                   & 12.71                                & 18.01                                   & +5.30                                 & 13.57                       & 17.80          & +4.23           & 12.72 & 16.46          & +3.74          & 13.00 & 17.42          & +4.42           \\
			\midrule
			\multirow{5}{*}{\shortstack[l]{\papertwo}}			                                     & \texttt{DeepSeek-v3.2}                  & 13.63                                & 15.23                                   & +1.60                                 & 13.04                       & 15.42          & +2.38           & 13.18 & 13.79          & +0.61          & 13.28 & 14.81          & +1.53           \\
			                                     & \texttt{GPT-5.1}                   & 14.24                                & 16.27                                   & +2.03                                 & 15.76                       & 16.90          & +1.14           & 13.55 & 16.71          & +3.16          & 14.52 & 16.63          & +2.11           \\
			                                     & \texttt{GPT-5.4}                   & 13.68                                & 15.51                                   & +1.83                                 & 13.81                       & 17.15          & +3.34           & 12.21 & 15.80          & +3.59          & 13.23 & 16.15          & +2.92           \\
			                                     & \texttt{Sonnet 4.6}                & 14.37                                & 15.50                                   & +1.13                                 & 13.43                       & 15.42          & +1.99           & 14.28 & 13.79          & $-0.49$        & 14.03 & 14.90          & +0.88           \\
			\cmidrule(lr){2-14}
			                                     & Average                   & 13.98                                & 15.63                                   & +1.65                                 & 14.01                       & 16.22          & +2.21           & 13.30 & 15.02          & +1.72          & 13.77 & 15.62          & +1.86           \\
			\midrule
			\multirow{5}{*}{\shortstack[l]{\paperthree}}			                                     & \texttt{DeepSeek-v3.2}                  & 6.52                                 & 18.25                                   & \textbf{+11.73}                       & 7.56                        & \textbf{18.25} & \textbf{+10.69} & 10.33 & \textbf{18.59} & \textbf{+8.26} & 8.14  & \textbf{18.36} & \textbf{+10.23} \\
			                                     & \texttt{GPT-5.1}                   & 6.00                                 & 16.70                                   & +10.70                                & 6.00                        & 15.86          & +9.86           & 9.00  & 15.54          & +6.54          & 7.00  & 16.03          & +9.03           \\
			                                     & \texttt{GPT-5.4}                   & 13.19                                & 14.59                                   & +1.40                                 & 13.86                       & 14.72          & +0.86           & 12.86 & 14.21          & +1.35          & 13.30 & 14.51          & +1.20           \\
			                                     & \texttt{Sonnet 4.6}                & 17.47                                & 11.38                                   & $-6.09$                               & 17.87                       & 13.29          & $-4.58$         & 16.62 & 10.18          & $-6.44$        & 17.32 & 11.62          & $-5.70$         \\
			\cmidrule(lr){2-14}
			                                     & Average                   & 10.79                                & 15.23                                   & +4.44                                 & 11.32                       & 15.53          & +4.21           & 12.20 & 14.63          & +2.43          & 11.44 & 15.13          & +3.69           \\
			\midrule
			\multicolumn{2}{l|}{Overall Average} & 12.49                     & 16.29                                & +3.80                                   & 12.97                                 & 16.52                       & +3.55          & 12.74           & 15.37 & +2.63          & 12.73          & 16.06 & +3.32                            \\
			\bottomrule
		\end{tabular}
	}
\end{table}

{\color{black}
\subsection{Experimental Details for Sec.~\ref{subsubsec:exp-alg}}\label{app:exp-alg}

\subsubsection{Generation Details}

Table~\ref{tab:task-list} lists the 35 evaluation tasks, 20 of which are drawn from MLR‑Bench. The generation prompts are presented in Fig.~\ref{fig:prompt-alg-chain1-r} (for ReasFlow) and Fig.~\ref{fig:prompt-alg-chain1} (for the baselines). We provide the baselines with more detailed requirements to ensure they can produce comparable results.

\begin{center}
	\begin{LLMBox}[width=\textwidth, fontupper=\small]{Prompt for algorithmic tests (ReasFlow).}
		\begin{markdown}
Please design an algorithm based on the following.

Task: \\{task\\}

Research Idea: \\{idea\\}

Please work in AUTO MODE and complete the full algorithm design
without waiting for user confirmation.
\end{markdown}
	\end{LLMBox}
	\captionof{figure}{ReasFlow prompt for algorithmic tests.}
	\label{fig:prompt-alg-chain1-r}
\end{center}

\begin{center}
	\begin{LLMBox}[width=\textwidth, fontupper=\small]{Prompt for algorithmic tests (Baseline).}
		\begin{markdown}
You are an expert algorithm designer. Given a research task and a research idea,
produce TWO outputs in the current working directory:

1. A complete Python implementation saved as `code/algorithm.py`
2. A detailed algorithm design report saved as `document/algorithm_design.md`

CRITICAL INSTRUCTIONS FOR FILE CREATION:
- Create each file with a SEPARATE `apply_patch` command (one patch per file)
- Create `code/algorithm.py` FIRST, then `document/algorithm_design.md` SECOND
- Keep each patch simple — avoid backticks, special characters, or long heredocs inside the content

The Python implementation should:
- Define a class with "Algorithm" or "Optimizer" in its name
- Implement an `optimize()` method (or equivalent main entry point)
- Implement a `get_convergence_history()` method that returns optimization history
- Be self-contained and runnable with standard ML libraries (numpy, scipy, torch, etc.)
- Include a `if __name__ == "__main__":` block with a toy example demonstrating usage

The algorithm design report MUST include all of the following sections:
1. **Algorithm Overview** — what problem it solves and the core intuition
2. **Key Ideas and Innovations** — what is novel about this approach
3. **Algorithm Steps** — detailed procedural description (numbered steps)
4. **Pseudocode** — in algorithmic/LaTeX-style notation
5. **Complexity Analysis** — time and space complexity with justification
6. **Implementation Notes** — key implementation considerations, dependencies, frameworks
7. **Hyperparameters** — table of hyperparameters with defaults and ranges
8. **Expected Behavior and Limitations** — what the algorithm should achieve and its known limitations

IMPORTANT: Do NOT include experimental design, dataset descriptions, or evaluation metrics.
Focus ONLY on the algorithm itself.

The `document/` and `code/` directories already exist in your working directory.

---

Task: \\{task\\}

Research Idea: \\{idea\\}

---

Please create both files now.
\end{markdown}
	\end{LLMBox}
	\captionof{figure}{Baseline prompt for algorithmic tests.}
	\label{fig:prompt-alg-chain1}
\end{center}

\begin{table}[h]
\centering
\caption{Task list for evaluation. Each evaluation task comprises a task description and
  a research-idea input.}
\label{tab:task-list}
\resizebox{\linewidth}{!}{%
\begin{tabular}{llp{8cm}}
\toprule
Source & Task ID & Description \\
\midrule
\multirow{10}{*}{ICLR 2025 workshops (MLR-Bench)}
 & \texttt{iclr2025\_mldpr}      & Machine Learning for Drug Pharmacology Research \\
 & \texttt{iclr2025\_scsl}       & Self-Supervised and Contrastive Learning \\
 & \texttt{iclr2025\_verifai}    & Verification and Formal Methods for AI \\
 & \texttt{iclr2025\_bi\_align}  & Bidirectional Alignment in LLMs \\
 & \texttt{iclr2025\_buildingtrust} & Building Trust in AI Systems \\
 & \texttt{iclr2025\_data\_problems} & Data-centric ML Problems \\
 & \texttt{iclr2025\_dl4c}       & Deep Learning for Code \\
 & \texttt{iclr2025\_question}   & Question Answering and Reasoning \\
 & \texttt{iclr2025\_scope}      & Scope and Generalisation in NLP \\
 & \texttt{iclr2025\_wsl}        & Weakly Supervised Learning \\
\midrule
\multirow{10}{*}{Optimization workshops (MLR-Bench)}
 & \texttt{neurips2023\_opt}              & NeurIPS 2023 optimization workshop \\
 & \texttt{neurips2024\_opt}              & NeurIPS 2024 optimization workshop \\
 & \texttt{neurips2023\_otml}             & optimization and ML (NeurIPS 2023) \\
 & \texttt{neurips2023\_heavytails}       & Heavy-tail phenomena in optimization \\
 & \texttt{neurips2023\_federated\_learning} & Federated optimization \\
 & \texttt{neurips2024\_fitml}            & Fitting ML models at scale \\
 & \texttt{icml2023\_sods}               & Sparse and over-parameterised deep systems \\
 & \texttt{icml2023\_differentiable}     & Differentiable programming \\
 & \texttt{icml2023\_fl}                 & Federated learning (ICML 2023) \\
 & \texttt{icml2024\_hdlearning}         & High-dimensional learning \\
\midrule
\multirow{15}{*}{Fine-grained optimization (OR-Bench)}
 & \texttt{opt\_adam\_variants}       & Adaptive optimisers (Adam variants) \\
 & \texttt{opt\_lora\_lowrank}        & Low-rank adaptation / LoRA \\
 & \texttt{opt\_quantization\_training} & Quantisation-aware training \\
 & \texttt{opt\_sparsification}       & Gradient and weight sparsification \\
 & \texttt{opt\_curriculum\_data}     & Curriculum and data ordering \\
 & \texttt{opt\_multiobjective}       & Multi-objective optimization \\
 & \texttt{opt\_zeroth\_order}        & Zeroth-order and black-box optimization \\
 & \texttt{opt\_memory\_efficient}    & Memory-efficient training \\
 & \texttt{opt\_second\_order}        & Second-order methods \\
 & \texttt{opt\_bilevel}              & Bilevel optimization \\
 & \texttt{opt\_online\_learning}     & Online learning and regret minimisation \\
 & \texttt{opt\_diffusion\_training}  & Diffusion model training optimization \\
 & \texttt{opt\_distributed}          & Distributed and asynchronous optimization \\
 & \texttt{opt\_hpo}                  & Hyperparameter optimization \\
 & \texttt{opt\_mixed\_precision}     & Mixed-precision training \\
\bottomrule
\end{tabular}%
}
\end{table}

\subsubsection{Evaluation Details}

We evaluate whether the generated Python code for the algorithm is directly runnable. The S.T. column in Table~\ref{tab:runnability} reports the number of script-test task records evaluated for each system. A first‑run success is defined as the script executing without runtime errors on its initial attempt. If it fails, the agent is instructed to retry the task until it either succeeds or the total time limit is reached. The reported S.R. counts only the initial validation attempt; later retries are not counted in the first-run success rate.
}

\subsection{Experimental Details for Sec.~\ref{subsubsec:exp-prover}}
\subsubsection{Generation Details}
In this subsection, we clarify the prompt used to generate the proof reports for further evaluation and comparison, which is shown in Fig.~\ref{fig:prompt-proof-paper2}.

\begin{figure}[htbp]
\centering
	\begin{NBLLMBox}[width=\textwidth, fontupper=\small]{Prompt for Proof Report Generation}
		\vspace{-2em}
		\begin{algorithm}[H]
			\caption{Subspace SCAFFOLD}
			\label{Alg:Subspace-SCALLION}
			\begin{algorithmic}
				\REQUIRE Initial model $x^{0}$ and control variables $\{c_{i}^{0}\}_{i=1}^{N}$, $c^{0}$; local learning rate $\eta_{l}$; global learning rate $\eta_{g}$; local steps $K$; number of sampled clients $S$; subspace dimension $r$; outer cycles $T$
				\STATE \textbf{Initialize subspace projector}: obtain $\mathbf{P}_0\in O(\mathbb{R}^{r\times d})$ by common random seed

				\FOR{$t=0,\cdots,T-1$}
				\STATE \textbf{Decompose model and control}:
				$x^t_{proj} = P_t x^t$, $x^t_{res} = x^t - P_t^\top x^t_{proj}$,
				$c^t_{proj} = P_t c^t$, $c^t_{res} = c^t - P_t^\top c^t_{proj}$

				\FOR{ all client $i$ in parallel}
				\STATE \textbf{Client control split}:
				$c_{i,proj}^t = P_t c_i^t$, $c_{i,res}^t = c_i^t - P_t^\top c_{i,proj}^t$

				\STATE Initialize $y_{i,proj}^{t,0} = x^t_{proj}$
				\FOR{$k=0,\ldots,K-1$}
				\STATE Reconstruct full model: $y_i^{t,k} = P_t^\top y_{i,proj}^{t,k} + x^t_{res}$
				\STATE Compute mini-batch gradient: $g_i^{t,k} = \nabla F(y_i^{t,k};\xi_i^{t,k})$

				\STATE Update projected model:
				$y_{i,proj}^{t,k+1} = y_{i,proj}^{t,k} - \eta_l(P_t g_i^{t,k} - c_{i,proj}^t + c^t_{proj})$
				\ENDFOR

				\STATE Update client control:
				$c_{i}^{t+1} = (I-P_t^\top P_t)c_{i}^{t}+\frac{1}{K}P_t^\top P_t\sum_{k=0}^{K-1} g_i^{t,k}$

				\ENDFOR

				\STATE Update projected model: $x_{proj}^{t+1} = x_{proj}^t - \eta_g \frac{\eta_l K}{N}\sum_{i} \frac{x^t_{proj} - y_{i,proj}^{t,K}}{\eta_l K}$

				\STATE \textbf{Backfill to full space}:
				$x^{t+1} = P_t^\top x_{proj}^{t+1} + x^t_{res}$,
				$c^{t+1} =  \frac{1}{NK}\sum_{i=0}^{N-1}\sum_{k=0}^{K-1}P_t^\top P_t g_i^{t,k} + c^t_{res}$

				\STATE Update projector: generate new $P_{t+1}$ for next round
				\ENDFOR
			\end{algorithmic}
		\end{algorithm}
		\vspace{-1em}
		Prove the convergence result of this algorithm under stochastic non-convex scenario, with the following assumptions:

		\begin{assumption}[$L$-smoothness]\label{ass:A1}
			Each client loss $F_i:\mathbb{R}^d\!\to\!\mathbb{R}$ satisfies
			$\|\nabla F_i(x)-\nabla F_i(y)\|\le L\|x-y\|$ for all $x,y$.
		\end{assumption}
		\begin{assumption}[Bounded variance and independence across samples]\label{ass:A2}
			For stochastic gradients $g_i(x;\xi)$,
			$\mathbb{E}[g_i(x;\xi)\mid x]=\nabla F_i(x)$ and
			$\mathbb{E}[\|g_i(x;\xi)-\nabla F_i(x)\|^2]\le \sigma^2$. In addition, all stochastic gradient noises $g_i(x;\xi)-\nabla F_i(x)$ are independent across different clients $i$.
		\end{assumption}
		\begin{assumption}[Shared random subspace]\label{ass:A3}
			In each outer round $k$, all clients use the same orthonormal
			$P_k\in O(\mathbb{R}^{r\times d})$, with $P_kP_k^\top=I_r$ and
			$\mathbb{E}_{P_k}[P_k^\top P_k]=(r/d)I_d$.
		\end{assumption}

		Note that you should not introduce bounded gradient assumption or assumptions on data heterogeneity. You should try your best to give the correct proofs and generate a latex file as a final report of the task. The report should be written like the theory section of an academic paper.
	\end{NBLLMBox}
    \caption{Prompt for generating proof reports for evaluation.}
	\label{fig:prompt-proof-paper2}
\end{figure}

\subsubsection{Evaluation Details}
In this subsection, we provide detailed implementation and results for evaluating the proof reports generated by ReasFlow and baselines. The complete prompt for the \texttt{ProofReviewAgent} is as presented in Fig.~\ref{fig:prompt-proof-review}. The detailed assessment including scores for each different dimension is as listed in Table~\ref{tab:prover_scores_detail}.

\newpage
\begin{center}
	\begin{LLMBox}[width=\textwidth, fontupper=\small]{System prompt for the \texttt{ProofReviewAgent}}
		\begin{markdown}
## Role

You are a top-tier machine learning theoretical scientist and a senior reviewer for prestigious academic conferences such as NeurIPS and ICML. You are renowned for your extremely rigorous, objective, and fastidious academic attitude.

## Task

Your task is to evaluate the quality of a theoretical proof generated by an AI. I will provide the [Problem Definition \\& Settings], [Target Algorithm \\& Conclusion], and the [AI-Generated Proof Process]. You must meticulously scrutinize every step of the proof and provide a comprehensive score (0-100) along with a detailed review report.

## Evaluation Criteria

Please strictly follow the four dimensions below for review and scoring. Note that mathematical proofs require absolute rigor; any single fatal logical gap may render the entire proof invalid.

### Dimension 1: Academic Formalism \\& Formatting (10 pts)

* **Full Score Standard**: Possesses the standard proof structure of an academic paper (e.g., clear Lemmas, Theorems, Proof environments, and professional LaTeX formatting).
* **Deductions**: If the response is merely a wall of text lacking mathematical formulas or fails to provide a basic proof framework (e.g., "Proof:", "Q.E.D."), deduct all points for this dimension.

### Dimension 2: Target Alignment \\& Relevance (10 pts)

* **Full Score Standard**: The proven conclusion, loss functions used, and algorithmic update steps are perfectly consistent with the [Problem Definition \\& Settings].
* **Deductions**: If the proof conclusion is irrelevant to the requested algorithm or setting (e.g., proving a standalone SGD when Federated Learning was required, or assuming convexity when a non-convex setting was specified), assign 0 points for this dimension.

### Dimension 3: Consistency of Assumptions \\& Notation (10 pts)

* **Full Score Standard**: Strictly uses only the provided assumptions (e.g., L-smoothness, bounded variance); notations must be consistent throughout the text and clearly defined.
* **Deductions**:
* Introducing undeclared strong assumptions (e.g., "assuming bounded gradients" or "assuming i.i.d. data"): Deduct 5 points per occurrence.
* Inconsistent notation or using undefined variables (e.g., using $x$ in one step and suddenly switching to $w$): Deduct 1-3 points depending on severity.

### Dimension 4: Logical Deduction \\& Correctness (50 pts) [Core Dimension]

Examine the formula derivations line-by-line. LLMs frequently "hallucinate" in this area. Categorize logical leaps into the following three types:

* **Trivial Skips**: Basic algebraic simplifications, combining like terms, or applying standard inequalities (e.g., Cauchy-Schwarz, Peter-Paul). No deduction.
* **Suspicious Skips**: Consolidating several complex equations into a non-trivial new expression without intermediate steps or cross-term analysis; or claiming "it is easy to see" or "similarly" for steps that are not immediately obvious. Deduct 5-10 points per "Suspicious Skip." If the proof contains excessive suspicious leaps that make verification impossible, the score for this dimension cannot exceed 20 points.
* **Explicit Errors / Fake Proofs**: Incorrect inequality directions, expectation operators penetrating non-linear functions, matrix dimension mismatches, or "fake proofs" intentionally fabricated to reach the target result.
Any single "Explicit Error" results in 0 points for this dimension.

### Dimension 5: Theoretical Significance \\& Strength (20 pts)

* **Criteria**: Evaluate the "Price-Performance Ratio" of the theorem. A high-quality proof should aim for the **strongest possible conclusion** using the **weakest possible assumptions**.
* **Scoring Logic**:
* **Strong Conclusion**: Does it provide a tight convergence rate (e.g., $O(1/\epsilon^2)$ vs $O(1/\epsilon)$)? Does it handle the most general case (e.g., non-convex)?
* **Weak Assumptions**: Does it avoid restrictive assumptions like "Bounded Gradients" or "Strong Convexity"?
* **Penalty**: If the proof achieves the goal by adding "trivializing" assumptions (making the problem too easy), deduct points from the total score.

## Output Format

Please output and save your review results in JSON format to ensure it can be parsed:

```
{
"Review_Process": {
"Step_by_Step_Critique": "Analyze the proof process line-by-line here. Explicitly identify which steps are normal, which are suspicious skips, and which are explicit errors. You must include specific formula references.",
"Assumption_Check": "Analyze whether the author abused assumptions or omitted critical given assumptions.",
"Alignment_Check": "Analyze whether the proof truly addresses the specified problem and algorithm.",
"Strength_Check": "Analyze the theoretical significance of the provided theorem."
},
"Scores": {
"Dim1_Formalism": [0-10],
"Dim2_Relevance": [0-10],
"Dim3_Consistency": [0-10],
"Dim4_Logical_Rigor": [0-50],
"Dim5_Strength": [0-20]
},
"Total_Score": [0-100],
"Final_Verdict": "Accept / Weak Reject / Strong Reject (A one-sentence summary of whether the proof is credible)"
}
```

Use the write() tool provided to save the above JSON review to a file if the user provides a file path.
\end{markdown}
	\end{LLMBox}
	\captionof{figure}{System prompt for the \texttt{ProofReviewAgent}.}
	\label{fig:prompt-proof-review}
\end{center}

\begin{table}[htbp]
	\centering
	\caption{Detailed Scores of the theoretical reports reviewed by the \texttt{ProofReviewAgent}. S1, S2, S3 corresponds to evaluations performed with \texttt{GPT-5.1}, \texttt{GPT-5.4} and \texttt{Gemini-3-Pro} as the reviewer's LLM backend, respectively. \textbf{Bold}: best; \underline{underlined}: second best.}
	\label{tab:prover_scores_detail}
	\newcolumntype{Z}{>{\centering\arraybackslash}p{1.4em}}
	\resizebox{\textwidth}{!}{
		\begin{tabular}{l|ZZZ|ZZZ|ZZZ|ZZZ|ZZZ}
			\toprule
			\multirow{2}{*}{AI Agent} & \multicolumn{3}{c|}{Dim 1 Score (10)} & \multicolumn{3}{c|}{Dim 2 Score (10)} & \multicolumn{3}{c|}{Dim 3 Score (10)} & \multicolumn{3}{c|}{Dim 4 Score (50)} & \multicolumn{3}{c}{Dim 5 Score (20)} \\
			\cmidrule(lr){2-4} \cmidrule(lr){5-7} \cmidrule(lr){8-10} \cmidrule(lr){11-13} \cmidrule(lr){14-16} & S1 & S2 & S3 & S1 & S2 & S3 & S1 & S2 & S3 & S1 & S2 & S3 & S1 & S2 & S3 \\
			\midrule
			ChatGPT (\texttt{GPT-5.3}) &  \underline{8} & 8 & \underline{2} & 4 & 2 & \underline{0}  & 5  & 1  & \underline{5}  & 0  & 0  & 0  & 12 & 4  & 0  \\
			Gemini (\texttt{Gemini-3.1-Pro}) & \textbf{10} & \underline{9} & \textbf{10} & \underline{9} & 6 & \textbf{10} & 8  & 3  & \textbf{10} & 38 & 0  & 0  & 16 & 6  & \underline{18} \\
			ReasLingo (\texttt{Gemini-3-Pro})  & \textbf{10} & \underline{9} & \textbf{10} & \underline{9} & \textbf{8} & \textbf{10} & \underline{9}  & 3  & \textbf{10} & 45 & 18 & 0  & \textbf{18} & 10 & 10 \\
			ReasLingo (\texttt{GPT-5.1}) & \textbf{10} & \underline{9} & \textbf{10} & \underline{9} & 6 & \textbf{10} & \underline{9}  & 5  & \textbf{10} & 42 & 8  & 0  & \underline{17} & 7  & 10 \\
			ReasLingo (\texttt{GPT-5.4}) & \textbf{10} & \underline{9} & \textbf{10} & \underline{9} & 5 & \underline{0}  & \textbf{10} & 5  & \textbf{10} & 46 & \underline{32} & \underline{40} & \underline{17} & 6  & 0  \\
			ProofGrader (\texttt{GPT-5.1}) & \textbf{10} & \underline{9} & \textbf{10} & \underline{9} & 2 & \underline{0}  & \textbf{10} & 3  & \textbf{10} & 46 & 16 & \textbf{50} & \underline{17} & 4  & 0  \\
			ProofGrader (\texttt{GPT-5.4}) & \textbf{10} & \underline{9} & \textbf{10} & \underline{9} & \underline{7} & \textbf{10} & \textbf{10} & 3  & \textbf{10} & 46 & 12 & \underline{40} & \textbf{18} & 8  & 5  \\
			QED (\texttt{GPT-5.4}) & \textbf{10} & \textbf{10} & \textbf{10} & \textbf{10} & 2 & \underline{0}  & \textbf{10} & \textbf{9}  & \textbf{10} & \textbf{50} & 24 & \textbf{50} & \textbf{18} & 6  & 0  \\
            \midrule
			ReasFlow (\texttt{Gemini-3-Pro})   & \textbf{10} & \underline{9} & \textbf{10} & \underline{9} & \textbf{8} & \textbf{10} & \underline{9}  & 6  & \textbf{10} & 45 & 24 & \textbf{50} & \textbf{18} & \textbf{15} & \textbf{20} \\
			ReasFlow (\texttt{GPT-5.1}) & \textbf{10} & \underline{9} & \textbf{10} & \underline{9} & \textbf{8} & \textbf{10} & \textbf{10} & 5  & \textbf{10} & 46 & 16 & \textbf{50} & \textbf{18} & 7  & \textbf{20} \\
			ReasFlow (\texttt{GPT-5.4}) & \textbf{10} & \textbf{10} & \textbf{10} & \underline{9} & \textbf{8} & \textbf{10} & \textbf{10} & \underline{8}  & \textbf{10} & \underline{47} & \textbf{38} & \textbf{50} & \textbf{18} & \underline{14} & \textbf{20} \\
			\bottomrule
		\end{tabular}
	}
\end{table}

{\color{black}
\subsection{Evaluation Details for Sec.~\ref{subsubsec:exp-exp}}\label{app:exp-exp}
\subsubsection{Generation Details}
We evaluate the agents by having them continue to design and implement experiments based on their outputs from Sec.~\ref{subsubsec:exp-alg}. The generation prompts are provided in Fig.~\ref{fig:prompt-alg-chain2-r} (for ReasFlow) and Fig.~\ref{fig:prompt-alg-chain2} (for the baselines). We provide the baselines with more detailed requirements to ensure they can produce comparable results.

\begin{center}
	\begin{LLMBox}[width=\textwidth, fontupper=\small]{Prompt for experimental tests (ReasFlow).}
		\begin{markdown}
You are given an algorithm design report and implementation script for the following research.

**Task:** \\{task\\}

**Research Idea:** \\{idea\\}

**Algorithm Design Report (summary):**
\\{`alg_report`[:3000]\\}

**Algorithm Script Location:** \\{`alg_script_hint`\\}

Please work in AUTO MODE:
1. Read the algorithm design report and script carefully
2. Design a comprehensive experimental plan (`experiment_plan.md`)
3. Execute all planned experiments
4. Generate `experiment_report.md` with full results and analysis
5. Create publication-quality figures

Complete all experiments without waiting for user confirmation.
Time limit: \\{N\\} hours.
\end{markdown}
	\end{LLMBox}
	\captionof{figure}{ReasFlow prompt for experimental tests.}
	\label{fig:prompt-alg-chain2-r}
\end{center}

\begin{center}
	\begin{LLMBox}[width=\textwidth, fontupper=\small]{Prompt for experimental tests (Baseline).}
		\begin{markdown}
You are an expert ML researcher. You previously designed an algorithm.
Now you must ACTUALLY EXECUTE experiments to evaluate it.

The algorithm design report and implementation code are available in the `document/` and `code/` directories
of your current working directory:
- `document/algorithm_design.md` — the algorithm design report
- `code/algorithm.py` — the algorithm implementation

## STEP-BY-STEP INSTRUCTIONS (follow in order):

### Step 1: Install dependencies
Run: `pip install numpy scipy torch matplotlib scikit-learn pandas`
Install any additional packages required by `code/algorithm.py`.

### Step 2: Verify the algorithm runs
Run: `cd code && python algorithm.py`
If it fails, fix the import errors or missing dependencies, then re-run.

### Step 3: Write experiment scripts
Create `code/run_experiments.py` that:
- Imports and instantiates the algorithm from `algorithm.py`
- Defines 2-3 baseline methods for comparison (e.g., random search, grid search, SGD)
- Runs all methods on 2-3 test problems (e.g., Rosenbrock, Rastrigin, or task-relevant benchmarks)
- Records metrics: convergence speed, final objective value, runtime
- Saves results to `results/` directory as CSV or JSON

### Step 4: Run experiments
Execute: `cd code && python run_experiments.py`
Capture all output and results.

### Step 5: Write experiment report
Save `document/experiment_report.md` containing:
1. **Experimental Setup** — datasets/problems, metrics, baselines, hyperparameters used
2. **Implementation Details** — how experiments were set up and run
3. **Main Results** — performance tables with ACTUAL numbers from your runs
4. **Ablation Studies** — effect of key hyperparameters
5. **Analysis and Discussion** — why the method works (or doesn't)

CRITICAL RULES:
- You MUST actually run code and report real numbers. Do NOT fabricate results.
- If a script fails, debug it and fix it before reporting.
- All numerical results must come from actual execution output.
- Create the `results/` directory if needed: `mkdir -p results`

---

Task: \\{task\\}
Research Idea: \\{idea\\}
\end{markdown}
	\end{LLMBox}
	\captionof{figure}{Baseline prompt for experimental tests.}
	\label{fig:prompt-alg-chain2}
\end{center}

\subsubsection{Evaluation Details}
We use both \texttt{GPT-5.4} and \texttt{Claude Sonnet 4.6} to evaluate the outputs, with three independent runs per model. For each dimension, the six scores are averaged to obtain a task-level score, and the reported system score is then averaged across tasks. Hallucination Suspicion is positive for a task if any of the six runs raises the flag. The prompts provided in Fig.~\ref{fig:prompt-alg-chain2-e} are adapted from MLR-Bench, with metrics we propose marked as \texttt{[NEW]}.

\begin{center}
	\begin{LLMBox}[width=\textwidth, fontupper=\small]{Prompt for evaluating experimental results (abridged).}
		\begin{markdown}
You will be given an experimental document produced by an AI agent. Unlike the standard MLR-Bench setup, this agent received an algorithm design report and script (not a full research proposal) as input, and was responsible for BOTH autonomously designing the experimental plan AND executing the experiments. Therefore, please evaluate both the quality of the experimental design (which the agent produced autonomously) and the quality of the execution results. 
    
Please evaluate along all dimensions below (1-10 scale) and respond in valid JSON.

---

## EVALUATION DIMENSIONS

### [NEW] DESIGN SOUNDNESS (1-10) 

How rigorous and well-structured is the experimental plan that the agent autonomously designed? Consider: appropriateness of baselines, clarity of evaluation metrics, inclusion of ablation studies, and internal logic of the plan.

9-10 - The experimental design is highly rigorous, with appropriate baselines, evaluation metrics, and ablation studies clearly defined.

7-8  - The design is mostly sound with only minor gaps (e.g., one missing baseline).

5-6  - The design has some weaknesses but covers the main experimental objectives.

3-4  - The design has significant gaps or logical flaws in the experimental plan.

1-2  - The experimental design is absent or fundamentally unsound.

### [NEW] DESIGN COMPLETENESS (1-10)

Does the experimental plan include all necessary components: baseline comparisons, appropriate evaluation metrics, and ablation studies?

9-10 - All necessary components are present and well-defined (baselines, metrics, ablations, statistical significance considerations). 

7-8  - Most components are present with minor omissions (e.g., no ablation study). 

5-6  - Some important components are missing (e.g., no baseline or no ablation). 

3-4  - Many key components are absent; plan is skeletal. 

1-2  - The experimental plan is severely incomplete or nonexistent.

### [NEW] ALGORITHM-EXPERIMENT ALIGNMENT (1-10)

Does the experimental design fully leverage the core contribution of the provided algorithm script and design report? Is there a clear logical connection between the algorithm and the experimental setup?

9-10 - The experiments are perfectly tailored to demonstrate the algorithm's core contribution; every experiment directly tests a key claim.

7-8  - Strong alignment with only minor disconnects (some tangential experiments).

5-6  - Partial alignment; some experiments are not directly motivated by the algorithm.

3-4  - Weak alignment; experiment and algorithm feel largely independent.

1-2  - No meaningful connection between the algorithm and experimental design.

### SOUNDNESS (1-10) [Most Important]

Are the experimental results based on REAL execution rather than synthetic/fabricated data? This is the most critical dimension. Check result files (`results/*.csv`,
`results/*.json`, `data/*.csv`, `data/*.json`) AND execution logs for evidence of actual runs.
Note: some agents output results directly to a `results/` directory rather than logs.

**MANDATORY RULE - apply before any other scoring:** If BOTH of the following are true: (a) `execution_artifacts` shows [NO DATA FILES FOUND] (no CSV/JSON result files exist), AND (b) log shows [NO LOG FOUND] or contains no actual execution traces (no stack traces, no printed values, no timing information - just setup text), THEN: Soundness MUST be scored 1-2, and `has_hallucination` MUST be true. A polished `experiment_report.md` without any execution evidence is fabricated by definition.

9-10 - Results clearly stem from genuine execution with verifiable logs and outputs.

7-8  - Results appear genuine with minor documentation gaps.

5-6  - Mostly genuine but with some suspicious or unverifiable claims.

3-4  - Significant portions appear fabricated or hallucinated.

1-2  - Results are clearly synthetic/fabricated; no genuine experiments were run.

### CONSISTENCY (1-10)

Does the execution faithfully follow the experimental plan? Are the reported results consistent with the experimental setup described?

9-10 - Execution precisely matches the plan; all reported numbers are traceable.

7-8  - Mostly consistent with minor deviations (well-explained).

5-6  - Some inconsistencies between plan and execution.

3-4  - Significant gaps between stated plan and actual execution.

1-2  - Execution is entirely inconsistent with or ignores the plan.

### COMPLETENESS (1-10)

Were all designed experiments actually completed and reported?

9-10 - All planned experiments completed; all results reported with full detail.

7-8  - Most experiments completed; minor omissions explained.

5-6  - Roughly half of planned experiments completed.

3-4  - Only a small fraction of planned experiments completed.

1-2  - Almost no experiments completed or reported.

### NOVELTY (1-10)

How novel and creative is the experimental design in terms of evaluation methodology?

9-10 - Innovative evaluation approach that provides unique insights.

7-8  - Some novel evaluation choices beyond standard benchmarks.

5-6  - Standard experimental design; nothing unusual.

3-4  - Very conventional; misses opportunities for deeper evaluation.

1-2  - Trivial or copied experimental design.

### INSIGHTFULNESS (1-10)

How deep and thoughtful is the analysis of the experimental results?

9-10 - Rich analysis connecting results to algorithmic mechanisms; identifies failure modes and proposes explanations.

7-8  - Good analysis with meaningful observations.

5-6  - Adequate analysis; mostly describes results without deeper interpretation.

3-4  - Shallow analysis; mostly just reports numbers.

1-2  - No meaningful analysis; results are presented without any interpretation.

### SIGNIFICANCE (1-10)

Based on the experimental results, how significant is the algorithm's demonstrated contribution to the field?

9-10 - Results demonstrate substantial improvements over baselines with clear statistical significance.

7-8  - Meaningful improvements demonstrated across most metrics/datasets.

5-6  - Modest improvements; results are mixed across conditions.

3-4  - Negligible improvements; algorithm barely outperforms baselines.

1-2  - Algorithm underperforms baselines; results are negative.

### [NEW] EXPERIMENTAL EFFECTIVENESS (1-10)

Does the experimental evidence actually support or meaningfully test the algorithm's claims? This dimension evaluates the Claim-to-Evidence loop: whether the experiments produce results that constitute genuine evidence (positive or negative) for the algorithm's stated contributions.

NOTE: Honestly reporting negative results (algorithm underperforms) is NOT penalized if the experimental design was sound and the analysis is insightful. What matters is whether the experiments were designed to TEST the claims and whether the results INFORM the reader about the claims' validity.

9-10 - Every major claim has direct experimental evidence; results clearly confirm or refute each claim with appropriate metrics and baselines; ablation studies isolate individual contributions.

7-8  - Most claims are supported by experimental evidence; one or two claims lack direct evidence; overall the experiments convincingly test the core contribution.

5-6  - Some claims have supporting evidence but others are tested only indirectly or with insufficient baselines; the evidence-claim mapping is partial.

3-4  - Weak connection between claims and evidence; experiments run but results do not clearly inform any specific claim; generic benchmarking without claim focus.

1-2  - No discernible connection between the algorithm's claims and the experimental results; experiments appear arbitrary or disconnected from the contribution.

### [NEW] WRITEUP READABILITY (1-10)

How well-structured is this experiment report for a downstream writing agent to produce the Experiments section of a research paper? Evaluate whether the report provides all elements needed for paper writing.

Consider:
- Does the report follow a clear structure (Experimental Setup, Baselines, Metrics, Main Results, Ablation Studies, Analysis/Discussion)?
- Are data tables formatted with mean +/- std across runs?
- Are figures generated with proper captions, axis labels, and legends that are paper-ready?
- Is reproducibility information included (hyperparameters, seeds, compute environment)?
- Are result claims explicitly tied to specific table/figure references?
- Does the report include a clear comparison narrative (what beats what and by how much)?
- Would a writing agent be able to extract a complete Experiments + Results section?

9-10 - Excellent: Report is publication-ready; structured sections, formatted tables with mean+/-std, captioned figures, full reproducibility info.

7-8  - Good: Most elements present; minor formatting or structural gaps.

5-6  - Moderate: Basic results present but significant gaps in formatting, tables, or figure quality for paper writing.

3-4  - Weak: Report is mainly unstructured text with inline numbers; writing agent would need major reformatting.

1-2  - Poor: No usable structure; raw output dumps or missing key sections.

---

## HALLUCINATION CHECK

**Step 1 - Hard rule (check first):**
If `execution_artifacts` = [NO DATA FILES FOUND] AND log = [NO LOG FOUND] or log contains
no execution traces (no output values, no timing, no errors):
→ set `has_hallucination` = true, Soundness = 1-2. Do NOT let a polished report override this.

**Step 2 - Soft checks (apply when Step 1 does not trigger, i.e., evidence files DO exist):**
IMPORTANT: When result files exist, the bar for flagging hallucination is HIGH.
Stylistic gaps, narrative imprecision, or partial report coverage are NOT hallucination.
Only flag `has_hallucination` = true here if:
- The KEY numerical claims in the report (e.g., "our method achieves 92\% accuracy")
  directly contradict the actual values in the result files (e.g., files show 61\%), OR
- The result files contain clearly synthetic/templated data (identical values across all
  conditions, suspiciously round numbers like exactly 0.5000, 1.0000 everywhere), OR
- The report claims results for datasets/baselines that are entirely absent from all files.

Do NOT flag hallucination merely because:
- The report narrative is incomplete or partially inconsistent with the experimental plan
- Some secondary metrics or ablations in the report lack corresponding files
- The evaluation setup differs from what the task description suggested

---

## INPUT

**Task Description:**
`{task}`

**Research Idea:**
`{idea}`

**Algorithm Design Report:**
`{algorithm_report}`

**Algorithm Script (`code/algorithm.py` summary or key excerpts):**
`{algorithm_script}`

**Experiment Report (`document/experiment_report.md`):**
`{experiment_report}`

**Execution Log (`logs/` or `log.txt`, truncated if long; may show [NO LOG FOUND] if agent stored outputs in result files instead):**
`{log}`

**Execution Artifacts (data files / CSV results / figures generated during experiments):** `{execution_artifacts}`

---

OUTPUT FORMAT

Respond with ONLY valid JSON matching this schema:

    {
      "Hallucination": {
        "has_hallucination": <true/false>,
        "details": "Evidence of fabrication or confirmation of genuine results"
      },
      "DesignSoundness":              {"score": <1-10>, "justification": "..."},
      "DesignCompleteness":           {"score": <1-10>, "justification": "..."},
      "AlgorithmExperimentAlignment": {"score": <1-10>, "justification": "..."},
      "Soundness":                    {"score": <1-10>, "justification": "..."},
      "Consistency":                  {"score": <1-10>, "justification": "..."},
      "Completeness":                 {"score": <1-10>, "justification": "..."},
      "Novelty":                      {"score": <1-10>, "justification": "..."},
      "Insightfulness":               {"score": <1-10>, "justification": "..."},
      "Significance":                 {"score": <1-10>, "justification": "..."},
      "ExperimentalEffectiveness":    {"score": <1-10>, "justification": "..."},
      "WriteupReadability":           {"score": <1-10>, "justification": "..."},
      "OverallAssessment": {
        "score": <1-10>,
        "strengths": ["...", "..."],
        "weaknesses": ["...", "..."]
      },
      "Confidence": <1-5>
    }
\end{markdown}
	\end{LLMBox}
	\captionof{figure}{Prompt for judging experimental results (abridged).}
	\label{fig:prompt-alg-chain2-e}
\end{center}

}

\subsection{Evaluation Details for Sec.~\ref{subsubsec:exp-intro}}\label{app:exp-intro}

To evaluate the quality of the \textit{Introduction} section generated by the \texttt{IntroductionAgent}, we compare it against a base model (\texttt{GPT-4o}) baseline across three papers. For each configuration, we perform five independent runs and report mean scores. The evaluation employs three complementary metrics, each scored on a 0-1 scale:

\begin{itemize}
	\item \textbf{Content Coverage}: measures how well the introduction captures the reference material. It is a weighted combination of raw weighted coverage (weight 0.34), anchor coverage for main results and theorems (0.18), core scientific path coverage for method and evidence chains (0.22), structure alignment for the problem$\to$gap$\to$method$\to$evidence flow (0.26), and a survey dominance penalty ($-0.04$) that penalizes overly background-heavy introductions.
	\item \textbf{Faithfulness}: quantifies hallucination avoidance, defined as $1 - (\text{\#unsupported claims} / \text{\#total claims})$. Claims are categorized into common background (lenient), paper-specific (strict), and overclaim/speculation (very strict) thresholds.
	\item \textbf{Content Quality}: a rubric-based score assessing six dimensions, domain/application emphasis, algorithm mechanism or formula details, complete background description, core research idea clarity, explicit research goals/contributions, and concrete experimental or theoretical depth.
\end{itemize}

\subsubsection{Generation Details}

Each paper's upstream agent outputs, including the experiment report, theoretical proof report, survey/related work draft, and algorithm description, serve as the shared reference material from which both ReasFlow and the baseline must compose the introduction.

\noindent\textbf{ReasFlow (\texttt{IntroductionAgent}).} The \texttt{IntroductionAgent} first collects structured information from all upstream agent outputs using dedicated extraction tools that parse survey reports, method descriptions, experimental results, and theoretical contributions into structured summaries, grounding all content in source materials rather than relying on the LLM's parametric knowledge. The agent then composes the introduction and enters an automated \emph{evaluation--refinement} loop: a built-in evaluation tool assesses the draft along three quality dimensions (Coverage, Faithfulness, Content Quality), and a refinement tool performs targeted local edits on low-scoring aspects while preserving high-quality content. This loop iterates until scores plateau or a maximum iteration count is reached.

\noindent\textbf{Baseline (\texttt{GPT-4o}).} The baseline receives the same upstream assets, the complete set of \texttt{.tex} files from the prover, experiment, survey, and algorithm agents, concatenated as context, together with a single prompt instructing the model to write a self-contained introduction. No structured extraction, automated evaluation, or iterative refinement is performed; the baseline produces the introduction in a single inference call.

\noindent\textbf{Runs and reproducibility.} For each of the three papers, five independent runs are performed for both ReasFlow and the baseline to account for stochastic variation, and results are averaged.

Before evaluation, two preprocessing steps transform each generated introduction and its reference material into structured representations amenable to automated scoring.

\textbf{Reference Preprocessing} (\texttt{preprocess\_refe.py}): All reference \texttt{.tex} files from the paper (experiment reports, theoretical proofs, related work sections) are fed to an LLM, which performs a structured decomposition into three categories, \textit{experiment\_results}, \textit{prover\_results}, and \textit{related\_work\_claims}. A second LLM call then classifies each item's importance as \texttt{critical} (directly supports the main conclusion), \texttt{important} (supports the conclusion but is not the most central), or \texttt{supplementary} (minor details not essential for the introduction).

\textbf{Prediction Claim Extraction} (\texttt{preprocess\_pred.py}): The generated introduction together with its bibtex file \texttt{references.bib} is parsed by an LLM to extract all explicit contribution and factual statements as minimal semantic units (\textit{prediction\_claims}), each classified as either a ``contribution'' or ``fact'' type. These structured representations, the categorized reference items and the extracted prediction claims, serve as inputs to the evaluation stage described below.

\subsubsection{Evaluation Details}

The evaluation proceeds in three stages. 

\textbf{Stage~1 (Reference Preprocessing)}: All reference \texttt{.tex} files from the paper are fed to an LLM, which extracts three structured categories, \textit{experiment\_results}, \textit{prover\_results}, and \textit{related\_work\_claims}, each annotated with an importance label (\texttt{critical}, \texttt{important}, or \texttt{supplementary}) based on its role in supporting the paper's main conclusions.

\textbf{Stage~2 (Claim Extraction)}: The generated introduction is parsed by an LLM to extract all explicit contribution and factual statements as minimal semantic units (\textit{prediction\_claims}). 

\textbf{Stage~3 (Evaluation)}: Two LLM calls are made, one for joint Coverage \& Faithfulness assessment, and one for Content Quality, using the structured references and extracted claims as context.

\noindent\textbf{Coverage scoring.} The final coverage score is a weighted combination of four sub-metrics:
\begin{equation*}
	\text{Coverage} = 0.34 \cdot C_\text{raw} + 0.18 \cdot C_\text{anchor} + 0.22 \cdot C_\text{core} + 0.26 \cdot C_\text{struct} - 0.04 \cdot P_\text{survey},
\end{equation*}
where $C_\text{raw}$ is the importance-and-category-weighted coverage (critical items weight 1.0, important 0.6, supplementary 0.0; experiment results scaled $\times 1.15$, prover results $\times 1.10$, related work $\times 0.55$); $C_\text{anchor}$ is the coverage rate of anchor items (main/core results); $C_\text{core}$ weights only experiment and prover results to capture the method-to-evidence chain; $C_\text{struct}$ measures alignment with the problem$\to$gap$\to$method$\to$contribution$\to$ evidence narrative arc; and $P_\text{survey}$ penalizes overly background-heavy introductions.

The complete prompts for the two evaluation calls are presented in Fig.~\ref{fig:prompt-intro-cov} and Fig.~\ref{fig:prompt-intro-cq}.

\begin{center}
	\begin{LLMBox}[width=\textwidth, fontupper=\small]{System Prompt for Coverage \& Faithfulness Evaluation}
\begin{markdown}
You are a rigorous and meticulous LLM-as-a-judge specialized in evaluating scientific introductions.

**Coverage (0.00--1.00)**
The proportion of reference information (experimentResults, proverResults, relatedWorkClaims) meaningfully covered by the introduction.
- Mark an item as **covered** when the introduction captures its substantive meaning, even if compressed, paraphrased, or merged with nearby claims.
- Do NOT require lexical overlap or one-to-one sentence matching.
- For method/theory/result items, a conservative summary counts as covered even if minor details are omitted.
- Reward introductions preserving the chain: problem/background, limitation/gap, method/contribution, evidence preview.

**Faithfulness (0.00--1.00)**
Faithfulness = 1 - (unsupportedClaims / totalClaims).
Classify each prediction claim before judging:
1. commonBackground: standard field knowledge,  faithful unless contradicted;
2. paperSpecific: claims about method, theory, experiments,  require direct support or conservative inference;
3. overclaimOrSpeculation: mark unfaithful if contradicted, overstated, numerically unsupported, or unreasonably inferred.

**Output Format (JSON)**
Return coverage (float in [0, 1]), faithfulness (float in [0, 1]), rationale, claimFaithfulnessDetails (list of claimIndex, claimText, status), and referenceCoverageDetails (experimentResults, proverResults, relatedWorkClaims with index, descriptionExcerpt, importance, status).
\end{markdown}
	\end{LLMBox}
	\captionof{figure}{System prompt for the joint Coverage \& Faithfulness evaluation of generated introductions.}
	\label{fig:prompt-intro-cov}
\end{center}

\begin{center}
	\begin{LLMBox}[width=\textwidth, fontupper=\small]{System Prompt for Content Quality Evaluation}
		\begin{markdown}
You are a rigorous LLM-as-a-judge focusing on **Content Quality**. Apply strict standards.

**Content Quality (0.00--1.00)**
Score based on the following dimensions (lacking any dimension incurs deductions):

1. **Domain emphasis**: Does the introduction clearly identify the research/application domain?
2. **Algorithm/formula detail**: Does it include key algorithmic steps or core mathematical formulas with correct explanations?
3. **Background completeness**: Is the background coherent and sufficient to set up the subsequent content?
4. **Core idea clarity**: Does it clearly present the paper's core method and innovation?
5. **Research goals**: Are research objectives and contributions explicitly stated and verifiable?
6. **Concrete depth**: Does it include specific experimental settings/results or theoretical conclusions, avoiding vague generalities?

**Bonus criteria** (should significantly raise the score if present):
- Deep discussion of mathematical principles, model assumptions, derivations, or convergence analysis.
- Detailed description of algorithm steps, optimization strategies, or evaluation design.

**Not scored:**
Literature review completeness (for reference only, not factored into contentQuality).

Output JSON: contentQuality (float in [0,1]), rationale ("brief explanation").
\end{markdown}
	\end{LLMBox}
	\captionof{figure}{System prompt for the Content Quality evaluation of generated introductions.}
	\label{fig:prompt-intro-cq}
\end{center}

\noindent Two composite scores are computed as weighted sums of the six sub-metrics (raw coverage, anchor coverage, core path coverage, structure alignment, faithfulness, and content quality). The \textit{Overall Balanced} score distributes weights evenly across coverage sub-metrics (0.10 each) while giving moderate emphasis to faithfulness (0.28) and content quality (0.32). The \textit{Overall Technical Focus} score up-weights core scientific path coverage (0.15) and content quality (0.40) to prioritize method and evidence depth over background breadth.

\noindent To assess evaluation robustness, we employ three independent evaluation models (\texttt{GPT-5.4}, \texttt{Claude Opus}, and \texttt{Gemini}) and report per-judger results, each averaged over 5 independent runs. The results are summarized in Table~\ref{tab:intro_scores_detail}. The \texttt{IntroductionAgent} consistently outperforms the base model across all metrics and all three papers. The most substantial improvement is observed in Content Quality, with an average gain of +52.3\% and a peak of \textbf{+121.4\%} on \paperthree (Opus judger), indicating that the agent produces introductions with significantly richer technical depth. Faithfulness also improves on average (+7.7\%), though individual judgers vary: GPT-5.4 and Opus report large gains, while Gemini, which already assigns high faithfulness to the base model, shows smaller margins. Notably, the improvements are most pronounced on \paperthree, where the base model struggles the most, suggesting that the \texttt{IntroductionAgent} scales better on harder papers.

\begin{table}[htbp]
	\centering
	\caption{Per-paper evaluation of the \texttt{IntroductionAgent} vs.\ the base model (\texttt{GPT-4o}) across three evaluation models (\texttt{GPT-5.4}, \texttt{Claude Opus}, \texttt{Gemini}). Each cell reports the mean over 5 independent runs. For each metric, the best ReasFlow score and the best advantage are highlighted in \textbf{bold}.}
	\label{tab:intro_scores_detail}
	\resizebox{\textwidth}{!}{
		\begin{tabular}{ll|ccc|ccc|ccc|ccc|ccc}
			\toprule
			\multirow{2}{*}{Paper}               & \multirow{2}{*}{Judger} & \multicolumn{3}{c|}{Coverage} & \multicolumn{3}{c|}{Faithfulness} & \multicolumn{3}{c|}{Content Quality} & \multicolumn{3}{c|}{Overall Balanced} & \multicolumn{3}{c}{Overall Tech.\ Focus}                                                                                                                                                         \\
			\cmidrule(lr){3-5} \cmidrule(lr){6-8} \cmidrule(lr){9-11} \cmidrule(lr){12-14} \cmidrule(lr){15-17}
			                                     &                         & Base                          & RF                                & Adv.                                 & Base                                  & RF                                       & Adv.             & Base  & RF             & Adv.              & Base  & RF             & Adv.             & Base  & RF             & Adv.             \\
			\midrule
			\multirow{4}{*}{\paperone}
			                                     & \texttt{GPT-5.4}                  & 0.482                         & 0.621                             & +28.9\%                              & 0.781                                 & 0.873                                    & \textbf{+11.8\%}          & 0.372 & 0.564          & +51.6\%           & 0.554 & 0.683          & +23.2\%          & 0.520 & 0.661          & +27.0\%          \\
			                                     & \texttt{Opus}                    & 0.354                         & \textbf{0.677}                    & \textbf{+91.4\%}                     & 0.820                                 & 0.880                                    & +7.3\%           & 0.320 & 0.620          & \textbf{+93.8\%}           & 0.510 & 0.739          & \textbf{+45.0\%} & 0.461 & 0.708          & \textbf{+53.6\%} \\
			                                     & \texttt{Gemini}                  & 0.489                         & 0.674                             & +37.7\%                              & 0.950                                 & \textbf{0.890}                                    & $-6.3$\%         & 0.610 & \textbf{0.830} & +36.1\%           & 0.678 & \textbf{0.795} & +17.3\%          & 0.658 & \textbf{0.788} & +19.8\%          \\
			\cmidrule(lr){2-17}
			                                     & Average                 & 0.441                         & 0.657                             & +48.8\%                              & 0.850                                 & 0.881                                    & +3.6\%           & 0.434 & 0.671          & +54.7\%           & 0.580 & 0.739          & +27.3\%          & 0.547 & 0.719          & +31.6\%          \\
			\midrule
			\multirow{4}{*}{\papertwo}
			                                     & \texttt{GPT-5.4}                  & 0.455                         & 0.492                             & +8.2\%                               & 0.766                                 & 0.942                                    & \textbf{+23.0\%}          & 0.400 & 0.490          & +22.5\%           & 0.552 & 0.643          & +16.4\%          & 0.510 & 0.603          & +18.3\%          \\
			                                     & \texttt{Opus}                    & 0.350                         & \textbf{0.519}                             & +48.3\%                              & 0.900                                 & 0.860                                    & $-4.4$\%         & 0.320 & 0.520          & \textbf{+62.5\%}           & 0.527 & 0.642          & \textbf{+21.6\%}          & 0.474 & 0.602          & \textbf{+27.0\%}          \\
			                                     & \texttt{Gemini}                  & 0.348                         & \textbf{0.519}                             & \textbf{+49.0\%}                              & 0.982                                 & \textbf{0.950}                           & $-3.3$\%         & 0.650 & \textbf{0.640}          & $-1.5$\%          & 0.656 & \textbf{0.705}          & +7.6\%           & 0.627 & \textbf{0.674}          & +7.4\%           \\
			\cmidrule(lr){2-17}
			                                     & Average                 & 0.384                         & 0.510                             & +32.7\%                              & 0.883                                 & 0.917                                    & +3.9\%           & 0.457 & 0.550          & +20.4\%           & 0.578 & 0.663          & +14.7\%          & 0.537 & 0.626          & +16.6\%          \\
			\midrule
			\multirow{4}{*}{\paperthree}
			                                     & \texttt{GPT-5.4}                  & 0.360                         & \textbf{0.495}                             & \textbf{+37.4\%}                              & 0.704                                 & 0.873                                    & \textbf{+24.1\%} & 0.304 & 0.580          & +90.8\%           & 0.465 & 0.648          & \textbf{+39.5\%}          & 0.418 & 0.622          & \textbf{+48.8\%}          \\
			                                     & \texttt{Opus}                    & 0.379                         & 0.424                             & +11.9\%                              & 0.770                                 & 0.928                                    & +20.5\%          & 0.280 & 0.620          & \textbf{+121.4\%} & 0.482 & 0.657          & +36.2\%          & 0.433 & 0.629          & +45.5\%          \\
			                                     & \texttt{Gemini}                  & 0.384                         & 0.452                             & +17.7\%                              & 0.941                                 & \textbf{1.000}                           & +6.3\%           & 0.450 & \textbf{0.780}          & +73.3\%           & 0.587 & \textbf{0.733}          & +25.0\%          & 0.548 & \textbf{0.719}          & +31.3\%          \\
			\cmidrule(lr){2-17}
			                                     & Average                 & 0.375                         & 0.457                             & +22.0\%                              & 0.805                                 & 0.934                                    & +16.0\%          & 0.345 & 0.660          & +91.5\%           & 0.511 & 0.679          & +32.9\%          & 0.466 & 0.657          & +40.9\%          \\
			\midrule
			\multicolumn{2}{l|}{Overall Average} & 0.400                   & 0.541                         & +35.3\%                           & 0.846                                & 0.911                                 & +7.7\%                                   & 0.412            & 0.627 & +52.3\%        & 0.557             & 0.694 & +24.6\%        & 0.517            & 0.667 & +29.2\%                           \\
			\bottomrule
		\end{tabular}
	}
\end{table}
\subsection{Evaluation Details for Sec.~\ref{subsubsec:exp-writer}}\label{app:exp-writer}

This appendix documents the full setup for the \texttt{WritingAgent} sub-experiment of Sec.~\ref{subsubsec:exp-writer}: the eleven compared systems and their configurations, the shared input bundle and task prompt, the four-phase evaluation pipeline, the equal-weight scoring rule, the reviewer-model ensemble, and the representative evaluation prompts. Per-dimension sub-score tables are provided at the end.

\subsubsection{Generation Details}
\noindent\textbf{Compared Systems. }
We compare the ReasFlow \texttt{WritingAgent} against ten baselines drawn from three families of paper-writing systems. The first family consists of two general-purpose coding agents, \textit{Claude Code} and \textit{Codex}, each run in two configurations: a plain \textit{vanilla} configuration with no additional domain skills, and an \textit{ARIS}~\cite{yang2026aris} configuration that augments the coding agent with a scientific-writing skill pack covering paper-structure planning, asset integration, and a compile-then-review loop. Each of the four resulting variants is run once under \texttt{GPT-5.1} and once under \texttt{GPT-5.4}, giving the eight coding-agent systems in a $2\times2\times2$ matrix. The second family is the scientific-writing agent \textit{ReasLingo}, a publicly available system on the ReasLab platform built on top of the Gemini CLI; we evaluate it twice, once with a \texttt{GPT-5.1} backend and once with \texttt{GPT-5.4}. Our own \texttt{WritingAgent} (Sec.~\ref{sec:writing_agent}) is run with a \texttt{GPT-5.1} backend, giving eleven systems in total.

The fairness of the comparison rests on giving every one of the eleven systems exactly the same input bundle that the ReasFlow \texttt{WritingAgent} itself receives when it writes \paperone. This bundle consists of an asset directory with the full upstream research artifacts (proved theorems, problem setup and algorithm pseudocode, integrated proof report, experiment code with raw numerical outputs, main experiment report, pre-generated experiment figures, literature survey, draft introduction, and reference \texttt{.bib} files); the target venue's LaTeX template, including the document class and the algorithm style dependencies needed for compilation; and the shared task prompt (reproduced below), which describes the task, the inputs, the required deliverables, the non-negotiable content-completeness requirements, and the compile-cleanliness criterion. The asset bundle and the template are byte-identical across all compared systems and identical to what the \texttt{WritingAgent} reads, and the task prompt is delivered verbatim as each system's initial instruction.

Every system is then run as a black box inside its own isolated working directory and is allowed to complete the task autonomously, producing its deliverables under a \texttt{./paper/} subdirectory. ReasLingo is the only compared system without a local CLI; we submit the same inputs through its hosted platform and collect the resulting \texttt{main.tex}, \texttt{references.bib}, and compiled PDF for evaluation. No system-specific tuning, prompt engineering, or intermediate supervision is applied, and evaluation operates on whatever deliverables each system produces when its own run terminates.

\noindent\textbf{Shared Task Prompt. }
The task prompt below is delivered verbatim to every compared system. It is the same prompt that the ReasFlow \texttt{WritingAgent} receives for \paperone, with no system-specific adaptation.

\begin{center}
	\begin{LLMBox}[width=\textwidth, fontupper=\small]{Shared task prompt provided to every compared system}
		\begin{markdown}
You are an AI agent tasked with producing a publication-ready research paper from a bundle of paper-writing assets placed in your working directory. The topic, methods, theory, experimental data, and references must all come from the assets; do not invent anything from memory or general knowledge.<br>

The working directory contains two input folders. The assets directory holds the full research materials for the paper; begin by reading its top-level README, which is the authoritative overview of the bundle and maps paper sections to their source files. The template directory holds the official LaTeX template for the target venue, including the document class, a skeleton example file, and the algorithm style dependencies. Write every deliverable inside a paper subdirectory of the working directory: the master TeX file must begin with the template's documentclass line, the references file must merge every bibliography file from the assets, and the final PDF must be produced by running latexmk on the master TeX file inside that paper subdirectory.<br>

The following content requirements are hard red lines and non-negotiable. Every algorithm environment in the assets must be preserved in the paper. Every theorem, lemma, corollary, proposition, and definition must be preserved in full; phrasing may be edited, but conditions, conclusions, constants, and assumptions must remain exact. Every proof must be preserved completely, with no "proof omitted" short-circuits; if proofs live in the appendix, the full version must appear there. Every number, table, and experiment figure from the assets must be presented faithfully, with numbers matching the raw data exactly and no fabrication or approximation. Every bibliography entry from the assets must be merged into the paper's references file, and every in-text citation must resolve to such an entry.<br>

The final PDF must compile with zero LaTeX errors and zero warnings of any kind, including undefined references, missing citations, overfull or underfull hboxes, and font warnings. Run latexmk iteratively: after each compile, read the log, fix every error and every warning, and recompile until the log is clean.<br>

Complete the entire task in a single autonomous run. Do not pause to ask for confirmation or clarification; for any uncertainty, make the most reasonable default choice and continue. Do not deliver partial results.<br>

The paper must be written in English. Keep the author block as an anonymous placeholder and do not write any real names, affiliations, emails, or acknowledgements.
\end{markdown}
	\end{LLMBox}
	\captionof{figure}{Shared task prompt delivered to all eleven compared systems, reproduced with only whitespace normalized for typesetting.}
	\label{fig:prompt-writer-shared}
\end{center}

\subsubsection{Evaluation Details}

Once every system has produced a manuscript, evaluation proceeds in four phases (one per reporting dimension). All phases consume, when relevant, a common Phase~0 artifact bundle that is extracted programmatically from each output; no phase reads another phase's output.

\noindent\textbf{Phase~0 (programmatic preparation).} For every system we recursively read the content-facing asset \texttt{.tex} files; reconstruct the full generated LaTeX source by following \texttt{\textbackslash input} and \texttt{\textbackslash include}; extract text from the generated PDF; render representative PDF pages as JPEG images for vision input; parse the LaTeX compilation log; extract citation keys from both the assets and the generated paper; and verify bibliography entries against arXiv and CrossRef. Phase~0 records only facts (sets of keys, file lists, page counts, compile error counts), never judgments.

\noindent\textbf{Phase~1 (Faithfulness, dimension~I).} A reviewer model, given the assets, the baseline-generated paper, citation-set facts, and programmatic reference/paper signals, scores five sub-dimensions on a $0$--$10$ integer scale: I1~theorem fidelity, I2~algorithm fidelity, I3~numerical fidelity, I4~figure fidelity, I5~citation fidelity. The reviewer must flag every \textit{fabrication} (statement with no asset source), \textit{distortion} (content present in assets but altered in the paper), and \textit{phantom citation} (plausible-looking but non-existent bib entry), supported by a quoted paper snippet as evidence.

\noindent\textbf{Phase~2 (Coverage, dimension~II).} A reviewer model scores five sub-dimensions: II1~theorem usage, II2~algorithm usage, II3~experiment usage, II4~figure usage, II5~citation usage. The rubric is deliberately abstract: reviewers identify landmark references and core assets from their training knowledge together with the programmatic signals extracted in Phase~0, without the rubric naming any paper-specific artifact. This keeps the evaluation generalizable across papers.

\noindent\textbf{Phase~3 (Writing Quality, dimension~III).} The reviewer is given only the generated paper and paper-side signals (no assets), so that writing quality is judged self-referentially. The submitted \texttt{references.bib} is pre-filtered to the cited subset so that systems which copy the entire source bibliography are not unfairly punished on prompt length while a majority of entries are never actually cited. Four sub-dimensions are scored: III1~clarity, III2~coherence (project focus), III3~argumentation, III4~evidence use.

\noindent\textbf{Phase~4 (Submittability, dimension~IV).} Three sub-components are scored: IV1~\textit{compile}, a programmatic check of the LaTeX compile log (number of errors, undefined references, overfull/underfull hboxes, font warnings; $10$ iff all zero; $0$ iff no PDF is produced); IV2~\textit{visual layout}, a two-step vision pipeline in which \texttt{gemini-3.1-flash-lite-preview} first produces a structured description of every PDF page (identifying placeholders, unresolved references, empty bib fields, rendering anomalies, page-level layout issues) and then the three reviewer models each score three visual sub-dimensions (IV2\_1~layout compliance, IV2\_2~rendering integrity, IV2\_3~content presentation) from those descriptions; and IV3~\textit{template compliance}, a programmatic check of the \texttt{\textbackslash documentclass} line and of the absence of banned package families. IV1 and IV3 are binary-style integer scores in $\{0,\dots,10\}$; IV2 is a real-valued mean across the three visual sub-dimensions and three reviewers.

\noindent\textbf{Scoring Rule (Equal Weight at Every Level). }
Let $S^{r}_{d,k}$ denote the integer score $\in[0,10]$ given by reviewer $r\in\{\texttt{GPT-5.1},\texttt{GPT-5.4},\texttt{Gemini-3-Pro}\}$ to sub-dimension $k$ of dimension $d\in\{\text{I, II, III, IV}\}$. Let $K_d$ be the number of sub-dimensions of $d$ ($K_\text{I}=K_\text{II}=5$, $K_\text{III}=4$, $K_\text{IV}=3$). Equal-weight aggregation proceeds in three steps:
\begin{align*}
	\bar{S}_{d,k} &= \tfrac{1}{3}\!\sum_{r} S^{r}_{d,k}, &
	S_d &= \tfrac{1}{K_d}\sum_{k=1}^{K_d} \bar{S}_{d,k}, &
	\text{Overall} &= \tfrac{1}{4}\bigl(S_\text{I} + S_\text{II} + S_\text{III} + S_\text{IV}\bigr).
\end{align*}
For Phase~IV, $S_\text{IV}$ is itself an equal-weight mean of D1, D2, D3 (each already an equal-weight mean over its own components). When no PDF is produced, D1 and D2 are both set to $0$ and only D3 contributes to $S_\text{IV}$.

\noindent\textbf{Reviewer Ensemble. }
To reduce reviewer-model bias we use three backends in parallel for every LLM-based sub-dimension: \texttt{GPT-5.1} (reviewer~1), \texttt{GPT-5.4} (reviewer~2), and \texttt{Gemini-3-Pro} (reviewer~3). Every reviewer receives the same prompt and returns a strict-JSON report with integer sub-scores and evidence-bearing findings; we aggregate by straight mean. For the Phase~4 visual stage, page descriptions are produced once by a single cheap vision describer (\texttt{gemini-3.1-flash-lite-preview}) and then consumed by the three scoring reviewers, so that vision noise does not propagate through the ensemble. Prompts are designed for cache-friendly layout: shared asset content is placed at the front of the prompt and reviewer-specific task structure at the back, which allows the three reviewer backends to reuse their prefix cache across all eleven compared systems.

\noindent\textbf{Evaluation Prompts.}
Every phase shares the same prompt layout. The prompt starts with a sequence of XML-style data blocks whose contents are filled in at run time from the Phase~0 artifacts of the system under review, and the phase-specific task description, scoring rubric, and output schema come after them. Placing the run-time data before the rubric lets the three reviewer backends reuse their prefix cache across all eleven compared systems, since only the part after the data blocks changes between phases.

Phases~1 and~2 use the same five data blocks. The $\langle\texttt{assets}\rangle$ block carries the full source materials for the paper under evaluation, concatenated into a single XML block by Phase~0 from the paper's proved theorems, problem setup and algorithm pseudocode, integrated proof report, main experiment report, and source bibliography files. The $\langle\texttt{reference\_signals}\rangle$ block carries the programmatically extracted anchor signals that Phase~0 computes directly from the assets (for example, which theorem files are declared core by the source proof report, or which citation keys appear in the draft introduction and related-work files). The $\langle\texttt{paper}\rangle$ block carries the manuscript produced by the system under review: Phase~0 reconstructs the full generated LaTeX source by following the paper's \texttt{input} and \texttt{include} statements, so that every section file appears in the block together with the paper's own references file. The $\langle\texttt{cite\_check}\rangle$ block carries the set-operation facts that Phase~0 computes between the source bibliography keys and the paper's cited keys (keys cited from the source, keys unused from the source, keys added by the paper, broken citations). The $\langle\texttt{paper\_signals}\rangle$ block carries the programmatic facts Phase~0 extracts from the paper itself, such as which figures are actually rendered and the section-heading structure. Phase~3 uses only the $\langle\texttt{paper}\rangle$ block (with its references file filtered down to the cited subset) and the $\langle\texttt{paper\_signals}\rangle$ block, so that writing quality is judged from the manuscript alone. Phase~4 uses two purpose-built blocks: $\langle\texttt{page\_descriptions}\rangle$ carries the concatenation of per-page structured descriptions produced by the vision stage below, and $\langle\texttt{compile\_facts}\rangle$ and $\langle\texttt{template\_facts}\rangle$ carry the programmatic compile-log counters and the documentclass / banned-package checks.

The phases are also distinguished by how many times each prompt is invoked. In Phases~1, 2, and~3 the prompt is sent, for each of the eleven systems under comparison, to three reviewer backends (\texttt{GPT-5.1}, \texttt{GPT-5.4}, \texttt{Gemini-3-Pro}) in parallel, giving 33 independent LLM calls per phase; the per-sub-dimension scores are then averaged across the three reviewers with equal weight. Phase~4 is a two-stage protocol. In the first stage, the cheap vision model \texttt{gemini-3.1-flash-lite-preview} is called once per rendered PDF page of each system, producing one structured description per page (several hundred calls in total across the eleven systems); the prompt used in this stage is shown in Fig.~\ref{fig:prompt-writer-phase4a}. In the second stage, those per-page descriptions are concatenated into the $\langle\texttt{page\_descriptions}\rangle$ block and the combined prompt is sent, again for each of the eleven systems, to the same three reviewer backends in parallel for a further 33 scoring calls; the rubric used in this stage is shown in Fig.~\ref{fig:prompt-writer-phase4b}. The rubric portion of each prompt, which is independent of any particular paper and constitutes the reviewer's actual standard of judgment, is reproduced in the LLM boxes that follow.

\begin{center}
	\begin{LLMBox}[width=\textwidth, fontupper=\small]{Phase~1 evaluation prompt: Faithfulness (dimension~I)}
		\begin{markdown}
Run-time data blocks (inserted at the top of the prompt before the rubric):

`<assets>` ... full research materials for the paper under evaluation ... `</assets>`<br>
`<reference_signals>` ... programmatic anchor signals from the source assets ... `</reference_signals>`<br>
`<paper>` ... every .tex file plus the references file of the manuscript produced by the system under review ... `</paper>`<br>
`<cite_check>` ... citation-set operations between source bibliography keys and the paper's cited keys ... `</cite_check>`<br>
`<paper_signals>` ... figures actually used by the paper and its section-heading structure ... `</paper_signals>`<br>

`<task>` You are the Faithfulness Reviewer for academic papers. Given a paper automatically generated by an AI system, together with the raw research assets the system had access to, judge whether every statement in the paper can be traced to an asset, and whether there is fabrication or distortion. Your output is a strict JSON report with integer scores from 0 to 10 on five sub-dimensions plus a list of findings, each supported by a quoted snippet from the paper as evidence; do not judge novelty, experimental design, or writing style, all of which belong to other reviewers. `</task>`<br>

`<definition>` Faithfulness means that every verifiable statement in the paper has a matching source in the assets, or is a clearly legitimate generation such as an architecture diagram in a Method section. Unfaithfulness takes three specific forms: fabrication of a claim with no asset source; distortion of content that does exist in the assets but is altered in the paper; and phantom citation of a plausible-looking but non-existent bibliographic entry. `</definition>`<br>

`<rubric>` The five sub-dimensions are scored independently. I1 Theorem Fidelity checks that every theorem, lemma, corollary, or proposition in the paper is mathematically consistent with a source in `<assets>`; cross-directory inclusion of asset files (so the paper is not a self-contained document) is itself a defect. I2 Algorithm Fidelity checks that the paper's pseudocode agrees with the source algorithm on step order, loop nesting, update rules, and communication timing. I3 Numerical Fidelity checks that every specific number in the paper (metric, hyper-parameter, dataset size, assumption constant) is traceable to a number in the source experiment report or algorithm specification. I4 Figure Fidelity checks that every figure used in the paper has a legitimate source, that its caption is consistent with that source, and that it is referenced from the main text. I5 Citation Fidelity checks that every newly added bibliography entry (those listed under `added_bib_keys` in `<cite_check>`) is a real reference, that every citation is placed in a contextually appropriate section, and that no citation in `broken_cites` is unresolved. Each sub-dimension is scored on the same 0-10 scale: 10 when fully consistent with the source; 8 or 9 for minor phrasing or symbol differences with no semantic change; 6 or 7 for one core-level issue or several minor ones; 4 or 5 for multiple core issues; 2 or 3 when core content is fabricated or severely distorted; 0 or 1 when fabrication or mathematical errors are pervasive. `</rubric>`<br>

`<output_schema>` Return JSON only, with per-sub-dimension integer scores and a list of findings. Every finding must carry its severity, its location in the paper, the relevant asset reference when applicable, a short issue description, and a quoted paper snippet as evidence; all findings of severity high or medium must be listed, while low may be omitted. `</output_schema>`
\end{markdown}
	\end{LLMBox}
	\captionof{figure}{Core prompt structure for Phase~1 (Faithfulness).}
	\label{fig:prompt-writer-phase1}
\end{center}

\begin{center}
	\begin{LLMBox}[width=\textwidth, fontupper=\small]{Phase~2 evaluation prompt: Coverage (dimension~II)}
		\begin{markdown}
Run-time data blocks (inserted at the top of the prompt before the rubric):

`<assets>` ... full research materials for the paper under evaluation ... `</assets>`<br>
`<reference_signals>` ... programmatic anchor signals from the source assets ... `</reference_signals>`<br>
`<paper>` ... the manuscript produced by the system under review ... `</paper>`<br>
`<cite_check>` ... citation-set operations between source and paper ... `</cite_check>`<br>
`<paper_signals>` ... figures used by the paper, figures present in assets but unused, and section-heading structure ... `</paper_signals>`<br>

`<task>` You are the Asset Usage Judgment Reviewer. Given the input assets and the generated paper, judge whether the paper made reasonable editorial choices about which materials to include. Reasonable usage requires two things simultaneously: on the positive side, the core content of the source assets must be represented in the paper; on the negative side, the paper must not indiscriminately include non-core, redundant, or debug material as if it were important content. Your output is a strict JSON report with integer scores from 0 to 10 on five sub-dimensions plus a list of findings. Do not judge content correctness (Phase 1), writing style (Phase 3), or compilation and layout (Phase 4). `</task>`<br>

`<rubric>` The five sub-dimensions each weigh the missing-core side and the junk-inclusion side independently. II1 Theorem Usage asks whether the core theorems (those declared core in `<reference_signals>`) are actually presented and whether technical lemmas are kept out of the main text rather than dumped wholesale. II2 Algorithm Usage asks whether the problem definition, assumptions, and pseudocode are all present and whether design-rationale drafts are rewritten into the method discussion rather than pasted verbatim. II3 Experiment Usage asks whether every standalone experiment section from the source is represented in the paper, whether raw data tables are accompanied by analysis rather than stacked into an appendix, and whether no experiment sections appear that are not present in the source. II4 Figure Usage asks whether the core figures identified via `<paper_signals>` are used and whether debug or intermediate figures have been kept out of the main text. II5 Citation Usage asks whether foundational references for the paper's topic are cited (reviewers identify these from training knowledge; the rubric itself never names any specific paper) and whether the overall citation count is reasonable. Each sub-dimension is scored on the same 0-10 scale: 10 when the core is fully covered and no junk has accumulated; 8 or 9 for one minor omission or a couple of pieces of junk with the core still intact; 6 or 7 for one core omission or three or more junk pieces; 4 or 5 for multiple core omissions or large-scale junk; 2 or 3 for both widespread missing core and large-scale junk; 0 or 1 when there is almost no reasonable editorial judgment. `</rubric>`<br>

`<output_schema>` Return JSON only with per-sub-dimension integer scores and findings. The rubric itself must not name any specific paper, method, or reference; reviewers identify landmarks from their own training knowledge. `</output_schema>`
\end{markdown}
	\end{LLMBox}
	\captionof{figure}{Core prompt structure for Phase~2 (Coverage).}
	\label{fig:prompt-writer-phase2}
\end{center}

\begin{center}
	\begin{LLMBox}[width=\textwidth, fontupper=\small]{Phase~3 evaluation prompt: Writing Quality (dimension~III)}
		\begin{markdown}
Run-time data blocks (inserted at the top of the prompt before the rubric):

`<paper>` ... every .tex file plus the references file, with the bib filtered to the cited subset only ... `</paper>`<br>
`<paper_signals>` ... section-heading structure and figures actually used by the paper ... `</paper_signals>`<br>

`<task>` You are the Writing Quality Reviewer. Judge the paper purely as academic prose, that is, language, flow, argumentation, and integration of evidence. The assets are intentionally withheld from this phase so that writing quality is judged self-referentially. Your output is a strict JSON report with integer scores from 0 to 10 on four sub-dimensions plus findings with quoted evidence. Do not judge factual correctness (Phase 1), whether source material was used correctly (Phase 2), or compilation, layout, and template compliance (Phase 4). `</task>`<br>

`<rubric>` Four orthogonal sub-dimensions are scored independently. III1 Clarity concerns sentence form, terminology consistency, precision, grammar, and typography at the sentence-to-paragraph level. III2 Coherence, the emphasized sub-dimension for this project, concerns paragraph-to-paragraph and section-to-section logical flow: whether forward and backward references resolve to real content, whether the paper has a complete narrative arc rather than a collection of disconnected sub-papers, and whether the abstract is aligned with what the body actually delivers. III3 Argumentation concerns whether every core claim comes with motivation, explicit statement, supporting evidence, and closure. III4 Evidence Use concerns whether figures, tables, and citations are integrated into the argument at the place where the reader needs them rather than left decorative. Each sub-dimension is scored on the same 0-10 scale: 10 when the paper reads like a polished, well-edited piece of work; 8 or 9 when it is solid with only minor non-distracting issues; 6 or 7 when it is readable but noticeably rough; 4 or 5 when it is consistently problematic and requires effort to follow; 2 or 3 when it has widespread problems that make it difficult to read; 0 or 1 when it is incoherent or unintelligible. `</rubric>`<br>

`<output_schema>` Return JSON only with per-sub-dimension integer scores and findings. Every finding must carry its severity, its location in the paper, a short issue description, and a quoted snippet as evidence. `</output_schema>`
\end{markdown}
	\end{LLMBox}
	\captionof{figure}{Core prompt structure for Phase~3 (Writing Quality).}
	\label{fig:prompt-writer-phase3}
\end{center}

\begin{center}
	\begin{LLMBox}[width=\textwidth, fontupper=\small]{Phase~4a evaluation prompt: per-page vision description (stage~1 of Submittability)}
		\begin{markdown}
Run-time data block (a single page image is inserted, one per invocation):

`<page_image>` ... one rendered page of the PDF produced by the system under review ... `</page_image>`<br>

`<task>` You analyze a single PDF page from an academic paper draft in the target venue's format. Your job is to identify the content and structural elements present on the page, flag visual defects that would concern a journal editor, and emit one coarse page-level quality signal. A downstream reviewer will aggregate these per-page analyses to score the whole PDF's visual submittability; you focus on one page and do not infer beyond what is visible. `</task>`<br>

`<context>` The paper was auto-generated by an AI writing system given a LaTeX template. A journal editor would desk-reject if the PDF contains raw LaTeX as text, uses filenames as section headings, has data dumps replacing prose, or has non-English content. `</context>`<br>

`<definition>` Content types controlled enum: prose, equations, table, figure, algorithm pseudocode, code, csv data, references, raw latex source, header only, blank. Structure elements: title, author block, abstract, section heading, subsection heading, caption, page number, running header, footnote, reference item. Anomalies are drawn from a controlled enum grouped into three families: layout and whitespace (text overflow, image cut off, content cut off, mostly blank, excessive whitespace, isolated float on page, large whitespace gap, float at top only, float at bottom only, table columns overflow, equation broken line, caption separated from float); rendering (rendering garbage, font inconsistency, latex commands unrendered); content presentation (filename as section heading, non-English characters, missing page number, placeholder text visible, unresolved reference, empty bibliography fields). A four-level page flag is also emitted: clean (no anomalies), minor (one or two low-severity anomalies), major (one severe anomaly such as whole page as raw LaTeX, filename as heading, page dominated by csv data, or non-English content), and severe (multiple severe anomalies or the page is unreadable as a paper page). Key distinctions to avoid false positives: typeset pseudocode is not raw latex source; natural end-of-section or end-of-document whitespace is not a layout anomaly; a word hyphenated across a page boundary is standard LaTeX hyphenation and not content cut off; placeholder text visible is flagged only for literal template tokens (bibliography "Author Name" / "Author X" / "Unknown Author", body text `[TODO]` / `XXX` / `FIXME` / `<NAME>` / `<CITE>` / `(cite?)`, and stub author patterns like "A. Author" or generic "Firstname Lastname"), and is not flagged for methodology prose describing asset sources, for math subscripts such as $c_1$ or $x_i$, for citation keys rendered as short labels, for CSIAM slots the journal fills at acceptance ("Vol. x, No. x, pp. X-Y", "doi:", "©20xx Global-Science Press"), or for policy-required anonymization ("Anonymous Authors", "Anonymous Institution"); empty bibliography fields is flagged only when an entry clearly lacks author, year, or venue, and is not flagged for long author lists continuing onto the next page or for "et al." abbreviations; unresolved reference is flagged only for "??" or "[??]" marks actually rendered where a reference should appear. `</definition>`<br>

`<output_schema>` Return JSON only with the fields `content_type` (list from the content-types enum), `structure_elements` (list from the structure-elements enum), `anomalies` (list from the anomalies enum, empty if none), `anomaly_details` (free text with specifics, or empty string), `page_flag` (one of clean, minor, major, severe), `notable_text` (any unusual literal string on the page, or empty string), and `summary` (one factual sentence of at most 150 characters). `</output_schema>`<br>

`<prohibited>` Do not assign a 0-10 score at this stage (only the four-level page flag is used); do not judge content correctness, asset usage, writing, or compile status (those are other phases); do not invent anomaly codes outside the enum; do not add any text outside the JSON. `</prohibited>`
\end{markdown}
	\end{LLMBox}
	\captionof{figure}{Stage~1 of Phase~4: per-page structured description produced by a cheap vision model (\texttt{gemini-3.1-flash-lite-preview}), invoked once per page of each system's rendered PDF.}
	\label{fig:prompt-writer-phase4a}
\end{center}

\begin{center}
	\begin{LLMBox}[width=\textwidth, fontupper=\small]{Phase~4b evaluation prompt: visual scoring (stage~2 of Submittability)}
		\begin{markdown}
Run-time data blocks (inserted at the top of the prompt before the rubric):

`<page_descriptions>` ... concatenation of every per-page JSON produced by stage 1, in page order ... `</page_descriptions>`<br>
`<compile_facts>` ... IV1 counters extracted programmatically from the compile log: number of errors, undefined references, overfull and underfull hboxes, font warnings ... `</compile_facts>`<br>
`<template_facts>` ... IV3 checks: the documentclass line of main.tex and the presence or absence of banned package families ... `</template_facts>`<br>

`<task>` You are the Visual Layout Quality Reviewer for stage 2 of Phase 4. Given the per-page structured descriptions plus the programmatic compile and template facts, simulate a journal editor's thirty-second skim and score submittability on three sub-components on a 0-10 integer scale. You do not see the page images directly; score only from `<page_descriptions>`, and distinguish isolated issues affecting one or two pages from systemic issues repeated throughout the document, weighting the latter more heavily. `</task>`<br>

`<rubric>` IV1 Compile is a programmatic score derived from `<compile_facts>`: 10 iff every counter is zero, 0 iff no PDF was produced at all, and intermediate values degrade with the counter totals. IV2 Visual Layout is the LLM-scored component and is decomposed into three axes. `IV2_1` Layout Compliance covers margins, overflow, whitespace distribution, and float positioning (with signals text overflow, image cut off, excessive whitespace, isolated float on page, large whitespace gap, float at top or bottom only, table columns overflow, equation broken across lines, caption separated from float, mostly blank, missing page number). `IV2_2` Rendering Integrity covers whether content is rendered as intended (signals latex commands unrendered and non-English characters are both critical and cap `IV2_2` sharply, rendering garbage is critical, font inconsistency is a minor deduction). `IV2_3` Content Presentation covers whether the PDF reads as an academic paper rather than a compilation artifact (critical signals filename as section heading, placeholder text visible, unresolved reference, empty bibliography fields; caps also apply when csv data dominates pages, when the abstract is missing on page 1, or when there is no references section at the end). Each IV2 sub-axis uses the same 0-10 scale: 10 when the axis is essentially flawless; 8 or 9 when one or two isolated issues appear with no systemic pattern; 6 or 7 for noticeable but localized problems; 4 or 5 for widespread problems or multiple critical defects; 2 or 3 for systemic severe issues; 0 or 1 when the axis catastrophically fails. IV3 Template is a programmatic score from `<template_facts>`: 10 iff the documentclass matches the target template and no banned packages are used. The aggregation is equal weight: the final IV2 is the mean over `IV2_1`, `IV2_2`, `IV2_3` and over the three reviewer models, and the final IV submittability score is the mean of IV1, IV2, and IV3; when no PDF is produced, IV1 and IV2 are both set to 0 and only IV3 contributes. `</rubric>`<br>

`<output_schema>` Return JSON only with per-sub-component integer scores, a per-page count summary (total pages and counts of clean, minor, major, severe pages), and findings that cite specific page numbers and quote short excerpts from `<page_descriptions>` as evidence. `</output_schema>`
\end{markdown}
	\end{LLMBox}
	\captionof{figure}{Stage~2 of Phase~4: the three reviewer backends (\texttt{GPT-5.1}, \texttt{GPT-5.4}, \texttt{Gemini-3-Pro}) score visual submittability from the stage-1 page descriptions and the programmatic compile and template facts.}
	\label{fig:prompt-writer-phase4b}
\end{center}

\noindent\textbf{Per-Dimension Sub-Score Tables.}
Tables~\ref{tab:writer_phase1_sub}--\ref{tab:writer_phase4_sub} report the sub-dimension breakdown underlying Table~\ref{tab:writer_scores}. Every entry is already averaged across the three reviewer models with equal weight; the rightmost column reproduces the dimension score from the main table.

\begin{table}[htbp]
	\centering
	\caption{Phase~1 (Faithfulness) sub-dimension scores. I1~theorem, I2~algorithm, I3~numerical, I4~figure, I5~citation fidelity.}
	\label{tab:writer_phase1_sub}
	\begin{tabular}{lcccccc}
		\toprule
		System                                        & I1           & I2            & I3           & I4            & I5            & I             \\
		\midrule
		ReasFlow \texttt{WritingAgent} (\texttt{GPT-5.1})      & \textbf{10.00} & \textbf{10.00} & 9.00          & \textbf{10.00} & 9.67          & \textbf{9.73} \\
		Claude Code + ARIS (\texttt{GPT-5.4})                  & 7.67          & 9.33          & \textbf{9.67} & \textbf{10.00} & 8.67          & 9.07          \\
		Claude Code + ARIS (\texttt{GPT-5.1})                  & 4.67          & \textbf{10.00}         & 8.33          & 8.00           & 7.33          & 7.67          \\
		Claude Code, vanilla (\texttt{GPT-5.1})                & 4.67          & 7.33          & 8.67          & 8.67           & \textbf{10.00} & 7.87          \\
		Claude Code, vanilla (\texttt{GPT-5.4})                & 9.33          & \textbf{10.00}         & 4.33          & 7.33           & \textbf{10.00} & 8.20          \\
		Codex + ARIS (\texttt{GPT-5.1})                        & 4.67          & \textbf{10.00}         & 5.67          & 9.67           & 8.00          & 7.60          \\
		Codex + ARIS (\texttt{GPT-5.4})                        & 6.67          & \textbf{10.00}         & 8.67          & 9.00           & 8.00          & 8.47          \\
		Codex, vanilla (\texttt{GPT-5.1})                      & 4.67          & 9.33          & 8.00          & 8.00           & 8.33          & 7.67          \\
		Codex, vanilla (\texttt{GPT-5.4})                      & 5.00          & 9.33          & 4.00          & 8.33           & 5.67          & 6.47          \\
		ReasLingo (\texttt{GPT-5.1})                           & 5.00          & \textbf{10.00}         & 9.33          & 9.33           & 8.67          & 8.47          \\
		ReasLingo (\texttt{GPT-5.4})                           & 8.00          & \textbf{10.00}         & 8.00          & 8.67           & 7.67          & 8.47          \\
		\bottomrule
	\end{tabular}
\end{table}

\begin{table}[htbp]
	\centering
	\caption{Phase~2 (Coverage) sub-dimension scores. II1~theorem, II2~algorithm, II3~experiment, II4~figure, II5~citation usage.}
	\label{tab:writer_phase2_sub}
	\begin{tabular}{lcccccc}
		\toprule
		System                                        & II1           & II2           & II3            & II4            & II5           & II            \\
		\midrule
		ReasFlow \texttt{WritingAgent} (\texttt{GPT-5.1})      & 8.67          & \textbf{9.33} & \textbf{10.00} & \textbf{10.00} & \textbf{9.33} & \textbf{9.47} \\
		Claude Code + ARIS (\texttt{GPT-5.4})                  & 5.00          & 6.33          & \textbf{10.00} & \textbf{10.00} & \textbf{9.33} & 8.13          \\
		Claude Code + ARIS (\texttt{GPT-5.1})                  & 4.67          & 7.67          & 5.33           & 6.33           & 7.67          & 6.33          \\
		Claude Code, vanilla (\texttt{GPT-5.1})                & 4.67          & 8.00          & 7.33           & 7.00           & 9.00          & 7.20          \\
		Claude Code, vanilla (\texttt{GPT-5.4})                & \textbf{9.33} & \textbf{9.33} & 8.33           & \textbf{10.00} & 8.67          & 9.13          \\
		Codex + ARIS (\texttt{GPT-5.1})                        & 5.00          & 8.00          & 9.67           & 9.00           & 8.67          & 8.07          \\
		Codex + ARIS (\texttt{GPT-5.4})                        & 3.67          & 6.00          & 2.67           & 4.00           & 5.33          & 4.33          \\
		Codex, vanilla (\texttt{GPT-5.1})                      & 4.33          & 8.00          & 2.67           & 6.67           & 6.67          & 5.67          \\
		Codex, vanilla (\texttt{GPT-5.4})                      & 3.33          & 6.00          & 3.33           & 8.67           & 5.33          & 5.33          \\
		ReasLingo (\texttt{GPT-5.1})                           & 6.00          & 8.67          & 8.33           & 7.00           & 8.00          & 7.60          \\
		ReasLingo (\texttt{GPT-5.4})                           & 3.67          & 6.00          & 2.67           & 4.33           & 7.33          & 4.80          \\
		\bottomrule
	\end{tabular}
\end{table}

\begin{table}[t]
	\centering
	\caption{Phase~3 (Writing Quality) sub-dimension scores. III1~clarity, III2~coherence (project focus), III3~argumentation, III4~evidence use.}
	\label{tab:writer_phase3_sub}
	\begin{tabular}{lccccc}
		\toprule
		System                                        & III1          & III2          & III3          & III4          & III           \\
		\midrule
		ReasFlow \texttt{WritingAgent} (\texttt{GPT-5.1})      & 8.33          & \textbf{7.67} & \textbf{8.33} & \textbf{8.33} & \textbf{8.16} \\
		Claude Code + ARIS (\texttt{GPT-5.4})                  & 8.00          & 6.00          & 7.67          & 7.67          & 7.33          \\
		Claude Code + ARIS (\texttt{GPT-5.1})                  & 6.67          & 5.33          & 5.67          & 4.00          & 5.42          \\
		Claude Code, vanilla (\texttt{GPT-5.1})                & 8.00          & 7.33          & 7.67          & 5.67          & 7.17          \\
		Claude Code, vanilla (\texttt{GPT-5.4})                & \textbf{8.67} & 7.33          & 8.00          & 6.67          & 7.67          \\
		Codex + ARIS (\texttt{GPT-5.1})                        & 8.33          & 6.67          & 8.00          & 8.00          & 7.75          \\
		Codex + ARIS (\texttt{GPT-5.4})                        & 7.33          & 4.67          & 5.33          & 5.00          & 5.58          \\
		Codex, vanilla (\texttt{GPT-5.1})                      & 8.33          & 7.33          & 7.33          & 6.33          & 7.33          \\
		Codex, vanilla (\texttt{GPT-5.4})                      & 5.67          & 4.00          & 4.33          & 4.33          & 4.58          \\
		ReasLingo (\texttt{GPT-5.1})                           & 7.00          & 5.33          & 5.67          & 7.00          & 6.25          \\
		ReasLingo (\texttt{GPT-5.4})                           & 7.00          & 4.33          & 4.33          & 4.67          & 5.08          \\
		\bottomrule
	\end{tabular}
\end{table}

\begin{table}[t]
	\centering
	\caption{Phase~4 (Submittability) sub-components. IV1~compile (programmatic), IV2~visual layout (three LLM sub-scores IV2\_1~layout compliance, IV2\_2~rendering integrity, IV2\_3~content presentation, each averaged over three reviewers), IV3~template compliance (programmatic). A dash (---) in a IV2 sub-column indicates no PDF was produced; in that case IV1$=$IV2$=$0 and only IV3 contributes to IV.}
	\label{tab:writer_phase4_sub}
	\begin{tabular}{lccccccc}
		\toprule
		System                                        & IV1            & IV2\_1         & IV2\_2          & IV2\_3          & IV2           & IV3            & IV             \\
		\midrule
		ReasFlow \texttt{WritingAgent} (\texttt{GPT-5.1})      & \textbf{10}   & 9.00          & \textbf{10.00} & \textbf{10.00} & 9.67          & \textbf{10}   & 9.87           \\
		Claude Code + ARIS (\texttt{GPT-5.4})                  & 6             & \textbf{10.00} & \textbf{10.00} & 3.67          & 7.89          & \textbf{10}   & 7.56           \\
		Claude Code + ARIS (\texttt{GPT-5.1})                  & 0             & ---           & ---            & ---            & 0.00          & \textbf{10}   & 2.00           \\
		Claude Code, vanilla (\texttt{GPT-5.1})                & 1             & 8.00          & 5.33           & 3.33           & 5.55          & \textbf{10}   & 4.62           \\
		Claude Code, vanilla (\texttt{GPT-5.4})                & \textbf{10}   & \textbf{10.00} & \textbf{10.00} & \textbf{10.00} & \textbf{10.00} & \textbf{10}   & \textbf{10.00} \\
		Codex + ARIS (\texttt{GPT-5.1})                        & \textbf{10}   & 7.67          & 9.67           & 4.33           & 7.22          & \textbf{10}   & 8.89           \\
		Codex + ARIS (\texttt{GPT-5.4})                        & \textbf{10}   & 2.33          & 5.33           & 1.67           & 3.11          & \textbf{10}   & 7.24           \\
		Codex, vanilla (\texttt{GPT-5.1})                      & \textbf{10}   & 2.33          & 9.67           & 1.67           & 4.56          & \textbf{10}   & 7.82           \\
		Codex, vanilla (\texttt{GPT-5.4})                      & \textbf{10}   & 3.00          & 1.33           & 1.00           & 1.78          & \textbf{10}   & 6.71           \\
		ReasLingo (\texttt{GPT-5.1})                           & \textbf{10}   & 8.67          & \textbf{10.00} & 2.67           & 7.11          & \textbf{10}   & 8.84           \\
		ReasLingo (\texttt{GPT-5.4})                           & \textbf{10}   & 2.67          & 9.67           & 1.67           & 4.67          & \textbf{10}   & 7.87           \\
		\bottomrule
	\end{tabular}
\end{table}

\noindent\textbf{Discussion. }
The sub-score tables above indicate that faithfulness alone is not discriminative enough to separate the compared systems. Most systems receive an I2 algorithm-fidelity score of at least $9.33$, because the task prompt explicitly marks algorithm pseudocode as a hard red line and nearly every system copies the pseudocode correctly; faithfulness differentiates the compared systems mainly on I1 theorem fidelity, where \texttt{WritingAgent} scores $10.00$ against a median of about $5.00$ for baselines that drop most theorems, and on I3 numerical fidelity, where systems that copy numbers carefully separate from systems that round aggressively or regenerate.

Coverage is where the overall ranking is mostly decided. The six systems at the bottom of the overall table all have II~Coverage no higher than $6$, typically because they drop entire experiment sections or leave out parts of the algorithm specification; in contrast, systems that place above this threshold all cover the core experiments and algorithm ingredients, and the remaining gap between them is accounted for by writing quality and submittability.

Submittability finally exposes failure modes that Phases~I to~III cannot see. Two systems with otherwise reasonable content scores,  \textit{Claude Code + ARIS} under \texttt{GPT-5.1} and \textit{Claude Code, vanilla} under \texttt{GPT-5.1},  fall to the bottom because of compilation or rendering failure: the former fails to produce a PDF at all because Chinese characters leak into the source and break the compile, while the latter compiles but accumulates $45$ errors, $19$ undefined citations, and $13$ overfull hboxes. Such failures are invisible from the content-facing phases, yet they decisively determine whether the manuscript could actually be submitted, which is why the overall score is an equal-weight mean across all four dimensions rather than a weighted combination that could conceal them.

\section{Unpolished ReasFlow-Generated Manuscripts}\label{app:generated-reasflow-papers}

The following pages reproduce the unpolished ReasFlow-generated manuscripts for the five case studies, in the order FedSLoP, Subspace SCAFFOLD, SUDA-Muon, MC-ADSGD, and Retraction-free EXTRA. These manuscripts preceded the human-expert polishing reflected in the arXiv papers cited in the main text.

\includepdf[pages=-,pagecommand={\thispagestyle{empty}}]{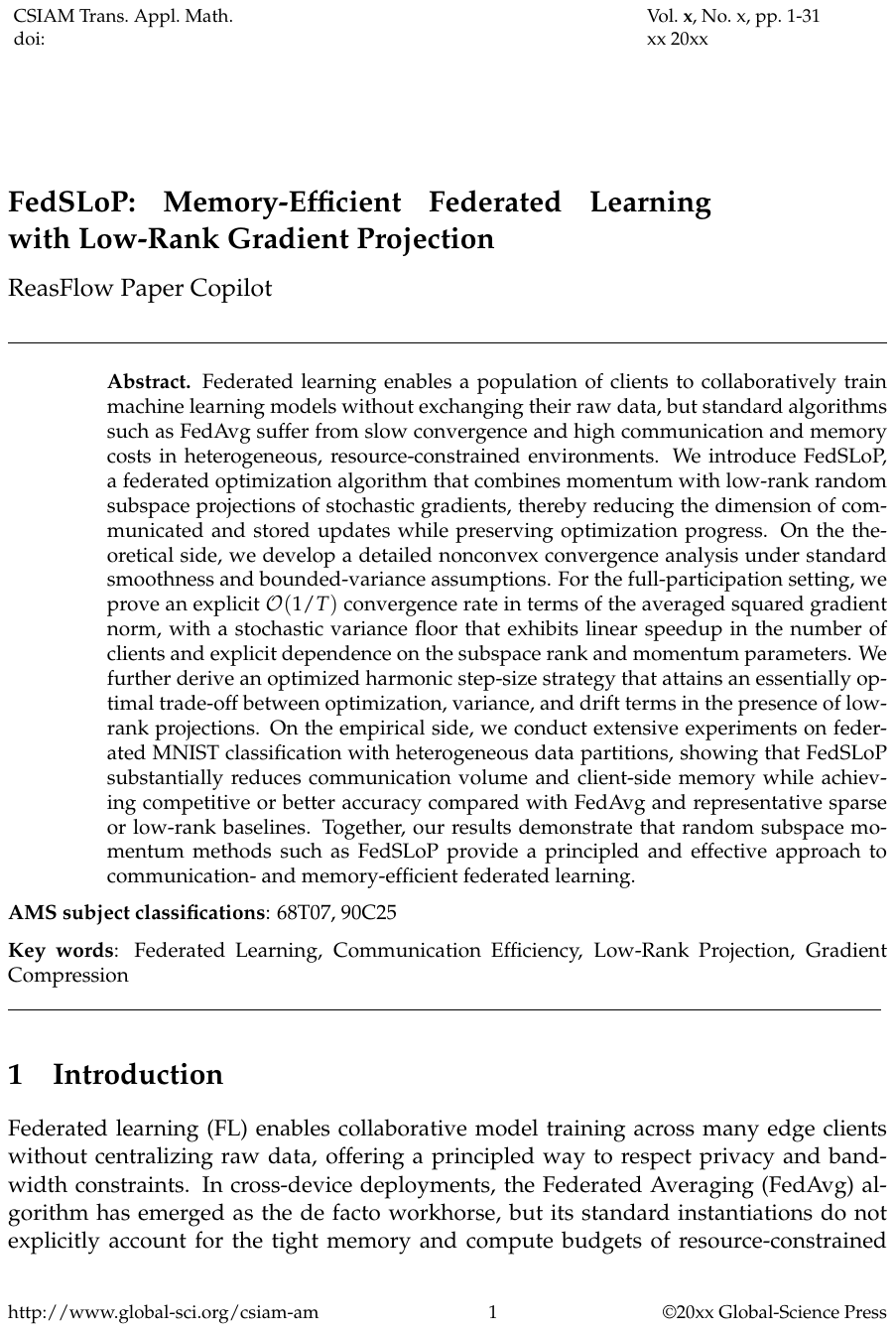}

\includepdf[pages=-,pagecommand={\thispagestyle{empty}}]{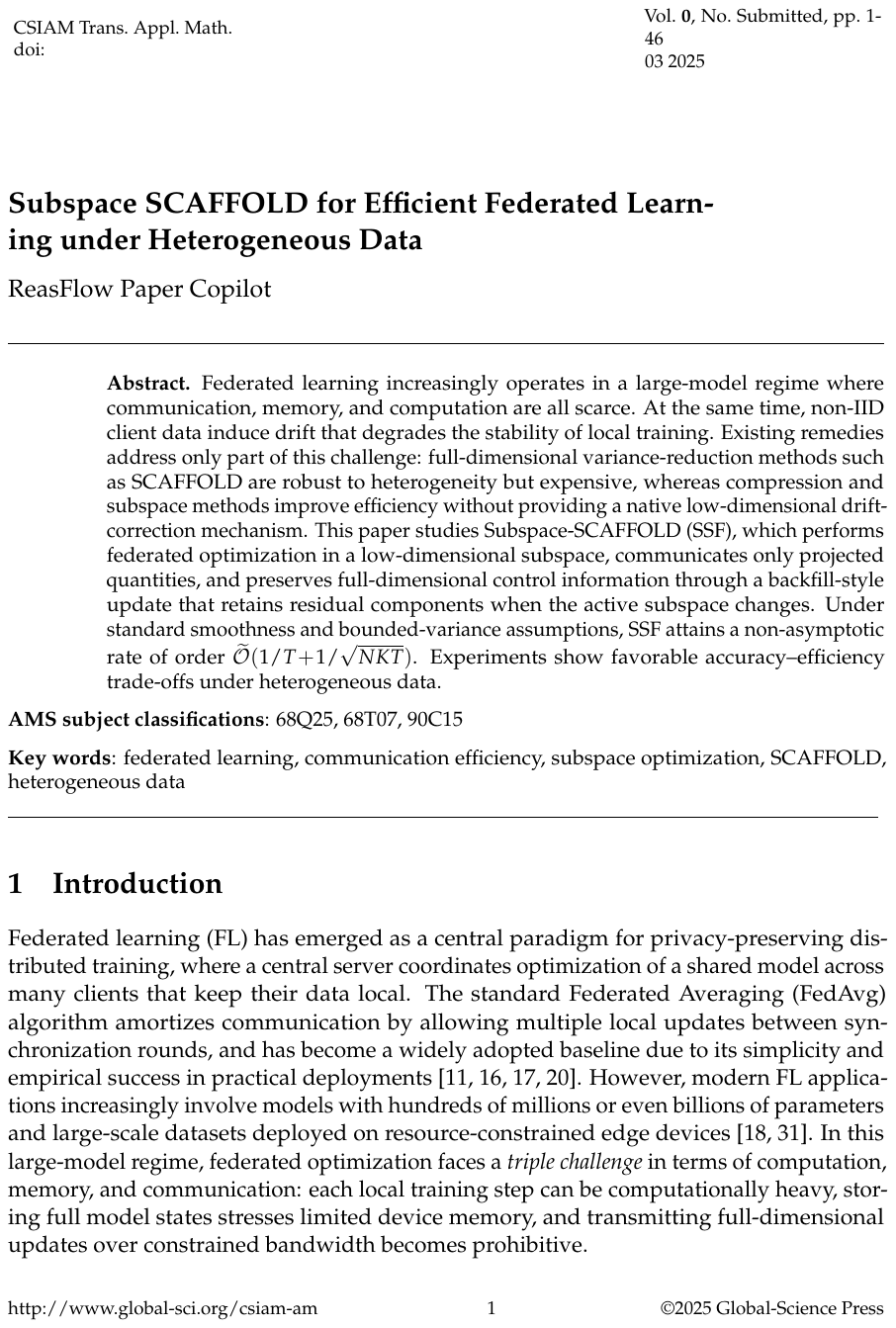}

\includepdf[pages=-,pagecommand={\thispagestyle{empty}}]{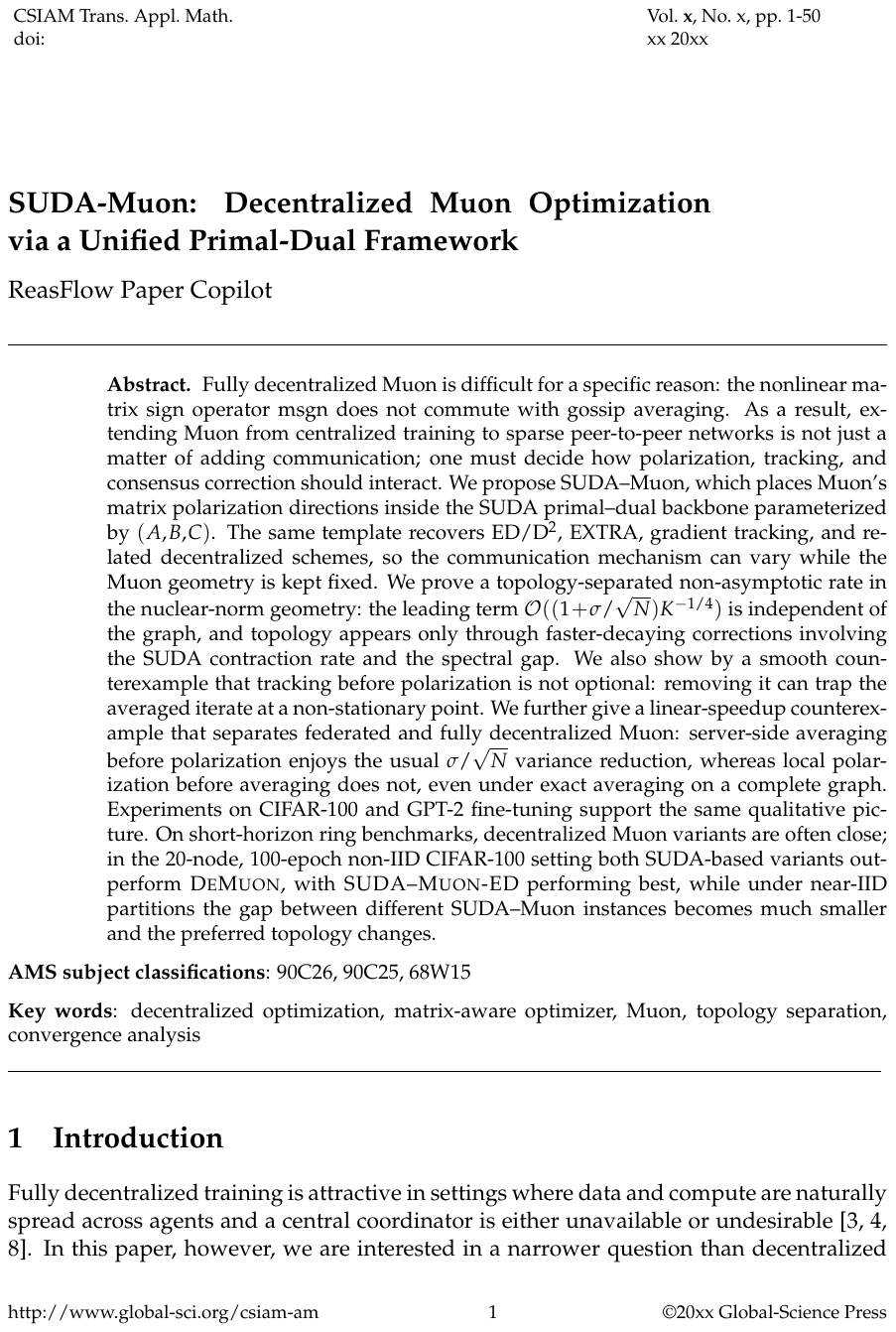}

\includepdf[pages=-,pagecommand={\thispagestyle{empty}}]{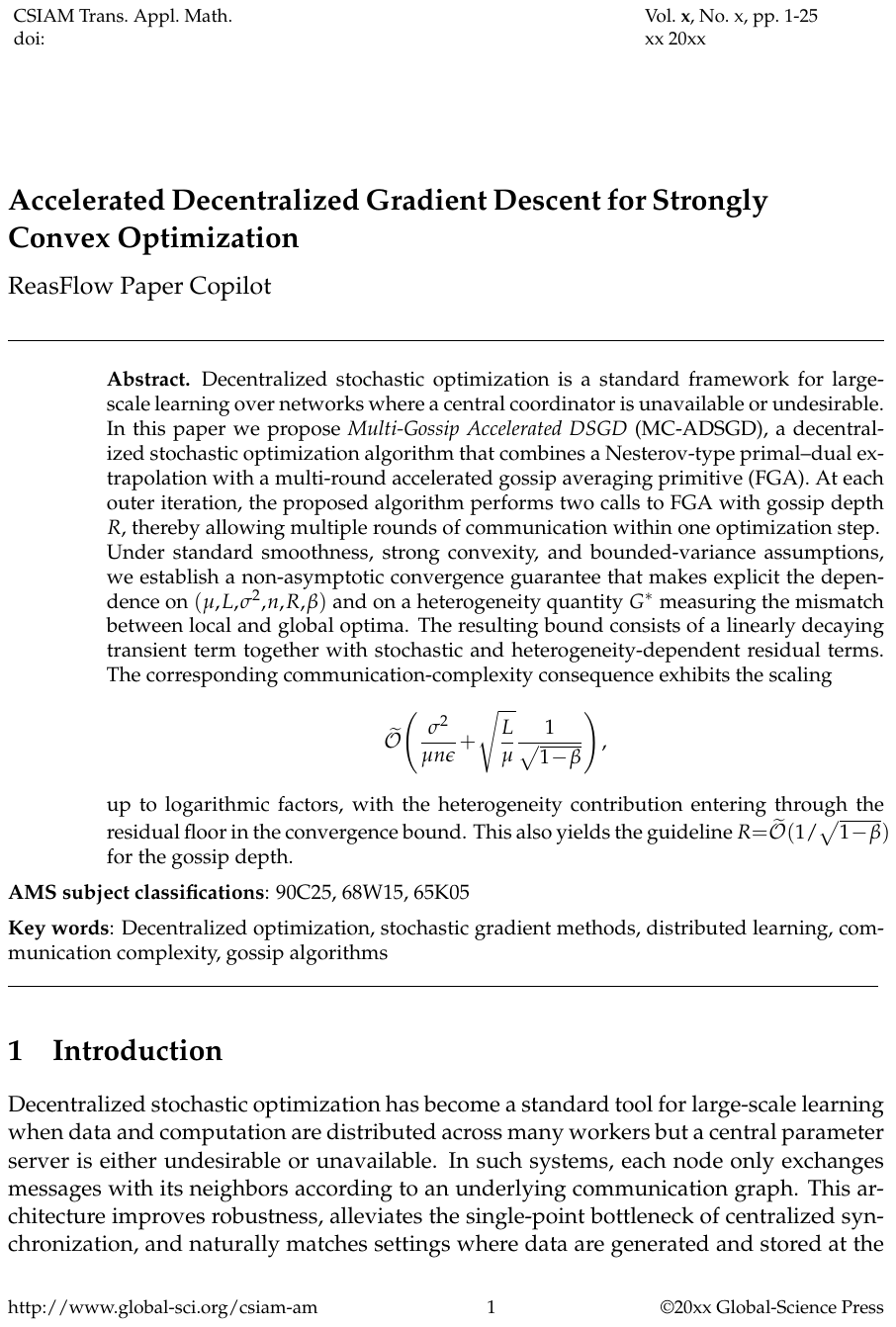}

\includepdf[pages=-,pagecommand={\thispagestyle{empty}}]{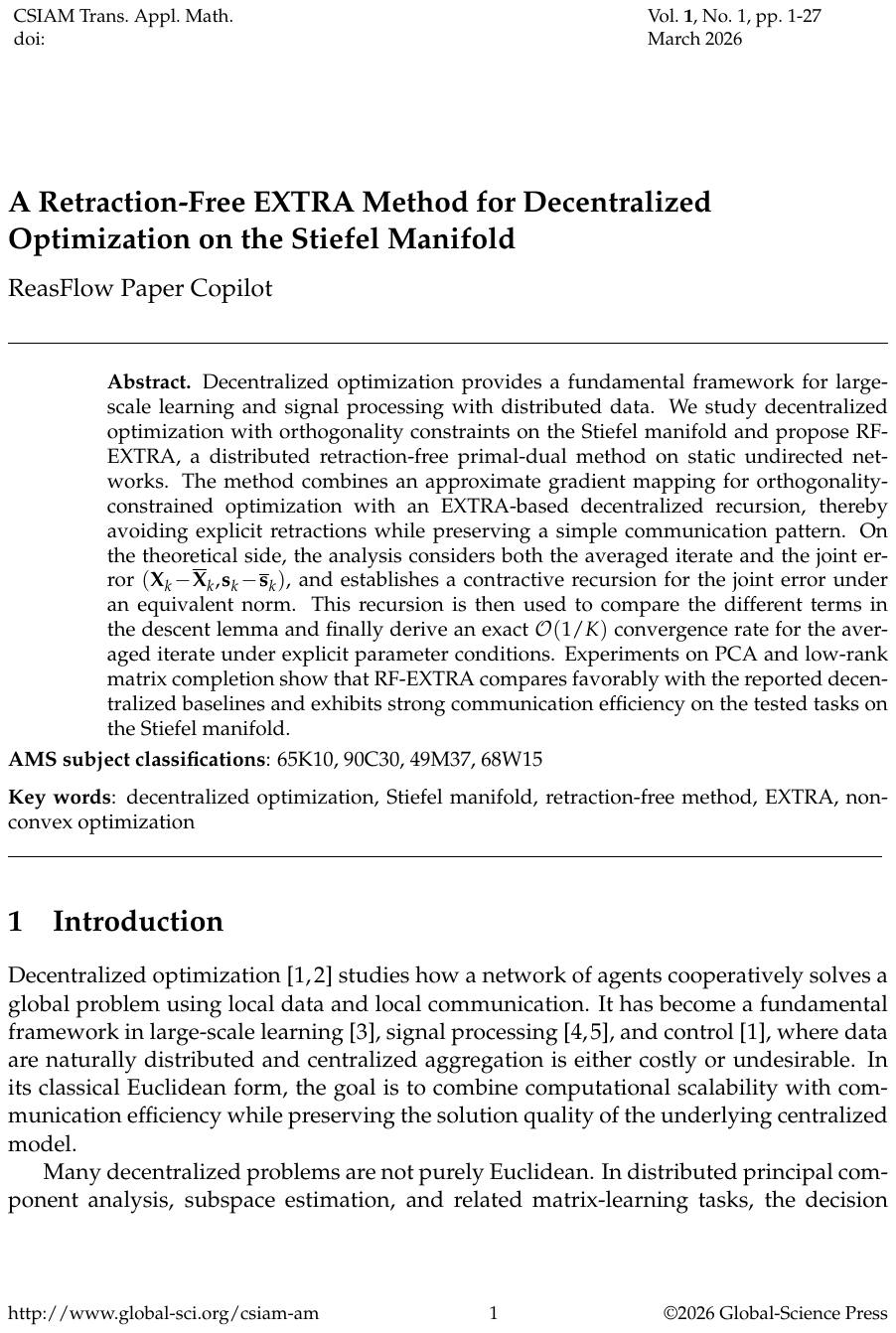}

\end{document}